\documentclass[10pt,twocolumn,letterpaper]{article}

\usepackage{iccv}
\usepackage[accsupp]{axessibility}

\usepackage{xcolor, booktabs, multirow, array, placeins, pifont, enumitem, bm, soul, nicefrac, boldline, tocloft, wrapfig, graphicx, caption, comment, xspace, colortbl}
\usepackage{anyfontsize}
\usepackage{symbols}
\usepackage{multirow}
\usepackage[labelformat=simple]{subcaption}

\usepackage{amsmath,amsfonts,bm}

\let\hat\widehat
\let\tilde\widetilde

\def\tabref#1{Tab.~\ref{#1}}

\def\1{\bm{1}}

\def\sp{space}

\def\va{{\bm{a}}}
\def\vb{{\bm{b}}}

\def\vf{{\bm{f}}}
\def\vg{{\bm{g}}}

\def\vn{{\bm{n}}}

\def\vp{{\bm{p}}}

\def\vt{{\bm{t}}}

\def\mB{{\bm{B}}}

\def\mF{{\bm{F}}}

\def\mM{{\bm{M}}}
\def\mN{{\bm{N}}}

\def\mQ{{\bm{Q}}}

\def\mS{{\bm{S}}}

\def\mX{{\bm{X}}}
\def\mY{{\bm{Y}}}
\def\mZ{{\bm{Z}}}

\DeclareMathAlphabet{\mathsfit}{\encodingdefault}{\sfdefault}{m}{sl}
\SetMathAlphabet{\mathsfit}{bold}{\encodingdefault}{\sfdefault}{bx}{n}
\newcommand{\tens}[1]{\bm{\mathsfit{#1}}}

\def\tF{{\tens{F}}}

\def\gA{{\mathcal{A}}}

\def\gC{{\mathcal{C}}}
\def\gD{{\mathcal{D}}}

\def\gG{{\mathcal{G}}}

\def\gL{{\mathcal{L}}}

\def\gN{{\mathcal{N}}}

\def\gP{{\mathcal{P}}}

\def\gT{{\mathcal{T}}}
\def\gU{{\mathcal{U}}}
\def\gV{{\mathcal{V}}}

\def\sR{{\mathbb{R}}}

\DeclareMathOperator*{\argmin}{arg\,min}

\usepackage{arydshln}
\usepackage[export]{adjustbox}
\newcommand*{\ShowNotes}{}
\definecolor{darkred}{rgb}{0.7,0.1,0.1}
\definecolor{darkgreen}{rgb}{0.1,0.7,0.1}
\definecolor{darkyellow}{rgb}{1.0,0.7,0.0} 
\definecolor{cyan}{rgb}{0.7,0.0,0.7}
\definecolor{dblue}{rgb}{0.2,0.2,0.8}
\definecolor{maroon}{rgb}{0.76,.13,.28}
\definecolor{burntorange}{rgb}{0.81,.33,0}
\definecolor{tealblue}{rgb}{0.212,0.459, 0.533}
\definecolor{darkred}{HTML}{C44D52}
\definecolor{darkblue}{HTML}{4C72B0}
\definecolor{pink}{HTML}{e050fa}

\definecolor{pp}{rgb}{0.43921569, 0.18823529, 0.62745098}
\definecolor{rr}{rgb}{0.5254902 , 0.00784314, 0.12941176}
\definecolor{bb}{rgb}{0.09019608, 0.23529412, 0.37647059}
\definecolor{yy}{rgb}{0.49803922, 0.3372549 , 0.0}
\definecolor{gg}{rgb}{0.02352941, 0.3372549 , 0.17647059}

\ifdefined\ShowNotes
  \newcommand{\colornote}[3]{{\color{#1}\bf{#2: #3}\normalfont}}
\else
  \newcommand{\colornote}[3]{}
\fi

\newcommand{\eat}[1]{}

\newcommand{\xmark}{\ding{55}}
\definecolor{checkmark}{HTML}{305AFF}
\definecolor{xmark}{HTML}{E62020}
\newcommand{\ccheck}{{\textcolor{checkmark}{\large\checkmark}}}
\newcommand{\ccross}{{\textcolor{xmark}{\large\xmark}}}

\newcommand{\ours}{{LTOAD}\xspace}
\newcommand{\myparagraph}[1]{{\noindent \bf #1.}}
\definecolor{OursColor}{HTML}{F2F3F4}

\newcommand{\citeGzero}{\cite{guo2024cada, marimont2021anomaly, mou2023rgi, zhang2021defect, rezende2015variational, rudolph2021same, yu2021fastflow, salehi2021multiresolution, zavrtanik2021draem, hyun2024reconpatch, tien2023revisiting, zhang2023destseg, li2021cutpaste, liu2025learning, fuvcka2025transfusion, 2024anogen, roth2022towards, cohen2020sub, defard2021padim, deng2022anomaly, yao2024glad}}
\newcommand{\citeGone}{\cite{zhao2023omnial, zhu2024fine, patra2024revisiting, Li2024MuSc, strater2024generalad, shi2023dissolving, yao2024glad, he2024learning,chen2024unified, gao2024onenip, lee2025continuous, isaac2024towards, he2024mambaad, you2022unified, zhang2023exploring, he2024diffusion, mengmoead, yao2024hierarchical, bai2024dual}}

\usepackage{graphbox}
\usepackage{algorithm}
\usepackage{algorithmic}

\usepackage{tabularx,booktabs}
\newcolumntype{Y}{>{\centering\arraybackslash}X}

\definecolor{iccvblue}{rgb}{0.21,0.49,0.74}
\usepackage[pagebackref, breaklinks, colorlinks, allcolors=iccvblue]{hyperref}

\def\paperID{6515}
\def\confName{ICCV}
\def\confYear{2025}

\title{Toward Long-Tailed Online Anomaly Detection \\ through Class-Agnostic Concepts}

\author{
    Chiao-An Yang$^{1}$  \quad
    Kuan-Chuan Peng$^{2}$ \quad
    Raymond A. Yeh$^{1}$ 
    \\
    $^{1}$Department of Computer Science, Purdue University
    \quad
    $^{2}$Mitsubishi Electric Research Laboratories
    \\
    $^{1}${\tt\small \{yang2300,rayyeh\}@purdue.edu} \quad
    $^{2}${\tt\small kpeng@merl.com}
}

\begin{document}
\maketitle
\begin{abstract}
Anomaly detection (AD) identifies the defect regions of a given image. Recent works have studied AD, focusing on learning AD without abnormal images, with long-tailed distributed training data, and using a unified model for all classes. In addition, online AD learning has also been explored. In this work, we expand in both directions to a realistic setting by considering the novel task of long-tailed online AD (LTOAD). We first identified that the offline state-of-the-art LTAD methods cannot be directly applied to the online setting. Specifically, LTAD is class-aware, requiring class labels that are not available in the online setting. To address this challenge, we propose a class-agnostic framework for LTAD and then adapt it to our online learning setting.
Our method outperforms the SOTA baselines in most offline LTAD settings, including both the industrial manufacturing and the medical domain. In particular, we observe $+$4.63\% image-AUROC on MVTec even compared to methods that have access to class labels and the number of classes. In the most challenging long-tailed online setting, we achieve +0.53\% image-AUROC compared to baselines. Our LTOAD benchmark is released here\footnote{\url{https://doi.org/10.5281/zenodo.16283852}}.
\end{abstract}
    
\section{Introduction}
\label{sec:intro}

Anomaly detection (AD)~\cite{cao2024survey, pang2021deep, zhang2024ader}  is the task of identifying defect regions in a given image.
This task is important due to its high valued applications in industrial manufacturing~\cite{saleh2022tire, ma2023review, yang2025defect, ad3, bergmann2019mvtec, zou2022spot, arodi2024cableinspect, fridman2021changechip} and medical~\cite{ad3, HeadCT, BrainMRI, Br35h} settings.

Recently, LTAD~\cite{ho2024ltad} explored the long-tailed (LT)~\cite{zhang2024systematic} setting of unsupervised one-class AD, where abnormal images are not given during training and the training set is unevenly distributed. This long-tailed distribution assumes that images of certain classes significantly outnumber others, which is a realistic assumption for real-world applications. 
Importantly, LTAD shows that prior work on evenly distributed data performs suboptimally on LT data. 
Meanwhile, recent works~\cite{doshi2020continual, liu2024unsupervised, tang2024incremental, gao2023towards, mcintoshunsupervised, wei2024fs} approach AD with online learning to leverage abnormal images during test time. However, they do not study the case where the classes follow a long-tail distribution, where the performance gap of head and tail classes in the online setting may be significant during offline training. This motivates us to study the long-tailed online AD (LTOAD) task by proposing a benchmark and a novel model.

\begin{figure}[t]
    \centering
    \includegraphics[width=\linewidth]{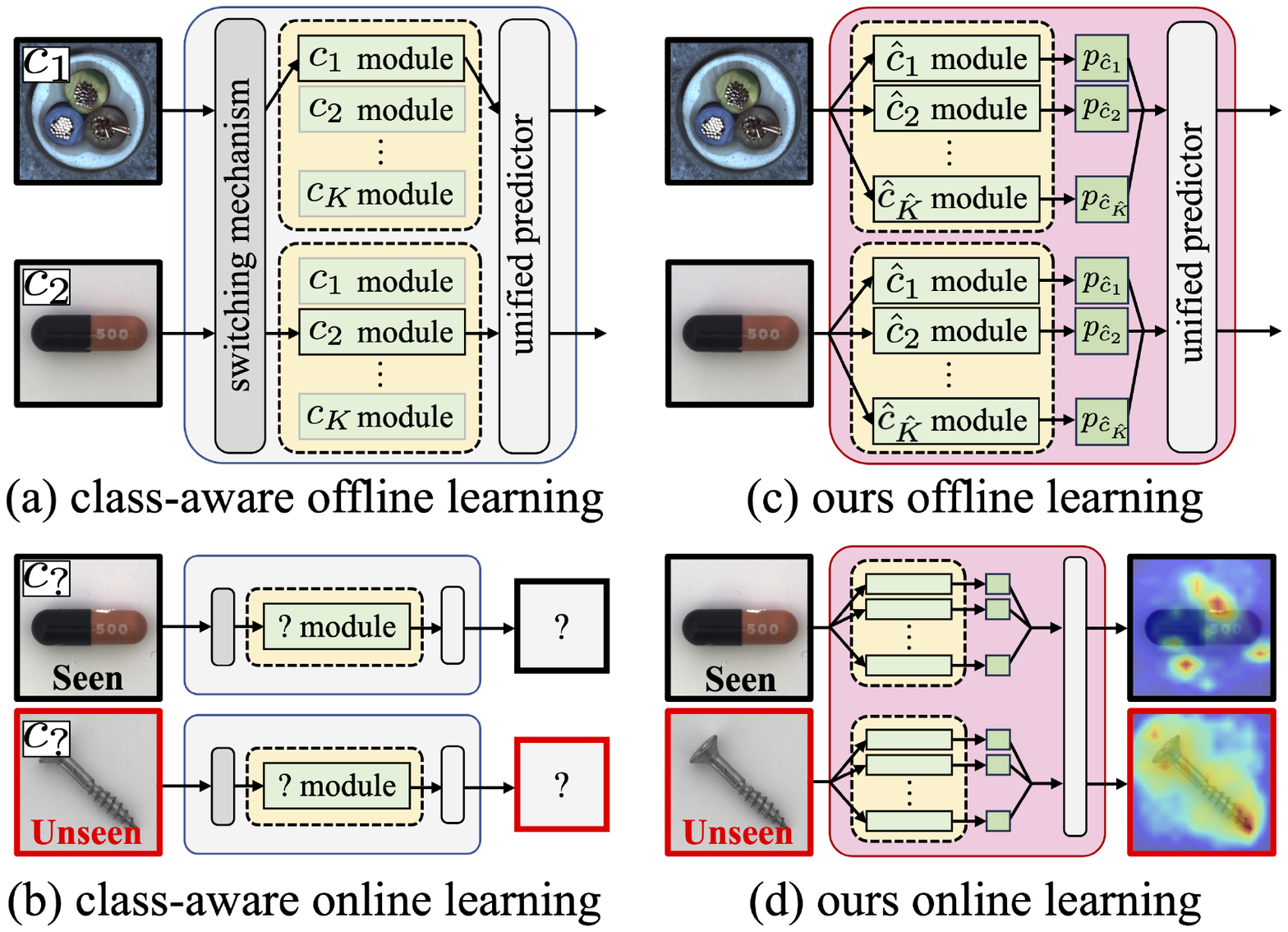}
    \vspace{-0.35cm}
    \caption{
        \textbf{Comparison of \ours~to class-aware anomaly detection methods on offline and online learning.} (a) Class-aware methods have a class-specific module for each class $c$ in the class set $\gC$. (b) These methods cannot work in online learning when the class labels are unavailable. (c) \ours~solves this problem by learning a concept set $\hat \gC$ to approximate $\gC$. We note that $\hat K = |\hat \gC|$ does not need to match $K = |\gC|$. (d) In online learning, \ours~can weight input images of {\bf seen} and {\color{red}{\bf unseen}} classes with existing concept-specific modules, \ie, $\{ p_{\hat c}\}_{\hat c \in \hat \gC}$.
    }
    \label{fig:teaser}
    \vspace{-0.2cm}
\end{figure}

At first glance, it may seem that one can take a state-of-the-art (SOTA) long-tailed offline model, \eg, LTAD~\cite{ho2024ltad}, and combine it with online learning to solve LTOAD. However, several limitations within the LTAD's framework prevent it from being effective in the online setting. 
First, LTAD is a class-aware method, where they assume that the class information is given. However, class labels are not available in the online setting. Instead, 
a \textbf{class-agnostic} method is required as illustrated in~\figref{fig:teaser}. 
Second, LTAD's encoder-decoder architecture does not leverage the latest vector quantization VAE~\cite{lu2023hierarchical, marimont2021anomaly, gangloff2022leveraging}, which is more effective, as shown in recent work.
Third, the prompt learning designs of LTAD and other AD works~\cite{AnomalyCLIP, qu2024vcpclip, chen2024hfp, jeong2023winclip, li2024fade, zhu2024toward, cao2024adaclip, huang2024adapting, cao2023segment, zhu2024fine, li2024promptad} ignore some aspects of abnormal prompts, \eg, LTAD does not support class-specific abnormal prompts. 

To address these challenges, we propose a novel model that is class-agnostic, with a vector quantized VAE, and a comprehensive prompt learning framework, for the task of long-tailed online anomaly detection. 
Experiments show that our method outperforms the SOTA baselines significantly by at least 4\% on several detection benchmarks without access to the class labels and the number of classes.

In the following sections, we first discuss how to make existing class-aware methods class-agnostic, specifically on LTAD (\secref{sec:concept}). Next, we define the task of long-tailed online AD and present our solution (\secref{sec:task}). Finally, we provide details of each proposed module (\secref{sec:details}), followed by the experimental results (\secref{sec:exp}) and related works (\secref{sec:rel}).

{\noindent\bf Our contributions are summarized as follows:}
\begin{itemize}
    \item We propose the task and benchmark for \textbf{long-tailed online anomaly detection (LTOAD)}. 
    \item We propose a class-agnostic framework for both offline and online learning that removes the constraint of requiring additional class information in previous work. Along with vector-quantization VAE and a comprehensive prompt learning scheme that benefits both offline and online learning.
    \item \ours~outperforms SOTA offline long-tailed methods and online baselines on MVTec, VisA, DAGM, and Uni-Medical. It also shows promising generalizability to cross-dataset settings.
\end{itemize}

\section{Making LTAD class-agnostic}
\label{sec:concept}

\subsection{Background}
\myparagraph{Anomaly Detection (AD)}
Given an image $\mX \in \sR^{H \times W \times 3}$ of height $H$ and width $W$, an AD model ${\tt F}_\theta$, parametrized by $\theta$, aims to predict an abnormal map $\hat\mY \in \{0,1\}^{H \times W}$ or an abnormal label $y \in \{0,1\}$) indicating whether the image is abnormal or not, where $\hat \mY_i = {\tt F}_\theta(\mX_i)$.

Beyond the standard setting, LTAD~\cite{ho2024ltad} proposes a more challenging setup of unsupervised unified AD that focuses on an unevenly distributed training set. Each class $c$ in the class set $\gC$ belongs to either the set of ``head classes'' $\gC^{\tt head}$ or ``tail classes'' $\gC^{\tt tail}$ depending on the number of times a class appears in the data set. That is, images from the classes in $\gC^{\tt head}$ significantly outnumber those from $\gC^{\tt tail}$. 

\myparagraph{Vision and language models (VLM) in AD}
Recent works 
\cite{huang2024adapting, li2024promptad, li2024fade, AnomalyCLIP, cao2023segment, li2024clipsam, cao2024adaclip, jeong2023winclip, chen2024hfp, qu2024vcpclip, zhang2023gpt, gu2024agpt, ho2024ltad}
have shown the benefit of foundation VLM for AD. A foundation model consists of a image encoder ${\tt E}^{\tt I}$ and an text encoder ${\tt E}^{\tt T}$. The encoders encode images and text
to a shared space such that cosine similarly between the embeddings captures the relationship between image and text. We use the notation $\left< \cdot \right>$ for cosine similarity that implicitly supports broadcasting, \ie, when $\left< \cdot \right>$ is between a matrix with a vector, we broadcast on the spatial dimension. 

In more detail, recent works~\cite{he2024mambaad, mengmoead, lu2023hierarchical, ho2024ltad} extract fine-grained image features from intermediate layers or blocks of ${\tt E}^{\tt I}$.  We denote these intermediate visual feature maps as $\mF^{\tt i}$ and the final visual feature vectors as $\vf^{\tt f}$, \ie, $(\mF^{\tt i}, \: \vf^{\tt f}) = {\tt E}^{\tt I}(\mX)$. The precise descriptions of $\mF^{\tt i}$ and $\vf^{\tt f}$ are provided in Appendix~\ref{supp:implement}.

\myparagraph{Class-aware AD}
Existing works~\cite{lu2023hierarchical, huang2024adapting, li2024promptad, li2024fade, cao2023segment, li2024clipsam, cao2024adaclip, jeong2023winclip, chen2024hfp, qu2024vcpclip, zhang2023gpt, gu2024agpt, ho2024ltad, wei2024fs, gao2023towards, mcintoshunsupervised} consist of class-aware modules in their architecture.
These works require an additional class label $c \in \gC$ for each image as input. Specifically, they assume that the class $c$ of the input image is also provided to the model, \ie, $\hat \mY_i = {\tt F}_\theta(\mX_i, c_i)$. 
As depicted in Fig.~\ref{fig:teaser}, a hard switching mechanism is used in their framework to select the output of a specific class-aware module, \ie $\mM_{c_i}$, to use for each instance. In this work, we do not assume that the class label is provided, as it would typically be unavailable in the online setting. Instead, we develop a class-agnostic approach, which we describe next.

\begin{figure*}[t]
    \centering
    \includegraphics[width=0.81\linewidth]{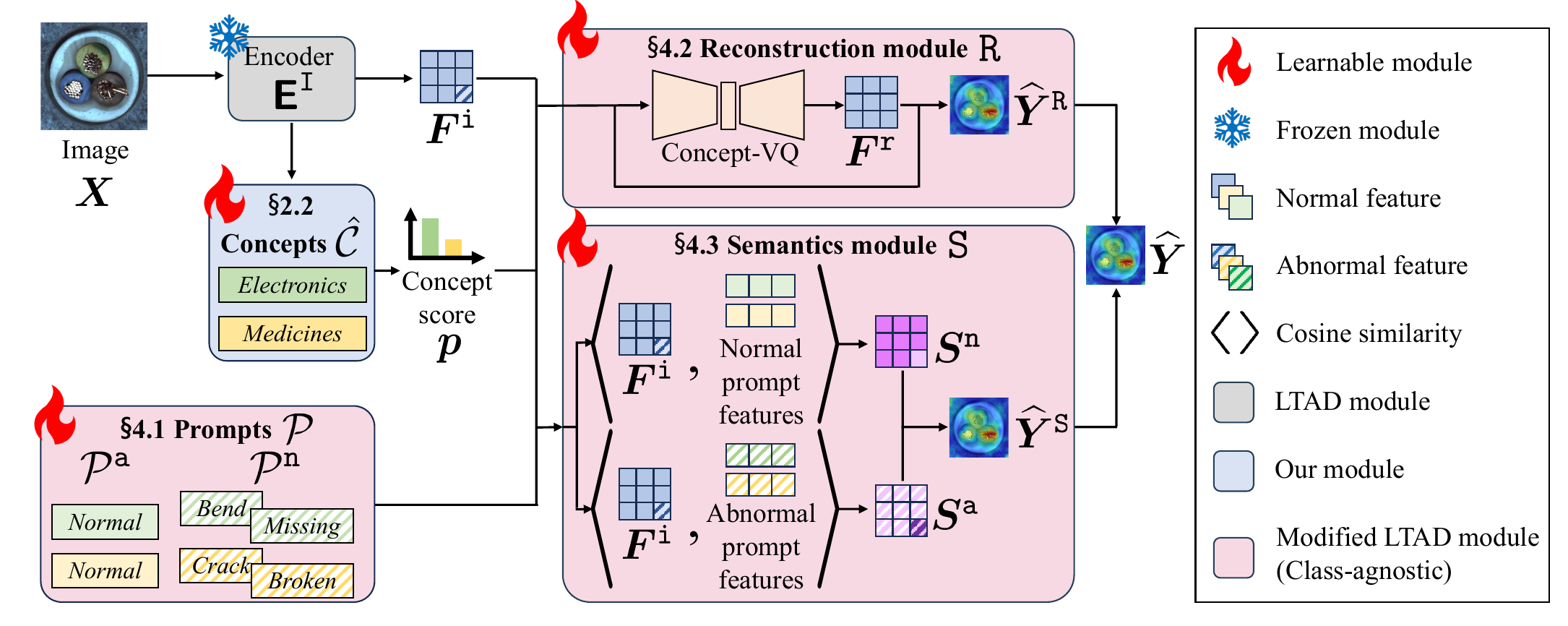}
    \vspace{-0.25cm}
    \caption{
        \textbf{Proposed class-agnostic pipeline.}  We construct $\hat \gC$, and the correspondent normal prompts $\gP^{\tt n}$ and abnormal prompts $\gP^{\tt a}$. The concept score $\vp$ is assigned to each image $\mX$ by computing the similarity between $\mX$ and each $\hat c \in \hat \gC$. It then controls the soft switching mechanism in our class-agnostic reconstruction module $\tt R$ and semantics module $\tt S$. $\tt R$ reconstructs $\mF^{\tt i}$ into $\mF^{\tt r}$ through Concept VQ and output $\hat \mY^{\tt R}$ by measuring the dissimilarity between $\mF^{\tt i}$ and $\mF^{\tt r}$. $\tt S$ compares the similarity map $\mS^{\tt n}$ of $\mF^{\tt i}$ and normal prompt features and the similarity map $\mS^{\tt a}$ of $\mF^{\tt i}$ and abnormal prompt features to output $\hat \mY^{\tt S}$. The final prediction $\hat \mY$ is aggregated from $\hat \mY^{\tt R}$ and $\hat \mY^{\tt S}$. 
    }
    \label{fig:pipeline}
    \vspace{-0.25cm}
\end{figure*}

\subsection{Concept set for removing class-awareness}
To remove the requirement of having the class label $c$, we introduce a concept set $\hat \gC$ where we assume that the class information $c$ can be represented as a composition of multiple concepts in $\hat \gC$. For example, the class `transistor' is related to and derived from concepts `semiconductors' and `circuits'. In other words, for each image of class $c$, instead of applying a hard one-hot label, we employ a soft weighting mechanism and assign a soft label $\vp \in \sR^{\hat K}$ where $\hat K = |\hat \gC|$. There are a few advantages. First, we can reduce $\hat K$ for model efficiency, since $\hat \gC$ does not need to match $\gC$ exactly and can be much smaller. Next, this soft weighting mechanism is applicable when an unseen class $\notin \gC$ appears in a realistic testing scenario, while previous class-aware works~\cite{ho2024ltad, lu2023hierarchical} would fail to function.

For this approach to be effective, the concept set $\hat \gC$ should be representative enough to cover the image classes $\gC$ of interest. Instead of manually selecting the set $\hat \gC$, we leverage the zero-shot capability of foundation models where $\hat \gC$ is learned with only the visual information of the training set and {\it without seeing any class labels.}

Given the training dataset $\gD^{\tt T} = \{ \mX_i \}_{i = 1}^N$ and a vocabulary set $\gV = \{ v_j\}_{j = 1}^{|\gV|}$, 
we first measure the pairwise similarity score $S_{ij}$ between $\vf^{\tt f}_i$ and each vocabulary $v_j$, \ie, $S_{ij} = \left< \vf^{\tt f}_i,\: {\tt E}^{\tt T}(v_j) \right>$. 
We then conduct a majority vote to pick the top-$\hat K$ vocabularies with the overall highest similarity scores across all images.
This voting result is used to initialize $\hat \gC$. Specifically, we create embeddings $\vt_{\hat c} = {\tt E}^{\tt T}(\hat c)$ for all $\hat c \in \hat \gC$ and make them learnable so that we can find more representative concept features during training.
 
At test time, a soft label, \ie, concept score $\vp \in [0, 1]^{\hat K}$, is predicted for each image $\mX$.
We compute $\vp$ by measuring its similarity to all concepts 
$\hat c \in \hat \gC$, \ie, 
\bea \vp \: = \: {\tt SoftMax}(\{\left< \vf^{\tt f}, \ \vt_{\hat c} \right> \}_{\hat c \in \hat \gC}).
\eea
As shown in~\figref{fig:teaser}, we incorporate a soft switching mechanism and the final output is a composition of each $\mM_{\hat c}$ weighted by $\vp$, \ie, $\sum_{{\hat c}=1}^{\hat K} p_{\hat c} \mM_{\hat c}$. We will now describe the high-level idea of how to incorporate soft switching into the AD pipeline.

\subsection{Proposed class-agnostic AD pipeline}
Our class-agnostic model is based on the general two-branch pipeline of LTAD~\cite{ho2024ltad} as it is the SOTA method, however, we note that the idea of soft switching can be applied to other class-aware models as well. Leveraging the previously acquired $\vp$, our soft switching mechanism can learn to composite outputs resulting from any multiple concepts or class-specific submodules. 

As shown in Fig.~\ref{fig:pipeline}, we deploy a similar overall pipeline as LTAD but with several key differences: (a) we introduce $\hat C$ as the cornerstone of our class-agnostic framework; (b) we replace the LTAD's class-aware modules with our class-agnostic LTOAD design; (c) for the model architecture in $\tt R$, we use our concept-based vector quantization VAE (Concept VQ); (d) we implement a comprehensive prompt learning. We provide the details of each module in~\secref{sec:details}.

\section{Long-tailed online anomaly detection}
\label{sec:task}

\subsection{Task formulation}
Given a model ${\tt F}_{\theta_0}$ where $\theta_0$ is the parameters trained offline on an LTAD dataset, the goal of \ours is to update the parameters $\theta_t$ in an online manner to improve the performance on a data stream $\tilde \mX_{\leq t} = \left[\tilde \mX_1, \: \cdots, \:\tilde \mX_t\right]$ where each image $\tilde \mX$ comes sequentially. Formally, we focus on improving the accuracy of ${\tt F}_{\theta_t}(\tilde \mX_{\leq t}) = \hat \mY_t$,
where $\hat \mY_t$ is the prediction at $t$. Note that $\tilde \mX_{\leq t}$ is an ordered list,~\ie, \ours is evaluated sequentially.

\myparagraph{Benchmark}
We evaluate \ours by considering the any-$\Delta$ inference setting where the model is updated on small batches of data samples of size $\Delta$. We split the data into two sets: {\bf (a)} an online training set $\gD^{\tt O}=\tilde\mX_{\leq N^{\tt O}}$
where $N^{\tt O}$ denotes the number of online training samples and {\bf (b)} an evaluation set $\gD^{\tt E}$.
Given the training set $\gD^{\tt O}$ and the model parameters $\theta_{t+1} \leftarrow \gA(\theta_{t} ; \: \gD^{\tt O}_t)$ 
is updated based on the $t$-th batch of samples $\gD^{\tt O}_t = \tilde \mX_{(t-1)\Delta:t\Delta}$ using an online learning algorithm $\gA$. The model is then evaluated on $\gD^{\tt E}$ at every step $t$.

Following standard metrics~\cite{ho2024ltad}, we use AUROC to evaluate the performance of detection (Det.) and segmentation (Seg.). We report the average, over $\gC^{\tt head}$, $\gC^{\tt tail}$, and $\gC$, at each intermediate steps $t' \in [1, \: T-1]$ in the data stream. 

\begin{table}[t]
\centering
\setlength{\tabcolsep}{3pt}

\resizebox{\linewidth}{!}{
\begin{tabular}[t]{ l l } 
\specialrule{.15em}{.05em}{.05em}
Config. & Definition 
\\
\cmidrule(lr){1-2}
{\it B} & 
{a {\it blurry} stream where the whole stream is ordered randomly}
\\
{\it B-HF} &
{a {\it blurry} stream where {\color{darkblue} $\gC^{\tt head}$} are more likely to appear first}
\\
{\it B-TF} & 
{a {\it blurry} stream where {\color{darkred} $\gC^{\tt tail}$} are more likely to appear first}
\\
{\it D2-HF} &
a {\it disjoint} stream with 2 sessions where all {\color{darkblue} $\gC^{\tt head}$} are in the first session
\\
{\it D2-TF} &
a {\it disjoint} stream with 2 sessions where all {\color{darkred} $\gC^{\tt tail}$} are in the first session
\\
{\it D5-HF} & 
a {\it disjoint} stream with 5 sessions where {\color{darkblue} $\gC^{\tt head}$} concentrate in the first few sessions
\\
{\it D5-TF} & 
a {\it disjoint} stream with 5 sessions where {\color{darkred} $\gC^{\tt tail}$} concentrate in the first few sessions
\\
{\it D5-M} & 
a {\it disjoint} stream with 5 sessions where {\color{darkblue} $\gC^{\tt head}$} and {\color{darkred} $\gC^{\tt tail}$} are mixed in each session
\\
\specialrule{.15em}{.05em}{.05em}
\end{tabular}
}
\vspace{-0.2cm}
\captionof{table}{
    \textbf{Configurations for $\gD^{\tt O}$.} 
    We define 8 configurations with combinations of different session type $\in \{ {\it blurry}, {\it disjoint} \}$ and ordering type $\in \{ \textit{{\color{darkblue}{head}}-first}, \: \textit{{\color{darkred}{head}}-first}, \: \textit{else} \}$.
}
\label{table:configuration}
\vspace{-0.35cm}
\end{table}

\myparagraph{Dataset setup}
As the data comes in a stream, a model's online learning performance is highly related to the ordering data in $\gD^{\tt O}$. To study the effect, we sequentially split $\gD^{\tt O}$ into \textit{sessions} which corresponds to a sublist of the dataset. In each session, the distribution of the data is similar. 

We consider two settings, {\it disjoint} and {\it blurry}, based on whether the class subset in each session overlaps or not. In a {\it disjoint} setting, the class subset in each session does not overlap with that of any other sessions, but in a {\it blurry} setting, the class subset of each session can overlap. 

Next, data order is also important in the performance due to overfitting and forgetting, \eg, if $\gD^{\tt O}$ has more $\gC^{\tt tail}$ at the beginning, the performance of $\gC^{\tt head}$ might drop first due to overfitting while the performance of $\gC^{\tt tail}$ plateaus at the end of the stream due to forgetting. 
Therefore, we also consider whether $\gC^{\tt head}$ or $\gC^{\tt tail}$ are more likely to appear first in $\gD^{\tt O}$ to study the effect of forgetting and overfitting. Overall, we design 8 settings for $\gD^{\tt O}$ with different session and ordering types combinations, as defined in~\tabref{table:configuration}. 
Please refer to our Appendix~\ref{supp:data_details} for more details.

\subsection{Anomaly adaptive online learning}
We propose an Anomalous Adaptive learning algorithm $\gA^{\tt AA}$ 
to tackle the issues of overfitting and forgetting for LTOAD.
As \ours is an unsupervised one-class classification task, a naive online algorithm $\gA^{\tt N}$ is to directly use the same loss function as offline learning. Note that $\gA^{\tt N}$ is not ideal as it does not leverage the unlabeled abnormal image in the online learning data.

To take advantage of this difference, our anomaly adaptive online learning algorithm $\gA^{\tt AA}$ computes a pseudo abnormal map $\hat \mM = \gT(\hat \mY)$ acquired by thresholding $\hat \mY$ with a function $\gT$ to adjust the behavior of the original losses. The algorithm is summarized in Alg.~\ref{alg:online}
where we denote these adjustments as a mask-dependent loss function $\tilde \gL(\hat \mM)$. As abnormal images in the online training set are unavailable in offline training, we value these samples more in the loss calculation and assign a larger weight $\beta$ to images with higher $\hat y$. 
Finally, we use the exponential moving average (EMA) for more stable updates. Please see the Appendix~~\ref{supp:implement} for details.
\begin{figure}[t]
\vspace{-.5cm}
\begin{minipage}{\linewidth}
\begin{algorithm}[H]
\caption{Anomaly Adaptive ($\gA^{\tt AA}$) Online learning}
\begin{algorithmic}[1]
\renewcommand{\algorithmicrequire}{\textbf{Input:}}
\renewcommand{\algorithmicensure}{\textbf{Output:}}
\REQUIRE offline-trained model with its parameters ${\tt F}_{\theta_0}$, online training data stream $\gD^{\tt O}$, batch size $\Delta$, threshold function $\gT$, mask-dependent loss function $\tilde \gL$, reduction operation $r$, hyper-parameters $\tau, \ \beta, \ \gamma$
\ENSURE $\Theta = \left[\theta_1, \ \cdots \ , \ \theta_T \right]$, where $T = |\gD^{\tt O}| / \Delta$
\STATE Initialize $\Theta = \left[ \ \right]$, $\tilde \theta_0 = \theta_0$
\FOR {$t = 1$ to $T$} 
\STATE Acquire batch $\gD^{\tt O}_t = \left[ \tilde \mX_i \right]_{i={(t-1)\Delta}, \cdots,  t\Delta}$
\FOR {$j = 1$ to $\Delta$}
\STATE Forward pass $\hat \mY_j = {\tt F}_{\tilde \theta_{t-1}}(\gD^{\tt O}_t[j])$
\STATE Compute pseudo abnormal label $\hat \mM_j = \gT(\hat \mY_j)$
\STATE Compute mask-dependent loss $\tilde \gL_j = \tilde \gL(\hat \mM_j)$
\STATE Compute gradient weight $\lambda_j \leftarrow \beta$ if $r(\hat \mY_j) \geq \tau$ else 1
\ENDFOR
\STATE $\tilde \theta_t \leftarrow \tilde \theta_{t-1} + \frac{1}{\Delta} \sum_j \lambda_j \nabla \tilde \gL_j$
\STATE Update with EMA $\theta_t \leftarrow \gamma \theta_{t-1} + (1 - \gamma) \tilde \theta_{t}$
\STATE $\Theta$.append($\theta_t$)
\ENDFOR
\RETURN $\Theta$
\end{algorithmic}
\label{alg:online}
\end{algorithm}
\end{minipage}
\vspace{-.3cm}
\end{figure}
\section{Model details}
\label{sec:details}
We now provide the description for each of the modules as overviewed in~\figref{fig:pipeline}. Additional implementation details are provided in the Appendix~\ref{supp:details}.

\subsection{Prompt learning}
\label{sec:prompt}
Based on the acquired $\hat \gC$, we construct normal and abnormal prompt sets $\gP^{\tt n}$ and $\gP^{\tt a}$ for \ours~to learn the textual semantics. $\gP^{\tt n}$ contains $\hat K$ prompts where each $P^{\tt n}_{\hat c} \in \gP^{\tt n}$ is initialized with the template of ``a normal $\hat c$.'' In contrast, $\gP^{\tt a}$ is initialized by asking a conversational AI~\cite{copilot, chatgpt} ``For $\hat c$, what are the 5 anomalies we are likely to observe?'' for each $\hat c$. 
Each normal prompt feature is defined as $\vn_{\hat c}= {\tt E}^{\tt T}(P^{\tt n}_{\hat c})$ while each abnormal prompt feature is defined as $\va_{\hat c, \: i}= {\tt E}^{\tt T}(P^{\tt a}_{\hat c, \: i})$ for $i \in [1, \: 5]$. In addition to the learnable $\hat \gC$, $\vn_{\hat c}$ and $\va_{\hat c, \: i}$ are also learnable so that \ours~can find more accurate abnormal semantics for each concept. In general, we learn $\vt_{\hat c}$, $\va_{\hat c}$, and multiple $\vn_{\hat c}$ for each concept.

\subsection{Reconstruction module}
\label{sec:rec}

The goal of the reconstruction module ${\tt R}$ is to reconstruct the features $\mF^{\tt i}$ of normal images for extracting cues about whether an image is anomalous. 
Since ${\tt R}$ is trained only on the representation of normal features, it would be unable to accurately reconstruct abnormal features. Therefore, a discrepancy between the input and reconstructed features provides a useful cue that the input is abnormal.
In more detail, given an input feature $\mF^{\tt i}$, ${\tt R}$ reconstructs it to $\mF^{\tt r}$ and output an abnormal map prediction:
\bea 
\hat \mY^{\tt R} = \frac{1}{2}\left(\mathbf{1} - \left< \mF^{\tt i}, \mF^{\tt r} \right>\right)
\eea
by measuring the discrepancy between $\mF^{\tt i}$ and $\mF^{\tt r}$ via cosine similarity. We now introduce the background of HVQ~\cite{lu2023hierarchical} before we describe the module's architecture and concept codebooks used to create $\mF^{\tt r}$.

\myparagraph{Background}
~\citet{lu2023hierarchical} proposes hierarchical vector-quantization (HVQ) to construct multiple expertise modules for each class to solve unified AD. A hard switching mechanism is applied to choose the correspondent modules for each class.
They also introduce a hierarchical architecture to connect high-level and low-level feature maps. Given multiscale feature maps $\{\mZ_l \}_{l = 1,\:\cdots,\:L}$, where $L$ is the number of layers, they build a hierarchical class-specific module ${{\tt Q}}_{l,c}$ for each $c \: \in \: \gC$. During training, each ${\tt Q}_{l,c}$ only sees $\mZ_l$ that comes from the images of class $c$. The output of each ${{\tt Q}}_{l,c}$ is denoted as $\mQ_{l}$. Each $\mQ_{l}$ is conditioned on $\mZ_l$, and the output of the previous module, \ie $\mQ_{l-1}$. Specifically, 
\begin{align}
    \mQ_{l} =
    \begin{cases}
        {{\tt Q}}_{l,c}(\mZ_{l}) ,& \text{if \ } l = 1 \\
        {{\tt Q}}_{l,c}(\mZ_{l} \oplus \mQ_{l-1}) ,& \text{otherwise}. \\
    \end{cases}
\end{align}
Here, $\oplus$ denotes concatenation on the channel dimension. The outputs $\{ \mQ_l \}_{l = 1, \cdots, L}$ are then fed into their proposed hierarchical decoder~\cite{lu2023hierarchical} for reconstruction. 

The class-specific hierarchical modules ${\tt Q}_{l,c}$ are quantization modules~\cite{oord2017vqvae, marimont2021anomaly, gangloff2022leveraging} that learn to quantify visual feature representations. For readability, we remove the subscripts of ${\tt Q}_{l,c}$. 
Each ${\tt Q}$ stores a learnable codebook $\mB$. The codebook contains $M$ codes of $D$ dimension, \ie $\mB \in \sR^{M \times D}$. Given a 2D input feature $\mZ$,  
${{\tt Q}}$ queries its codebook and finds the nearest prototypes $\mQ$. For each pixel location $(h,\: w)$, they find $\mQ[h,\: w] = \vb_{m^*}$, where $m^* \triangleq \argmin_m \| \mZ[h,\: w] - \mB[m] \|^2_2$.

\myparagraph{Module architecture and concept codebooks}
Given an input $\mF^{\tt i}$, we first project it onto a visual-text shared space through a layer-wise projection module ${\tt E}^{\tt p}$~\cite{ho2024ltad}, \ie, $\mF^{\tt p} = {\tt E}^{\tt p}(\mF^{\tt i})$.
For feature reconstruction, we use a VQ-autoencoder dubbed ``Concept-VQ,'' modified from class-specific HVQ~\cite{lu2023hierarchical} for our concept-specific objective. We also use a prompt-conditioned generator ${\tt G}$ for data augmentation to generate pseudo-normal features $\mF^{\tt n}$ and pseudo-abnormal features $\mF^{\tt a}$. These pseudo features provide additional guidance in training $\tt R$.
Finally, we use the soft label $\vp$ as a weighting mechanism for our concept-specific designs in Concept-VQ and ${\tt G}$.

Our Concept-VQ contains an encoder ${\tt E}^{\tt R}$, a decoder ${\tt D}^{\tt R}$, and hierarchical quantization modules between them. For each $l \in \left[1, \: \cdots \: , \: L\right]$ and each $\hat c \in \hat \gC$, we construct a concept-specific quantization module ${\tt Q}_{l, \: \hat c}$.
To capture the text-visual representation of $\gD^{\tt T}$, our concept codebook $\mB_{l, \: \hat c}$ in each ${\tt Q}_{l, \: \hat c}$  
is initialize $\mB_{l, \: \hat c}$ by sampling $M$ codes around its correspondent $\vt_{\hat c}$.

\myparagraph{Prompt-conditioned data augmentation}
During training, prior works~\cite{you2022unified, ho2024ltad, lu2023hierarchical, gao2024onenip} create pseudo-abnormal features for data augmentation by adding noise to the input features, \ie, {\it feature jittering}. These prior works ignore the semantics of the $\mF^{\tt p}$. Instead, we build a generator $\tt G$ that synthesizes pseudo features conditioned on prompt features $\va$ and $\vn$.
We generate pseudo-normal features $\mF^{\tt n}$ conditioned on $\mF^{\tt p}$ and $\vn$ while we generate pseudo-abnormal features $\mF^{\tt a}$ conditioned on $\mF^{\tt p}$ and $\va$.

\subsection{Semantics module}
\label{sec:sem}
The semantics module ${\tt S}$ leverages the cross-modality knowledge of abnormality in ${\tt E}^{\tt I}$ and ${\tt E}^{\tt T}$ to extract cues on whether an image is anomalous. 
It learns the semantic similarities or dissimilarities between visual features and normal texts, \eg, {\it "normal,"} versus abnormal texts, \eg, {\it ``broken''} by comparing the aggregated normal similarity scores $\mS^{\tt n}$ and abnormal scores $\mS^{\tt a}$. The higher $\mS^{\tt a} - \mS^{\tt n}$ is, the more likely it is abnormal. 
Given the projected feature $\mF^{\tt p}$, we denote the similarity difference as a functionality of ${\tt S}$, \ie, ${\tt S}(\mF^{\tt P}) = \mS^{\tt a} - \mS^{\tt n}$.
{
Then, the final prediction $\hat \mY^{\tt S} \triangleq ({\tt S}(\mF^{\tt p}) + \mathbf{2})/4$, where we scale to the range of $[0, \: 1]$. 
}

\myparagraph{Module architecture}
Given $\mF^{\tt p}$, we first compute the similarity map $\mS^{\tt n}_{\hat c}$ between it and the concept-specific normal prompt features, \ie, $\mS^{\tt n}_{\hat c} = \left< \mF^{\tt p}, \: \vn_{\hat c} \right>$. The overall normal similarity map $\mS^{\tt n}$ is then $\sum_{\hat c} p_{\hat c} \mS^{\tt n}_{\hat c}$. Similarly, we also compute the overall abnormal similarity map $\mS^{\tt a}$ for $\mF^{\tt p}$. 

For segmentation, the final prediction $\hat \mY$ is the elementwise harmonic mean~\cite{jeong2023winclip, li2024promptad} of the two intermediate predictions, \ie,
\begin{align}
    \hat \mY = \left( \alpha(\hat \mY^{\tt R})^{-1} + (1-\alpha)(\hat \mY^{\tt S})^{-1} \right)^{-1},
    \label{eq:final_y}
\end{align}
where $\alpha$ controls the weight between them. For detection, we apply an {\tt AvgPool2d} with a kernel size of 16 on $\hat \mY$ before we reduce the map to its maximum value as $\hat y$. We denote this reduction operation as $\hat y = r(\hat \mY)$.

\section{Experiments}
\label{sec:exp}

We first study the offline LTAD setting as an offline model parameter $\theta_0$ is the starting point for an online method. 
Importantly, we show that class-agnostic models are effective in offline LTAD without degrading performance. Finally, we conduct experiments on \ours with our $\gA^{\tt AA}$ in the online setting and evaluate cross-dataset generalizability. Additional experiment details are in the Appendix~\ref{supp:offline_results}.

\begin{table}[t]

        \centering
\setlength{\tabcolsep}{3pt}

\resizebox{\linewidth}{!}{
\begin{tabular}[t]{c@{}c c ccc c ccc c ccc c cccc} 
\specialrule{.15em}{.05em}{.05em}
\multirow{2}{*}{Method} & \multirow{2}{*}{CA} &
\multicolumn{2}{c}{\it exp100} && 
\multicolumn{2}{c}{\it exp200} &&
\multicolumn{2}{c}{\it step100} && 
\multicolumn{2}{c}{\it step200}
\\
& & 
Det. & Seg. &&
Det. & Seg. &&
Det. & Seg. &&
Det. & Seg. 
\\
\cmidrule(lr){1-17}
RegAD~\cite{huang2022registration} & \ccross & 
82.43 & 95.20 && N/A & N/A && 
81.54 & \ul{95.10} && N/A & N/A 
\\
{AnomalyGPT~\cite{gu2024agpt}} & \ccross &
87.44 & 89.68 && 85.80 & 90.15 && 
85.95 & 89.28 && 82.47 & 89.45 
\\
{LTAD~\cite{ho2024ltad}} & \ccross &
{88.86} & 94.46 && 86.05 & 94.18 && 
{87.36} & 93.83 && {85.60} & 92.12
\\
HVQ~\cite{lu2023hierarchical} & \ccross &
87.43 & \bf {95.25} && 85.34 & \ul{94.79} && 
85.39 & 94.17 && 83.00 & \ul{92.61}
\\
\rowcolor{OursColor} {LTOAD*} & \ccross &
\ul{93.12} & 95.01 && \ul{91.88} & 94.36 && \ul{92.02} & 94.72 && \ul{87.95} & 93.34
\\
\cmidrule(lr){1-17}
UniAD~\cite{you2022unified} & \ccheck &
87.70 & 93.95 && {86.21} & 93.26 && 
83.37 & 91.47 && 81.32 & 92.29
\\
MoEAD~\cite{mengmoead} & \ccheck &
84.73 & 94.34 && 83.23 & 93.96 &&
84.40 & 93.76 && 83.13 & 92.76
\\

\rowcolor{OursColor} {\ours} & \ccheck &
\bf 93.42 & \ul{95.21} && \bf 92.02 & \bf 94.94 && 
\bf 92.33 & \bf 95.11 && \bf 88.62 & \bf 94.00
\\
\specialrule{.15em}{.05em}{.05em}
\end{tabular}
}
\vspace{-0.15cm}
\caption{
    \textbf{Comparison ($\uparrow$) on offline MVTec} in image-level AUROC for anomaly detection (Det.) and pixel-level AUROC for anomaly segmentation (Seg.). The column CA (class-agnostic) indicates whether a method requires class names or the number of classes during training or not (require: \ccross; not require: $\ccheck$). RegAD needs at least 2 images per class and thus is not applicable on {\it exp200} and {\it step200}. We mark the best and second best performances in \textbf{bold} and \ul{underline}. 
}
\label{table:mvtec-main-offline}
\vspace{-0.2cm}

\end{table}

\begin{table}[t]

        \centering
\setlength{\tabcolsep}{3pt}

\resizebox{\linewidth}{!}{
\begin{tabular}{c@{}c ccc c ccc c ccc c cccc} 
\specialrule{.15em}{.05em}{.05em}
\multirow{2}{*}{Method} & \multirow{2}{*}{CA} &
\multicolumn{2}{c}{\it exp100} && 
\multicolumn{2}{c}{\it exp200} &&
\multicolumn{2}{c}{\it step100} && 
\multicolumn{2}{c}{\it step200}
\\
& &
Det. & Seg. &&
Det. & Seg. &&
Det. & Seg. &&
Det. & Seg. 
\\
\cmidrule(lr){1-17}
RegAD~\cite{huang2022registration} & \ccross &
71.36 & 94.40 && 71.36 & 94.40 &&
71.80 & 94.99 && 71.65 & 94.52 
\\
{AnomalyGPT~\cite{gu2024agpt}} & \ccross & 
70.34 & 80.32 && 69.78 & 79.48 &&
71.98 & 82.30 && 69.78 & 81.97 
\\
{LTAD~\cite{ho2024ltad}} & \ccross &
{80.00} & 95.56 && {80.21} & \ul{95.36} &&
{84.80} & \ul{96.57} && {84.03} & \ul{96.27}
\\
HVQ~\cite{lu2023hierarchical} & \ccross &
78.63 & \ul{96.18} && 78.03 & 95.32 &&
76.68 & 95.63 && 75.52 & 95.50
\\
\rowcolor{OursColor} {LTOAD*} & \ccross &
\ul{84.89} & 96.42 && \ul{84.84} & 96.44 && \ul{87.71} & 97.07 && \ul{86.52} & 96.89
\\
\cmidrule(lr){1-17}
UniAD~\cite{you2022unified} & \ccheck &
77.31 & 95.03 && 76.87 & 94.80 && 
78.83 & 96.04 && 77.64 & 95.66 
\\
MoEAD~\cite{mengmoead} & \ccheck &
73.89 & 94.34 && 73.58 & 93.98 &&
74.12 & 94.37 && 76.77 & 94.82
\\
\rowcolor{OursColor} {\ours} & \ccheck &
\bf 85.26 & \bf 96.91 && \bf 85.24 & \bf 96.97 && \bf 88.32 & \bf 97.54 && \bf 87.04 & \bf 97.23
\\
\specialrule{.15em}{.05em}{.05em}
\end{tabular}
}
\vspace{-0.15cm}
\caption{
   \textbf{Comparison ($\uparrow$) on offline VisA}. Format follows~\tabref{table:mvtec-main-offline}.
}
\vspace{-0.2cm}
\label{table:visa-main-offline}

\end{table}

\subsection{Offline long-tailed settings}

\myparagraph{Experiment setup} 
Following the LTAD benchmark~\cite{ho2024ltad}, we report on MVTec~\cite{bergmann2019mvtec} and VisA~\cite{zou2022spot}. 
Each dataset has multiple long-tailed settings dependent on the degree of imbalance between $\gC^{\tt head}$ and $\gC^{\tt tail}$. The imbalance type $\in \{ \textit{exp}, \: \textit{step}\}$ describes whether the numbers of classes decay exponentially or drop like a step function; while the imbalance factor $\in \{ \textit{100}, \: \textit{200}\}$ measures the ratio between the number of the most populated class to the least one. A larger imbalance factor means a more challenging setting. We follow exactly the same LTAD training and testing settings unless otherwise specified.
Additional results on DAGM~\cite{wieler2007weakly} and Uni-Medical~\cite{zhang2023exploring, bao2024bmad} are provided in the Appendix~\ref{supp:offline_results}.

\myparagraph{Baselines \& implementation details}
We compare with RegAD~\cite{huang2022registration}, UniAD~\cite{you2022unified}, and LTAD~\cite{ho2024ltad}. We benchmark the SOTA MoEAD~\cite{mengmoead} and HVQ~\cite{lu2023hierarchical} to highlight the gain of our concept-specific quantization modules to their class-specific ones. Only UniAD, MoEAD, and \ours are class-agnostic and do not need a class label at the test time. We provide a class-aware version of our model, LTOAD*. 

\myparagraph{Quantitative \& qualitative comparison}
\tabref{table:mvtec-main-offline} and ~\ref{table:visa-main-offline} show the comparisons of \ours with the baselines on MVTec and VisA.
We observe that \ours outperforms the SOTA on most settings while using no class information. \ours achieves a larger gain over Det. On average across all settings, \ours achieves $+4.63$\% / $+0.58$\% gain versus the second best baseline excluding LTOAD* on MVTec Det. / Seg. As for VisA, \ours achieves $+4.30$\% / $+1.29$\% gain. 

\begin{figure}[t]
    \vspace{-0.2cm}
    \centering
    \includegraphics[width=0.9\linewidth]{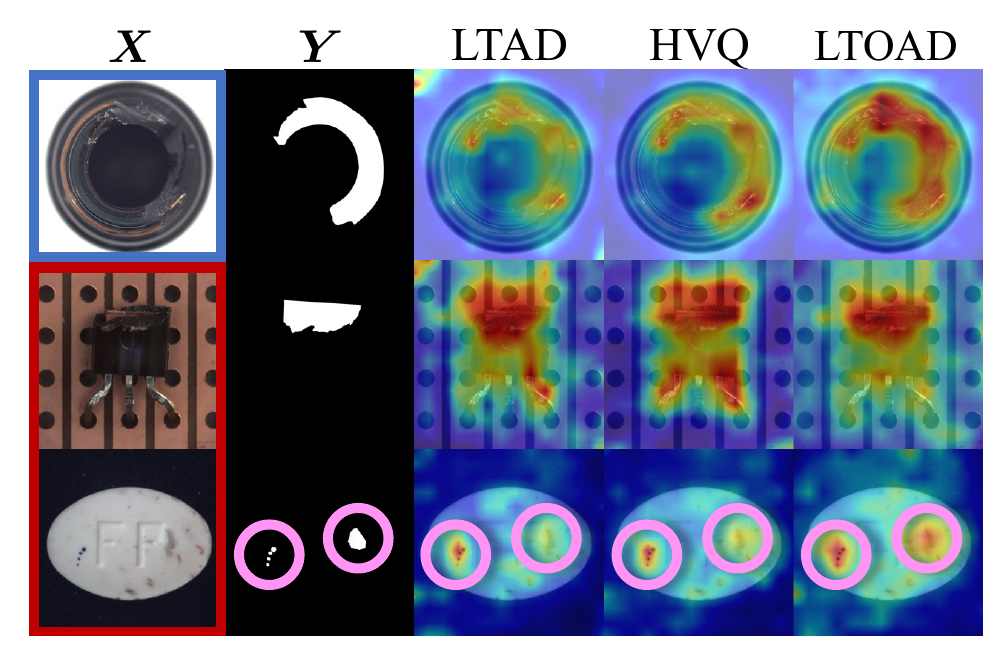}
    \vspace{-0.2cm}
    \captionof{figure}{
        \textbf{Qualitative comparison} among LTAD~\cite{ho2024ltad}, HVQ~\cite{lu2023hierarchical}, and \ours on MVTec offline LTAD~\cite{ho2024ltad} {\it exp100} benchmark. Inputs from {\color{darkblue}{$\gC^{\tt head}$}} / {\color{darkred}{$\gC^{\tt tail}$}} are outlined in {\color{darkblue}{blue}} / {\color{darkred}{red}}. Smaller ground truth anomaly masks are circled in {\color{pink}{pink}}.
    }
    \label{fig:qualitative-comparison}    
\end{figure}
Next, comparing LTAD and LTOAD*, we find that being class-agnostic is beneficial. We hypothesize that this is because the classes can now share knowledge through the soft labels.
Finally, we show qualitative results in~\figref{fig:qualitative-comparison}. We observe that \ours predictions are closer to ground truth than HVQ and LTAD. 

\subsection{Online long-tailed benchmark}

\myparagraph{Experiment setup}
For the online setting, we report on the same datasets as the offline benchmark~\cite{ho2024ltad}, \ie., MVTec, VisA, and DAGM. We split the testing set into $\gD^{\tt O}$ and $\gD^{\tt E}$ with a 3:7 ratio. For each dataset, we create the 8 settings defined in~\tabref{table:configuration}.

For baseline comparisons, we extend HVQ and LTAD to have online capability by applying our naive online learning algorithm $\gA^{\tt N}$ since they are the best offline models on MVTec. Note that both methods are not class-agnostic and needs class labels as extra information in online learning, which is not available/necessary for our approach.

\begin{table}[t]
\centering
\setlength{\tabcolsep}{3pt}

\resizebox{\linewidth}{!}{
\begin{tabular}{@{}c@{\hspace{.3em}}c@{\hspace{.2em}}c@{\hspace{.3em}}c@{\hspace{.4em}}c@{\hspace{.4em}}c@{\hspace{.4em}}c@{\hspace{.4em}}c@{\hspace{.4em}}c@{\hspace{.4em}}c@{\hspace{.4em}}c@{\hspace{.4em}}c@{\hspace{.4em}}c@{}}
\specialrule{.15em}{.05em}{.05em}
{Method} & 
{CA} & 
{Online} &
{\it B} &
{\it B-HF} &
{\it B-TF} &
{\it D2-HF} &
{\it D2-TF} &
{\it D5-HF} &
{\it D5-TF} &
{\it D5-M} &
{Avg.}
\\
\cmidrule(lr){1-12}
LTAD~\cite{ho2024ltad} & \ccross & \ccross &
93.46 & 93.46 & 93.46 & 93.46 & 93.46 & 93.46 & 93.46 & 93.46 & 93.46
\\
LTAD~\cite{ho2024ltad} & \ccross & $\gA^{\tt N}$ &
94.12 & 93.80 & 94.07 & 93.70 & 94.08 & 93.49 & 93.84 & 93.65 & 93.84
\\
{HVQ~\cite{lu2023hierarchical}} & \ccross & \ccross & 
95.02 & 95.02 & 95.02 & 95.02 & 95.02 & 95.02 & 95.02 & 95.02 & 95.02
\\
{HVQ~\cite{lu2023hierarchical}} & \ccross & $\gA^{\tt N}$ & 
\bf 95.44 & \bf 95.33 & 95.28 & 95.22 & \bf 95.29 & 95.03 & 95.21 & 95.05 & 95.23
\\
\cmidrule(lr){1-12}
{\ours} & \ccheck & \ccross &
95.08 & 95.08 & 95.08 & 95.08 & 95.08 & 95.08 & 95.08 & 95.08 & 95.08
\\
{\ours} & \ccheck & $\gA^{\tt N}$ &
95.41 & 95.22 & \bf 95.32 & \bf 95.23 & 95.04 & 94.91 & 94.72 & 94.79 & 95.08
\\
\rowcolor{OursColor} {\ours} & \ccheck & $\gA^{\tt AA}$ &
{95.36} & 95.30 & \bf 95.32 & \bf 95.23 & 95.28 & \bf 95.26 & \bf 95.25 & \bf 95.21 & \bf 95.28
\\
\specialrule{.15em}{.05em}{.05em}
\end{tabular}
}
\caption{
    \textbf{Comparison ($\uparrow$) on online MVTec} in pixel-level AUROC for anomaly segmentation across 8 configurations of $\gD^{\tt O}$. All models are pre-trained on MVTec {\it exp100}~\cite{ho2024ltad}.
}
\vspace{-0.44cm}
\label{table:mvtec-main-online}

\end{table}

\begin{figure}[t]
    \includegraphics[width=\linewidth]{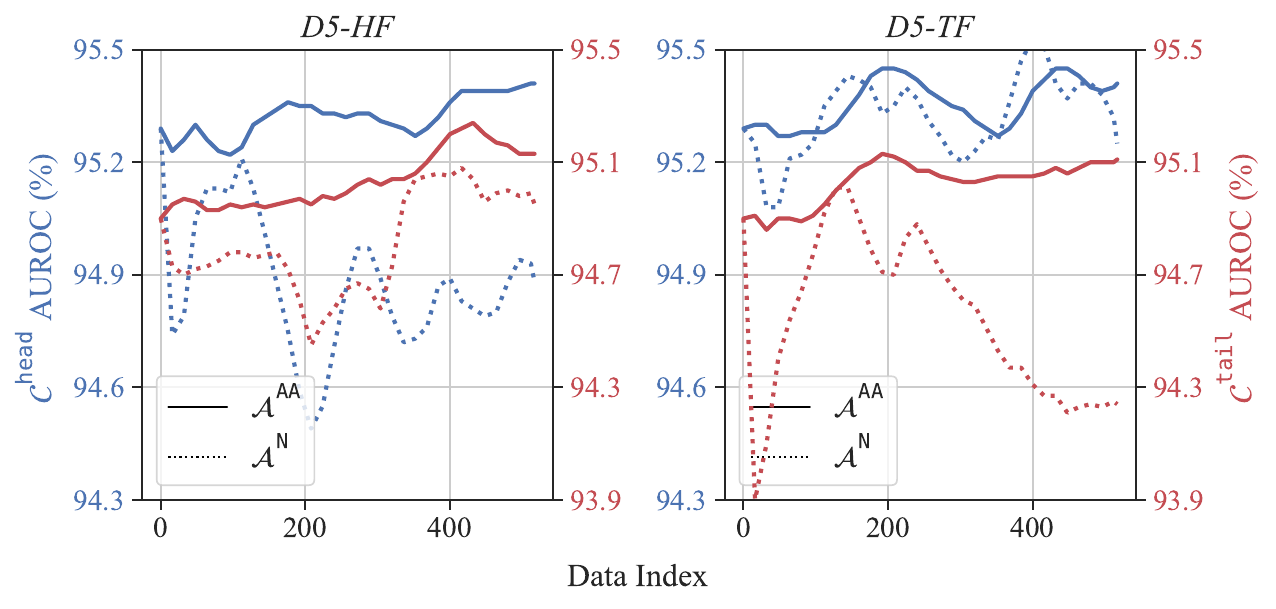}
    \vspace{-0.65cm}
    \caption{
        \textbf{Performance curve} of $\gA^{\tt N}$ and $\gA^{\tt AA}$ in pixel-level AUROC on {\color{darkblue}{$\gC^{\tt head}$}} and {\color{darkred}{$\gC^{\tt tail}$}}. They are offline-trained on {\it exp100}~\cite{ho2024ltad} and online-trained on {\it D5-HF} or {\it D5-TF}.
    }
    \vspace{-0.2cm}
    \label{fig:online-curve}
\end{figure}

\begin{table}[t]
\centering

\setlength{\tabcolsep}{2pt}
\resizebox{\linewidth}{!}{
\begin{tabular}{cccccccccccc} 
\specialrule{.15em}{.05em}{.05em}
\multirow{2.5}{*}{Method} & \multirow{2.5}{*}{Online} &
\multicolumn{2}{c}{MVTec} && 
\multicolumn{2}{c}{VisA} &&
\multicolumn{2}{c}{DAGM}
\\
\cmidrule(l){3-4} \cmidrule(l){6-7} \cmidrule(l){9-10} 
& &
$\rightarrow$ VisA & $\rightarrow$ DAGM &&
$\rightarrow$ MVTec & $\rightarrow$ DAGM &&
$\rightarrow$ MVTec & $\rightarrow$ VisA
\\
\cmidrule(lr){1-10}
{\ours} & \ccross &
84.38 & 91.27 && 82.73 & 79.16 && 79.51 & 81.13
\\
\rowcolor{OursColor} {\ours} & \ccheck &
\bf 92.37 & \bf 96.87 && \bf 89.84 & \bf 96.79 && \bf 88.37 & \bf 92.41
\\
\specialrule{.15em}{.05em}{.05em}
\end{tabular}
}
\vspace{-0.15cm}
\captionof{table}{
    \textbf{Cross-dataset} anomaly segmentation in pixel-level AUROC. All models are pre-trained on their {\it exp100}~\cite{ho2024ltad}.
}

\vspace{-0.05cm}
\label{table:general-exp100-R}

\end{table}

\myparagraph{Quantitative comparison}
We report the online Seg. performance at final time step $t = T$ on MVTec {\it exp100} in~\tabref{table:mvtec-main-online}. The results on other settings/datasets are in the Appendix~\ref{supp:online_results}.
Overall, $\gA^{\tt N}$ performs well in the simplest setting ({\it B}), {\it but  degrades} on more difficult settings, \eg, {\it D2-TF, D5-HF}, \etc. On average, \ours with $\gA^{\tt AA}$ performs the best even without class labels as extra information. We plot \ours's performance curve of ${\tt F}_{\theta_t}$ over $t$ with $\Delta$=16 under the two most difficult settings ({\it D5-HF} and {\it D5-TF}) in~\figref{fig:online-curve}, where $\gA^{\tt N}$ falls off on $\gC^{\tt head}$/$\gC^{\tt head}$ under {\it D5-HF}/{\it D5-TF} while $\gA^{\tt AA}$ improves steadily. Empirically, LTOAD is also more efficient than LTAD because $\hat K < K$.

\begin{table}[t]
    \centering
\setlength{\tabcolsep}{3pt}

\resizebox{\linewidth}{!}{
\begin{tabular}{ cccl } 
\specialrule{.15em}{.05em}{.05em}
Set ID & Domain & Size & Elements
\\
\cmidrule(lr){1-4}
\multirow{3}{*}{
    $\gC$ 
}
& 
\multirow{3}{*}{
    N/A
}
& 
\multirow{3}{*}{
    15
}
&
\textcolor{blue}{\ul{zipper}}, \textcolor{teal}{pill}, \textcolor{teal}{capsule}, grid, \textcolor{purple}{transistor}, 
\\  
&&&
carpet, metal nut, \textcolor{brown}{wood}, \textcolor{brown}{leather},  screw, 
\\
&&&
tile, \textcolor{orange}{cable}, toothbrush, \textcolor{violet}{hazelnut}, \textcolor{red}{bottle*}
\\
\cmidrule(lr){1-4}
\multirow{2}{*}{
    {\it Random}
}
& 
\multirow{2}{*}{
    Out
}
&
\multirow{2}{*}{
    10
}
&
refinery, finger, vault, exam, salad, 
\\ 
&&&
sailor, dove, dealer, well, humanoid 
\\
\cmidrule(lr){1-4}
{\it Small}
& In & 3 &
\textcolor{purple}{semiconductor}, \textcolor{blue}{\ul{zipper}}, \textcolor{brown}{beech}
\\
\cmidrule(lr){1-4}
\rowcolor{OursColor}
&&&
\textcolor{purple}{semiconductor}, \textcolor{blue}{\ul{zipper}}, \textcolor{brown}{beech}, 
\\ 
\rowcolor{OursColor} &&&
\textcolor{red}{microscopy*}, \textcolor{teal}{antibiotics},  \textcolor{teal}{medicines}, 
\\ 
\cellcolor{OursColor}
\multirow{-3}{*}{
    \ours $\hat \gC$
}
&
\cellcolor{OursColor}
\multirow{-3}{*}{
    In
}
&
\cellcolor{OursColor}
\multirow{-3}{*}{
    10
}
&
\cellcolor{OursColor}
\textcolor{orange}{circuit}
, \textcolor{brown}{mahogany}, \textcolor{brown}{hardwood}, \textcolor{violet}{walnut}
\\
\specialrule{.15em}{.05em}{.05em}
\end{tabular}
}
\vspace{-0.15cm}
\captionof{table}{
\textbf{Classes vs. concepts on MVTec.} Words with similar semantics have the same color, \eg, {\it wood}-related words are colored in \textcolor{brown}{brown} and {\it electronics}-related words are colored in \textcolor{purple}{purple}. 
}
\vspace{-0.35cm}
\label{table:mvtec-concept}

\end{table}

\myparagraph{Cross-dataset evaluation}
\tabref{table:general-exp100-R} shows that \ours can adapt to unseen domains even if it is not designed for cross-dataset evaluation. For each $\gD_1 \rightarrow \gD_2$ experiment, we pre-train \ours on {\it exp100} of $\gD_1$ then evaluated on {\it B} of $\gD_2$ without and with online learning. \ours with online learning outperforms the one without it.

\subsection{Additional analysis}
\myparagraph{Analysis on initialization of $\hat \gC$}
We now analyze the effect of our concept learning component. All experiments are done under MVTec {\it exp100} and {\it B}.
We highlight the importance of our concept learning compared to two baselines: initializing with a {\it random} concept set or with a {\it small} one. 

\tabref{table:mvtec-concept} shows the learned initialization of $\hat \gC$ versus the ground truth $\gC$. We observe that our concept learning can find the $\hat \gC$ initialization which represents similar semantics in $\gC$, \eg, the {\it bottle$^*$} images (see the 1st image in Fig.~\ref{fig:qualitative-comparison}) in MVTec are bottom-only views, which look like an {\it microscopy$^*$}.

Next, we experiment with the two baselines to generate initialization for $\hat \gC$. In {\it Random} of~\tabref{table:mvtec-concept}, 
instead of learning from $\gD^{\tt T}$, \ie, in-domain concepts, we randomly select 10 nouns in ALIGN's $\gV$ unrelated to the industrial settings of MVTec, \ie, out-domain concepts. In other words, the semantics of $\gC$ and {\it Random} do not overlap. In {\it Small} of~\tabref{table:mvtec-concept}, instead of choosing $\hat K = 10$, we select only the top-$\hat K=3$ vocabularies as $\hat \gC$. The semantics of {\it Small} partially covers the semantics of $\gC$. \tabref{table:mvtec-ablation-offline} shows the quantitative results.

We report the performance of {\it Random} and {\it Small} in \tabref{table:mvtec-ablation-offline}. Even with an out-domain initialization, compared to HVQ in~\tabref{table:mvtec-main-offline}, which uses $\gC$ for their class-specific quantization modules, {\it Random} still performs competitively. Specifically, we observe a $+$4.11\% gain in Det. versus HVQ. On top of that, with in-domain $\hat \gC$ as a more aligned representation, \ours in~\tabref{table:mvtec-ablation-offline} achieves the best image-level AUROC. In particular, we further improve it by $+$1.88\%.
We have a similar observation initializing with {\it Small}, which outperforms HVQ on Det.  Having sufficiently large $\hat \gC$, \ours with $\hat K = 10$ gives more gain.

{
Finally, although some $c \in \gC$ are not presented in our $\hat \gC$, LTOAD still adapts and outperforms the baselines in \tabref{table:mvtec-ablation-offline}. This suggests that LTOAD can work even with a foundation model whose $\gV$ and $\gC$ are disjoint. However, this is not necessarily true for prior class-aware methods as they implicitly assume that $\gC \subseteq \gV$.
}

\begin{table}[t]
    \centering
    \centering
\setlength{\tabcolsep}{3pt}

\resizebox{\linewidth}{!}{
\begin{tabular}[t]{c@{\hspace{0.3em}}c@{\hspace{0.3em}} c@{\hspace{0.3em}}c@{\hspace{0.3em}}c@{\hspace{0.3em}}c c ccc c ccc c cccc} 
\specialrule{.15em}{.05em}{.05em}
\multirow{2.5}{*}{Exp.} & \multirow{2.5}{*}{$\hat C$} & \multirow{2.5}{*}{$\hat K$} & \multirow{2.5}{*}{${\tt R}$} & \multirow{2.5}{*}{${\tt S}$} & \multirow{2.5}{*}{${\tt G}$} && \multicolumn{3}{c}{Detection} & \multicolumn{3}{c}{Segmentation}
\\
\cmidrule(r){8-10} \cmidrule(r){11-13}
& &&&&& & 
 $\gC^{\tt tail}$ & $\gC^{\tt head}$ & $\gC$ & $\gC^{\tt tail}$ & $\gC^{\tt head}$ & $\gC$
\\
\cmidrule(lr){1-17}
{\it Random} & O-D & 10 & \ccheck & \ccheck & \ccheck &&
84.73 & 99.31 & 91.54 & 94.46 & 95.39 & 94.89
\\
{\it Small} & I-D & 3 & \ccheck & \ccheck & \ccheck &&
85.66 & 98.89 & 91.83 & 94.24 & 95.30 & 94.74
\\
\midrule
{\it NoR} & I-D & 10 & \ccross & \ccheck & \ccross &&
53.84 & 56.25 & 54.97 & 60.73 & 63.46 & 62.00
\\
{\it NoS} & I-D & 10 & \ccheck & \ccross & \ccheck  &&
85.16 & 99.21 & 91.72 & 94.57 & 95.24 & 94.88
\\
{\it NoG} & I-D & 10 & \ccheck & \ccheck & \ccross &&
85.63 & \bf 99.56 & 92.13 & 94.08 & 95.62 & 94.80
\\
\rowcolor{OursColor} \ours & I-D & 10 & \ccheck & \ccheck & \ccheck && 
\bf 88.14 & 99.45 & \bf 93.72 & \bf 94.83 & \bf 95.64 &  \bf 95.36
\\
\specialrule{.15em}{.05em}{.05em}
\end{tabular}
}
\vspace{-0.15cm}
\caption{
    \textbf{Analysis and ablation on \ours} using the same metrics as~\tabref{table:mvtec-main-online} under MVTec \textit{exp100}~\cite{ho2024ltad} and {\it B}. 
    We report the performances on $\gC^{\tt head}$ and $\gC^{\tt tail}$. 
    I-D/O-D: in/out-domain. 
}
\label{table:mvtec-ablation-offline}
\vspace{-.2cm}

\end{table}

\begin{table}[t]
    \centering
    
\centering
\setlength{\tabcolsep}{3pt}
\small
\resizebox{\linewidth}{!}{
\begin{tabular}[t]{c@{\hspace{0.3em}}c@{\hspace{0.2em}} c@{\hspace{0.2em}}c@{\hspace{0.2em}}c@{\hspace{0.3em}}c c ccc c ccc c cccc} 
\specialrule{.15em}{.05em}{.05em}
\multirow{2.5}{*}{Exp.} & {$\hat \gC$} & \multicolumn{3}{c}{$\gP^{\tt a}$} && \multicolumn{3}{c}{Detection} & \multicolumn{3}{c}{Segmentation}
\\
\cmidrule(r){2-2} \cmidrule(r){3-5} \cmidrule(r){7-9} \cmidrule(r){10-12}
& L & L & M & CS &&
 $\gC^{\tt tail}$ & $\gC^{\tt head}$ & $\gC$ & $\gC^{\tt tail}$ & $\gC^{\tt head}$ & $\gC$
\\
\cmidrule(lr){1-17}
{\it FixC} & \ccross & \ccheck & \ccheck & \ccheck &&
84.65 & 98.13 & 90.94 & 94.57 & 95.16 & 94.85
\\
{\it FixP} & \ccheck & \ccross & \ccheck & \ccheck &&
85.59 & \bf 99.45 & 92.06 & 94.52 & 94.75 & 95.04
\\
{\it SingleP} & \ccheck & \ccheck & \ccross & \ccheck &&
85.41 & 99.37 & 91.92 & 94.64 & 95.44 & 95.02
\\
{\it ShareP} & \ccheck & \ccheck & \ccheck & \ccross &&
87.03 & 99.35 & 92.78 & 95.00 & 95.63 & \bf 95.39
\\
\rowcolor{OursColor} 
\ours & \ccheck & \ccheck & \ccheck & \ccheck && 
\bf 88.14 & \bf 99.45 & \bf 93.72 & \bf 94.83 & \bf 95.64 &  95.36
\\
\specialrule{.15em}{.05em}{.05em}
\end{tabular}
}
\vspace{-0.15cm}
\caption{
\textbf{Ablations on concept $\hat \gC$ \& abnormal prompt $\gP^{\tt a}$ learning} under MVTec {\it exp100}~\cite{ho2024ltad} on the same metrics as~\tabref{table:mvtec-ablation-offline}. Acronyms: L: learned; M: multiple; CS: concept-specific.
}
\vspace{-.35cm}
\label{table:mvtec-ablation-prompt}

\end{table}

\myparagraph{Ablation studies on ${\tt R}$, ${\tt S}$, and ${\tt G}$}
In \tabref{table:mvtec-ablation-offline}, we also ablate \ours components (see Sec.~\ref{sec:rec} and \ref{sec:sem}) to justify their design. All experiments are done under the MVTec {\it exp100} and {\it B} setting.
{\it NoR} ablates ${\tt R}$. We also remove ${\tt G}$, $\hat \mY^{\tt R}$ from Eq.~\ref{eq:final_y} and related losses (see Appendix~\ref{supp:training}).
In contrast, {\it NoS} ablates ${\tt S}$. We also remove $\hat \mY^{\tt S}$ and related losses. 
We find from {\it NoR} that ${\tt S}$ alone is not an effective branch while {\it NoS} is far more competitive.
However, when trained jointly (\ours) and compared to {\it NoS},
${\tt S}$ helps ${\tt R}$ improve the Det. accuracy on $\gC^{\tt tail}$ notably. Specifically, ${\tt S}$ improves the image-level AUROC on $\gC^{\tt tail}$ by 2.98\%.

{\it NoG} replaces our prompt-conditioned ${\tt G}$ with unconditional {\it feature jittering}~\cite{you2022unified, ho2024ltad, lu2023hierarchical} for data augmentation. With the guidance of $\gP^{\tt n}$ and $\gP^{\tt a}$, ${\tt G}$ creates high-quality pseudo-features especially beneficial for $\gC^{\tt tail}$. We observe a gain of $+$2.51\% / $+$0.75\% in Det. / Seg. on $\gC^{\tt tail}$.
\begin{table}[t]
    \setlength{\tabcolsep}{6pt}
\resizebox{\columnwidth}{!}{
\begin{tabular}{lc@{\hspace{3mm}}c@{\hspace{3mm}}c@{\hspace{3mm}}c@{\hspace{3mm}}c@{\hspace{3mm}}c@{\hspace{3mm}}c@{\hspace{3mm}}c>{\columncolor{OursColor}}c}
\toprule
\multirow{2.5}{*}{\textbf{\shortstack{Unsupervised AD Method\\Conditions}}} & \multicolumn{9}{c}{\textbf{Unsupervised AD Categories}} \\ 
\cmidrule(l){2-10}
 & $\gG_0$ & $\gG_1$ & $\gG_2$ & $\gG_3$ & $\gG_4$ & $\gG_5$ & $\gG_6$ & $\gG_7$ & \ours \\
\midrule
unified model & \ccross & \ccheck & \ccheck & \ccross & \ccheck & \ccheck & \ccross & \ccheck & \ccheck \\
online learning capability & \ccross & \ccross & \ccross & \ccross & \ccross & \ccross & \ccheck & \ccheck & \ccheck \\
incorporate VLM/LLM(s) & \ccross & \ccross & \ccross & \ccheck & \ccheck & \ccheck & \ccross  & \ccross &\ccheck \\
can handle long-tailed data & \ccross & \ccross & \ccross & \ccross & \ccross & \ccheck & \ccross & \ccross & \ccheck \\
class-agnostic & \ccross & \ccheck & \ccross & \ccross & \ccross & \ccross & \ccross & \ccheck & \ccheck \\
\bottomrule
\end{tabular}
}
\vspace{-0.25cm}
\caption{
\textbf{Characteristic comparison} between \ours and prior unsupervised AD works (\eg, $\gG_0$: \citeGzero, $\gG_1$: \citeGone, $\gG_2$: \cite{lu2023hierarchical}, $\gG_3$: \cite{huang2024adapting, li2024promptad, li2024fade}, $\gG_4$: \cite{cao2023segment, li2024clipsam, cao2024adaclip, jeong2023winclip, chen2024hfp, qu2024vcpclip, zhang2023gpt, gu2024agpt},  $\gG_5$: \cite{ho2024ltad}, $\gG_6$: \cite{wei2024fs, gao2023towards, mcintoshunsupervised}, $\gG_7$: \cite{liu2024unsupervised, tang2024incremental}).
}
\label{tab:related_works}
\vspace{-0.35cm}

\end{table}

\myparagraph{Ablation studies on learnable concepts/prompts}
~\tabref{table:mvtec-ablation-prompt} analyzes the effect of each attribute detailed in~\tabref{tab:related_works_prompt}, where we demonstrate that our prompt learning design is comprehensive. All experiments are performed in the setting MVTec {\it exp100} and {\it B}. {\it FixC} freezes $\hat \gC$. {\it FixP} freezes $\gP^{\tt a}$.
In {\it SingleP}, $\gP^{\tt a}$ stores only one abnormal prompt for each concept. In {\it ShareP}, all concepts share the same abnormal prompts. Each attribute contributes significantly to the Det. performance of $\gC^{\tt tail}$ and $\gC$. 

\section{Related work}\label{sec:rel}

In~\tabref{tab:related_works}, we group the literature of unsupervised AD 
based on their assumptions or target tasks. 
Notably, \ours (1) is a unified model, (2) has online learning capability, (3) uses VLM/LLM priors, (4) handles long-tailed training data distributions, and (5) is class-agnostic. 

\myparagraph{Unsupervised anomaly detection}
$\gG_0$ methods train an independent predictor per class. 
In contrast, UniAD~\cite{you2022unified} challenges to learn one unified predictor for all classes. 
More follow-up research falls into this group ($\gG_1$).
Inspired by prior work~\cite{masoudnia2014mixture, riquelme2021scaling, lu2022heterogeneity}, MoEAD~\cite{mengmoead} uses mixture-of-experts. HVQ~\cite{lu2023hierarchical} uses class-specific quantization modules in their VQ-VAE~\cite{oord2017vqvae} architecture, but HVQ also induces the constraint of requiring class information as inputs, making it not class-agnostic, \ie, $\gG_2$.

\myparagraph{Foundation model based AD}
With advancement in foundation models, recent works focused on zero-shot~\cite{gu2024filo, cao2023segment, AnomalyCLIP, jeong2023winclip, cao2024adaclip, qu2024vcpclip, Li2024MuSc, zhu2024fine} or few-shot~\cite{li2024fade, li2024promptad, zhu2024toward, bai2024dual, liao2024coft, gu2024agpt} unsupervised AD where the normal training images are limited (or none), they~\cite{cao2023segment, AnomalyCLIP, jeong2023winclip, cao2024adaclip, qu2024vcpclip, li2024fade, li2024promptad, gu2024agpt} often turn themselves to the rich knowledge in foundation models~\cite{radford2021learning, jia2021scaling, zhang2023meta, touvron2023llama, han2024onellm} or conversational AI~\cite{chatgpt}. 
These methods can be grouped into either $\gG_3$ or $\gG_4$, depending on whether they are unified model. To query the language models precisely, $\gG_3$ and $\gG_4$ need the class names or labels as extra inputs and are thus not class-agnostic. 

\begin{table}[t]
    \setlength{\tabcolsep}{6pt}
\resizebox{\columnwidth}{!}{
\begin{tabular}{lc@{\hspace{3mm}}c@{\hspace{3mm}}c@{\hspace{3mm}}c@{\hspace{3mm}}c>{\columncolor{OursColor}}c}
\toprule
\multirow{2.5}{*}{\textbf{\shortstack{Prompt Design Conditions of \\Unsupervised AD Methods}}} & \multicolumn{6}{c}{\textbf{Prompt Design Categories in AD}} \\ \cmidrule(l){2-7} 
 & $\gG_8$ & $\gG_{9}$ & $\gG_{10}$ & $\gG_{11}$ & $\gG_{12}$ & \ours \\
\midrule
learnable class/concept names & \ccheck & \ccross & \ccross & \ccross & \ccross & \ccheck \\
support $>$1 APs  & \ccross & \ccross & \ccheck & \ccheck & \ccheck & \ccheck \\
learnable APs & \ccross & \ccheck & \ccross & \ccross & \ccheck & \ccheck \\
class/concept specific APs & \ccross & \ccross & \ccross & \ccheck & \ccheck &  \ccheck \\
\bottomrule
\end{tabular}
}
\vspace{-0.2cm}
\caption{\textbf{Characteristic comparison \wrt prompt design} between \ours and prior unsupervised AD works which use LLM(s) with prompts (\eg, $\gG_8$: \cite{ho2024ltad, qu2024vcpclip}, $\gG_{9}$: \cite{chen2024hfp}, $\gG_{10}$: \cite{jeong2023winclip, li2024fade, zhu2024toward, cao2024adaclip, huang2024adapting}, $\gG_{11}$:~\cite{cao2023segment}, $\gG_{12}$:~\cite{zhu2024fine, li2024promptad}). 
AP: abnormal prompts.}
\label{tab:related_works_prompt}
\vspace{-0.2cm}

\end{table}

Although normal images are limited in zero/few-shot AD, the distribution is even since the number of images per class is the same. Consequently, LTAD~\cite{ho2024ltad}, \ie, $\gG_5$, explores the long-tailed setting where the training set is highly unevenly distributed. They rely on the pre-trained ALIGN~\cite{jia2021scaling} and learn the pseudo-class names.
We highlight the differences between \ours prompt learning to related works in~\tabref{tab:related_works_prompt}. In summary,
\ours supports (a) learnable concept names, (b) learnable abnormal prompts (AP), (c) multiple APs, and (d) concept-specific APs.

\myparagraph{Online learning AD}
Online (or continual/incremental) learning~\cite{koh2021online, wang2024comprehensive, ghunaim2023real, raghavan2024delta} investigates how the model learns continuously on a data stream. 
Recent works~\cite{wei2024fs, gao2023towards, mcintoshunsupervised, liu2024unsupervised, tang2024incremental}, \ie $\gG_6$ and $\gG_7$, explore unsupervised online learning AD.
Differently, we propose the task of long-tailed online
anomaly detection to study the effect of head/tail classes. 

\section{Conclusion}
We propose the benchmark of long-tailed online anomaly detection, studying the online performance of head and tail classes. 
We identify the shortcomings of class-aware methods and propose class-agnostic \ours. It learns to approximate the ground truth class set while using foundation models for our novel concept and abnormal prompt learning framework. We also propose $\gA^{\tt AA}$, a learning algorithm to adapt the online stream with potentially abnormal inputs. \ours outperforms the SOTA in most offline and online settings without additional class information. Although not specifically designed for domain generalization, \ours can adapt to unseen domains. We believe that LTOAD will advance the field toward a more realistic anomaly detection setting that is closer to real-world applications.

\vspace{3pt}
{\noindent\bf Acknowledgments.}
CAY and KCP were supported by Mitsubishi
Electric Research Laboratories. CAY and RAY were in part supported by an NSF Award \#2420724.

{
    \small
    \bibliographystyle{ieeenat_fullname}
    \bibliography{main}
}

\clearpage
\setcounter{page}{1}
\section*{\Large Appendix}

\setcounter{section}{0}
\renewcommand{\theHsection}{A\arabic{section}}
\renewcommand{\thesection}{A\arabic{section}}
\renewcommand{\thetable}{A\arabic{table}}
\setcounter{table}{0}
\setcounter{figure}{0}
\renewcommand{\thetable}{A\arabic{table}}
\renewcommand\thefigure{A\arabic{figure}}
\renewcommand{\theHtable}{A.Tab.\arabic{table}}
\renewcommand{\theHfigure}{A.Abb.\arabic{figure}}
\renewcommand\theequation{A\arabic{equation}}
\renewcommand{\theHequation}{A.Abb.\arabic{equation}}

\noindent The appendix is organized as follows:
\begin{itemize}[topsep=0pt, leftmargin=16pt]
\item In~\secref{supp:details}, we provide additional implementation and training details of our method.
\item In~\secref{supp:data_details}, we provide more details of our new benchmark: long-tailed online AD.
\item In~\secref{supp:offline_results}, we provide additional results on 
Uni-Medical~\cite{zhang2023exploring, bao2024bmad} and LTAD~\cite{ho2024ltad} benchmarks.
\item In~\secref{supp:online_results}, we provide additional results on \ours benchmarks.
\end{itemize}

\section{Details of LTOAD}
\label{supp:details}

\subsection{Training details}
\label{supp:training}

\begin{figure}[h]
    \centering
    \includegraphics[width=0.8\linewidth]{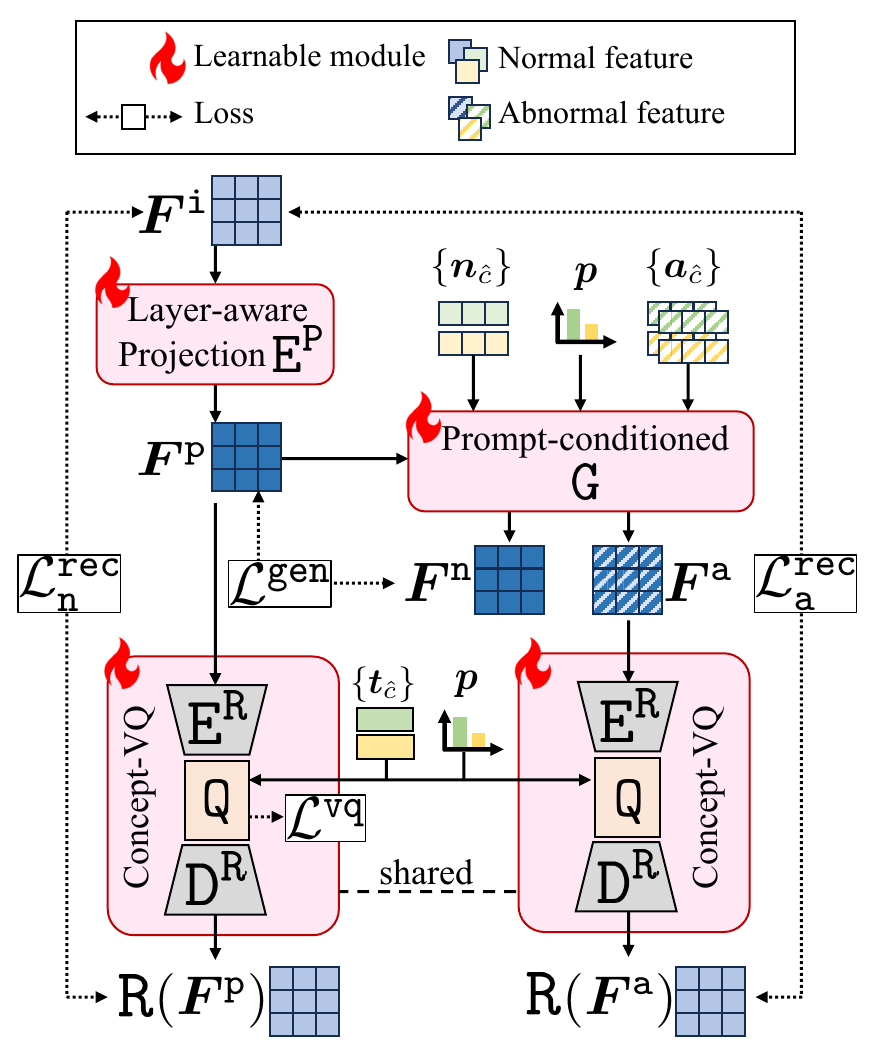}
    \caption{
        \textbf{Reconstruction module ${\tt R}$.} 
        $\:$Our Concept-VQ autoencoder consists of an encoder ${\tt E}^{\tt R}$, a decoder ${\tt D}^{\tt R}$, and quantization modules ${\tt Q}$. It learns to reconstruct input features. As data augmentation, we use ${\tt G}$ to generate pseudo normal $\mF^{\tt n}$ and pseudo abnormal $\mF^{\tt a}$ conditioned on prompts $\{ \vn_{\hat c}\}$ and $\{ \va_{\hat c} \}$, respectively. 
    } 
    \label{fig:rec}
\end{figure}

\myparagraph{Reconstruction module ${\tt R}$}
We detail the training pipeline and details of ${\tt R}$ introduced in Sec.~\ref{sec:rec}.
The pipeline is shown in Fig.~\ref{fig:rec}.

We maximize the similarly of $\mF^{\tt i}$ and reconstructed features $\mF^{\tt r} =  {\tt R}(\mF^{\tt p})$.
At the same time, 
the reconstruction of  $\mF^{\tt a}$ should also be as close to $\mF^{\tt i}$ as possible, \ie, our reconstruction loss $\gL^{\tt rec}$ is defined as:
\fontsize{9.5pt}{9.5pt}
\begin{align}
    \gL^{\tt rec} = \underbrace{
        {\tt GAP} \left( \mathbf{1} - \left< \mF^{\tt i},\ {\tt R}(\mF^{\tt p}) \right> \right)
    }_{\gL^{\tt rec}_{\tt n}} \ + \ \underbrace{
        {\tt GAP} \left( \mathbf{1} - \left< \mF^{\tt i},\ {\tt R}(\mF^{\tt a}) \right>  \right)
    }_{\gL^{\tt rec}_{\tt a}}.
    \label{eq:loss_rec}
\end{align}

In addition to $\gL^{\tt rec}_{\tt a}$ in~\equref{eq:loss_rec} which trains ${\tt G}$, we also push the similarities between pseudo-normal features and the input normal features 
using a generator loss $\gL^{\tt gen} = {\tt GAP} \left(\mathbf{1} - \left< \mF^{\tt p},\ \mF^{\tt n} \right> \right).$

\begin{figure}[h]
    \centering
    \includegraphics[width=0.75\linewidth]{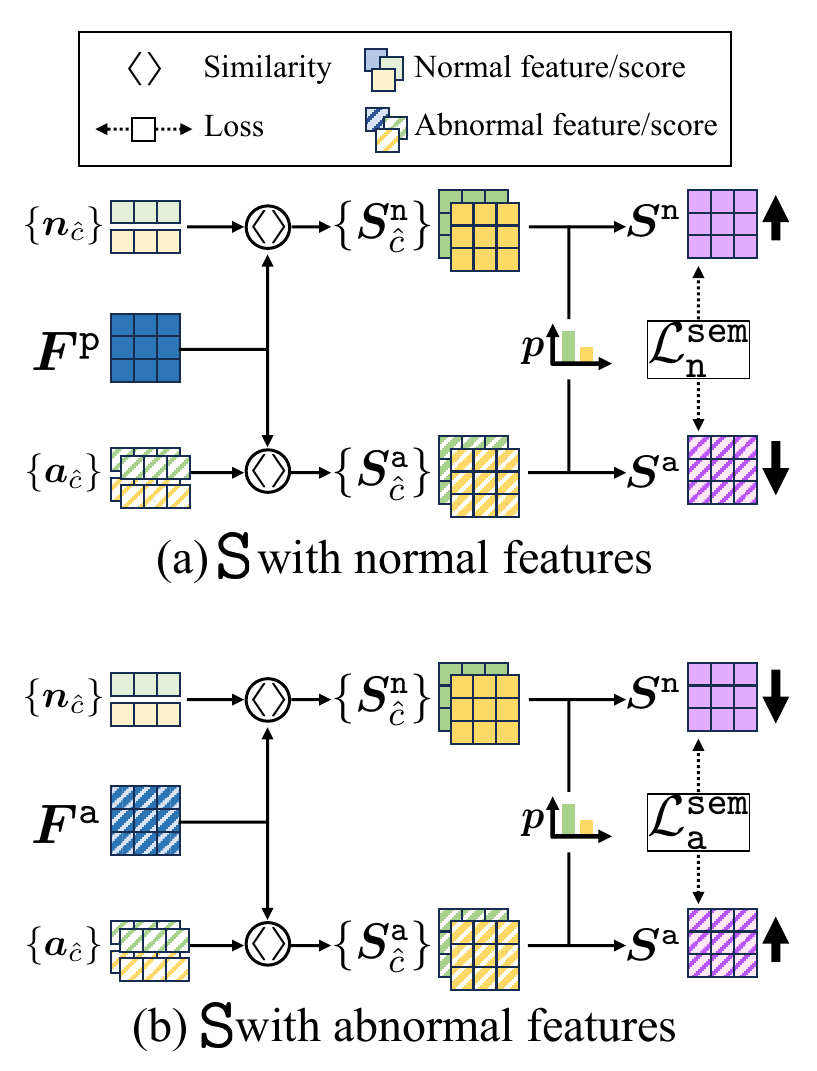}
    \caption{
        \textbf{Semantics module $\tt S$.}
        $\:$(a) Given normal features $\mF^{\tt p}$, $\gL^{\tt sem}_{\tt n}$ maximizes $\mS^{\tt n} - \mS^{\tt a}$. Here $\mS^{\tt n}$ is the overall per-pixel similarity between $\mF^{\tt p}$ and $\{ \vn_{\hat c} \}$ while $\mS^{\tt a}$ is the one between $\mF^{\tt p}$ and $\{ \va_{\hat c} \}$.~(b) On the other hand, given abnormal features $\mF^{\tt a}$, $\gL^{\tt sem}_{\tt a}$ maximizes $\mS^{\tt a} - \mS^{\tt n}$.
    }
    \label{fig:sem}
\end{figure}

\myparagraph{Semantics module ${\tt S}$}
We detail the training pipeline and details of ${\tt S}$ introduced in Sec.~\ref{sec:sem}.
The pipeline is shown in Fig.~\ref{fig:sem}.

We maximize the difference between normal semantic scores $\mS^{\tt n}$ and abnormal semantic scores $\mS^{\tt a}$ for normal $\mF^{\tt p}$ and vice versa for pseudo-abnormal $\mF^{\tt a}$.
We modify the explicit anomaly margin~\cite{li2024promptad} loss for ${\tt S}$. Specifically, the loss is defined as $\gL^{\tt sem}_{\tt n} = \max(-\delta,\: {\tt S}(\mF^{\tt p}))$, where $\max(\cdot)$ is a clipping function and $\delta \in \sR^+$ is a margin term we introduce to further encourage the gap. In contrast, pseudo-abnormal $\mF^{\tt a}$ is trained by minimizing $\gL^{\tt sem}_{\tt a} = \max(-\delta,\: -{\tt S}(\mF^{\tt a}))$ as shown in Fig.~\ref{fig:sem}(b). Together, our semantics loss $\gL^{\tt sem}$ is defined as the sum of the aforesaid two terms, \ie, $\gL^{\tt sem} = \gL^{\tt sem}_{\tt n} + \gL^{\tt sem}_{\tt a}$.
The final loss function $\gL$ for our offline learning is 
\begin{align}
    \gL = \lambda_{\tt vq} \gL^{\tt vq} + \lambda_{\tt rec} \gL^{\tt rec} + \lambda_{\tt gen} \gL^{\tt gen} + \lambda_{\tt sem} \gL^{\tt sem},
    \label{eq:loss}
\end{align} where the $\lambda$s control the weight of each individual loss.
\subsection{Concept codebooks}
\label{supp:codebook}

As mentioned in \secref{sec:rec}, we initialized $\mB_{l, \: \hat c}$ by
sampling $M$ codes around its correspondent $\vt_{\hat c}$.
Specifically, we initialize  $\mB_{l, \: \hat c} = \mN_{l, \: \hat c} + \vt_{\hat c}$ where $\mN_{l, \: \hat c} \in \sR^{M \times D}$ is sampled from a Gaussian distribution $\gN(\mathbf{0}, \: \sigma)$. 
Throughout the training, the center of $\mB_{l, \: \hat c}$ is always set to $\vt_{\hat c}$ and we learn the deviation $\mN_{l, \: \hat c}$ to fit the data representation. We set $\alpha=0.02$.

For each pixel location $(h,\: w)$, we find 
$\mQ_{l, \: \hat c}[h, \: w] = \mB_{l, \: \hat c}[m^*]$
where $m^* = \arg_m \min \left< \mZ_l[h, \: w], \ \mB_{l, \: \hat c}[m] \right>$.
We aggregate $\gL^{\tt vq}_{l, \: \hat c}$ to have $\gL^{\tt vq} = \sum_{l, \: \hat c} p_{\hat c} \gL^{\tt vq}_{l, \: \hat c}$, where $p_{\hat c}$ is the element of $\vp$ associated with $\hat c$.
The final outputs of our quantization modules are the weighted sum from each concept, \ie, $\sum_{\hat c} p_{\hat c} \mQ_{l, \: \hat c}$ for each $l \in [1, \: L]$.

\subsection{Prompt-conditioned data augmentation}
\label{supp:aug}

We detail the architecture of ${\tt G}$ introduced in Sec.~\ref{sec:rec}.
Given the input visual feature map $\mF^{\tt p} \in \sR^{h \times w \times d^{\tt f}}$, where $d^{\tt f}$ is the output dimension of text encoder ${\tt E}^{\tt T}$, and a set of prompt features, ${\tt G}$ generate pseudo-normal $\mF^{\tt n}$ or pseudo-abnormal feature map $\mF^{\tt a} \in \sR^{h \times w \times d^{\tt f}}$ depending on whether the prompt features are $\vn$ or $\va$.

The normal prompt features $\vn \in \sR^{K \times d^{\tt f}}$ and abnormal prompt features $\va \in \sR^{K \times 5 \times d^{\tt f}}$ contains multiple features from all concepts. We randomly select one feature as the input feature for text-condition, denoted as $\vg$.

We first pixel-wise concatenate $\mF^{\tt p}$ and $\vg$ . We feed it to the 5-layer CNN with a hidden dimension of $d^{\tt h}$. We use ReLU as activation layers. We also insert batch normalization layers in between. 

\subsection{Semantics similarity scores}
We detail how we acquire the overall abnormal similarity map $\mS^{\tt n}$ in Sec.~\ref{sec:sem}.
During training, we pick the ``hardest'' abnormal prompts to enhance the gap between normal and abnormal semantics. 
Specifically, for $\mF^{\tt p}$, the hardest abnormal prompt is the one with the highest similarity score since it will most likely confuse the model. In other words, we pick  $\mS^{\tt a}_{\hat c} = {\tt maximum} (\{ \left< \mF^{\tt p}, \: \va_{\hat c, \: i} \right> \}_{i=1,\: \cdots, \: 5})$. Here ${\tt maximum}(\cdot)$ picks the largest value on the $i$-dimension. On the other hand, for $\mF^{\tt a}$, the hardest abnormal samples are the ones with the lowest similarity score, \ie, $\mS^{\tt a}_{\hat c} = {\tt minimum} (\{ \left< \mF^{\tt a}, \: \va_{\hat c, i} \right> \}_{i=1,\: \cdots, \: 5})$. Here ${\tt minimum}(\cdot)$ picks the smallest value on the $i$-dimension.
Finally, the overall abnormal similarity map is $\mS^{\tt a} = \sum_{\hat c} p_{\hat c} \mS^{\tt a}_{\hat c}$. 

During inference, we pick the most probable $\mS^{\tt a}_{\hat c}$ , \ie $\mS^{\tt a}_{\hat c} = {\tt maximum} (\{ \left< \mF^{\tt p}, \: \va_{\hat c, \: i} \right> \}_{i=1,\: \cdots, \: 5})$.

\subsection{Implementation details}
\label{supp:implement}
For our foundation model ${\tt E}^{\tt I}$ and ${\tt E}^{\tt T}$, we use the ALIGN~\cite{jia2021scaling} implementation from Hugging Face Transformer~\cite{huggingface-align}, which is open source and publicly available (Apache 2.0).
Their ${\tt E}^{\tt I}$ is based on ${\tt efficientnet\_b7}$~\cite{tan2019efficientnet} containing multiple EfficientNet blocks.
We use the same setting in ~\cite{ho2024ltad} and define our $\mF^{\tt i}$ as the concatenation of output feature maps from the following blocks: the $3^{\tt rd}$, $10^{\tt th}$, $17^{\tt th}$, and $37^{\tt th}$.

We sample $M = 16$ codes of dimension 640 for each codebook $\mB_{l, \: \hat c}$ while setting the number of codebooks $\hat K = 10$. As a comparison, HVQ~\cite{lu2023hierarchical} samples 512 codes of dimension 256 for their $K$ codebooks. Overall, we require far fewer codes than since $16 \times 10 \times 640 \leq 512 K \times 256$ for all $K \geq 1$.

For $\hat \mY$ (see Eq.~\ref{eq:final_y}), we set $\alpha = 0.3$.
In $\gA^{\tt AA}$~(see Alg.~\ref{alg:online}), we set 
$\gamma = 0.3$, $\beta = 5$, $\tau = 0.2$, and $\gT(\hat \mY) = 0.95 \cdot r(\hat \mY)$. For online learning on the same dataset, we clip the gradient norm by $1\mathrm{e}{-3}$. For offline learning and online learning across different datasets, we clip the gradient norm by $1\mathrm{e}{-1}$.

For all experiments, we use AdamW optimizer with a learning rate of $1\mathrm{e}{-4}$. During training, we use the balance sampler for sampling long-tailed data distribution in LTAD~\cite{ho2024ltad}.
All experiments are conducted on an NVIDIA RTX 6000 GPU. 

\subsection{Concepts and prompts initialization}
\label{supp:concepts_prompts}
We now provide the initialization of the $\hat \gC$ and $\gP^{\tt a}$ we used for all datasets.

\myparagraph{MVTec} Our $\hat \gC$ contains the following vocabularies:
{\it semiconductor}, 
{\it zipper}, 
{\it beech}, 
{\it walnut}, 
{\it circuit}, 
{\it microscopy}, 
{\it mahogany}, 
{\it hardwood}, 
{\it medicines}, 
and {\it antibiotics}.
For each $\hat c \in \hat \gC$, we acquire the 5 abnormal prompts in $\gP^{\tt a}_{\hat c}$ by asking Copilot~\cite{copilot} the following query: 
``{\it Out of 100 $[\hat c]$ which looks generally the same, one appears to have a broken region. List 5 examples.}''

\myparagraph{VisA}
Our $\hat \gC$ contains the following vocabularies:{\it circuit}, 
{\it electronics}, 
{\it candles}, 
{\it bananas}, 
{\it supplements}, 
{\it semiconductor}, 
{\it sensors}, 
{\it vitamin}, 
{\it usb}, 
and {\it wheel}.
For each $\hat c \in \hat \gC$, we acquire the 5 abnormal prompts in $\gP^{\tt a}_{\hat c}$ by asking Copilot~\cite{copilot} the following query: 
``{\it Out of 100 $[\hat c]$ which looks generally the same, one appears to have a broken region. List 5 examples.}''

\myparagraph{DAGM} Our $\hat \gC$ contains the following vocabularies:
{\it ultrasound}, 
{\it fetal}, 
{\it wavelengths}, 
{\it topography}, 
{\it microscopy}, 
{\it imaging}, 
{\it irregularities}, 
{\it electromagnetic}, 
{\it noise}, 
and {\it seismic}.
For each $\hat c \in \hat \gC$, we acquire the 5 abnormal prompts in $\gP^{\tt a}_{\hat c}$ by asking Copilot~\cite{copilot} the following query: 
``{\it Out of 100 $[\hat c]$ which looks generally the same, one appears to have a broken region. List 5 examples.}''

\myparagraph{Uni-Medical}
Our $\hat \gC$ contains the following vocabularies: {\it neuroscience}, {\it brain},
 {\it vascular},
 {\it microscopy},
 {\it imaging},
 {\it mri},
 {\it mitochondrial},
 {\it neurons},
 {\it ultrasound},
 and {\it cerebral}.
For each $\hat c \in \hat \gC$, we acquire the 5 abnormal prompts in $\gP^{\tt a}_{\hat c}$ by asking ChatGPT~\cite{chatgpt} the following query: 
``{\it Out of 100 $[\hat c]$ images which look generally the same, one appears to have a broken region. List 5 examples.}''

\section{Long-tailed online AD benchmark}
\label{supp:data_details}

We re-address the design motivation of our long-tailed online AD (\ours) benchmark. Then we document the additional details for setting up the \ours for the following datasets, \ie, MVTec~\cite{bergmann2019mvtec}, VisA~\cite{zou2022spot}, and DAGM~\cite{wieler2007weakly}.

Our benchmark focuses on evaluating the performance of $\tF_{\theta_t}$ for each step $t$ in online training stream $\gD^{\tt O}$ where ${\theta_0}$ is pre-trained on an long-tailed $\gD^{\tt T}$, \ie, LTAD~\cite{ho2024ltad}.
We design $\gD^{\tt O}$ not to be long-tailed because, in the industrial environment, we need quick adaptation in the online learning process. If $\gD^{\tt O}$ is also long-tailed, it is difficult for the model to learn anomaly detection effectively due to insufficient anomaly samples.
We will provide all of the used meta files of $\gD^{\tt T}$, $\gD^{\tt O}$, and $\gD^{\tt E}$ along with our code once accepted.

\begin{table}[h]

\vspace{1.5em}
\centering
\setlength{\tabcolsep}{3pt}

\resizebox{\linewidth}{!}{%
\begin{tabular}{ ccl } 
\specialrule{.15em}{.05em}{.05em}
Dataset & $|\gC|$ & Elements
\\
\cmidrule(lr){1-3}
\multirow{2}{*}{MVTec}
& 
\multirow{2}{*}{15}
&
\textcolor{darkblue}{hazelnut}, 
\textcolor{darkblue}{leather}, 
\textcolor{darkblue}{bottle}, 
\textcolor{darkblue}{wood}, 
\textcolor{darkblue}{carpet}, 
\textcolor{darkblue}{tile}, 
\textcolor{darkblue}{metal nut},
\\
&&
\textcolor{darkred}{toothbrush}, \textcolor{darkred}{zipper}, \textcolor{darkred}{transistor}, \textcolor{darkred}{grid}, \textcolor{darkred}{pill}, \textcolor{darkred}{capsule},  \textcolor{darkred}{cable}, \textcolor{darkred}{screw}
\\  
\cmidrule{1-3}
\multirow{2}{*}{VisA}
& 
\multirow{2}{*}{12}
&
\textcolor{darkblue}{pcb3}, 
\textcolor{darkblue}{pcb2}, 
\textcolor{darkblue}{pcb1}, 
\textcolor{darkblue}{pcb4}, 
\textcolor{darkblue}{macaroni1}, 
\textcolor{darkblue}{macaroni2},
\\
&&
\textcolor{darkred}{candle}, 
\textcolor{darkred}{cashew}, 
\textcolor{darkred}{fryum}, 
\textcolor{darkred}{capsules}, 
\textcolor{darkred}{chewinggum},  
\textcolor{darkred}{pipe fryum}, 
\textcolor{darkred}{screw}
\\
\cmidrule{1-3}
\multirow{2}{*}{DAGM}
& 
\multirow{2}{*}{10}
&
\textcolor{darkblue}{Class10}, 
\textcolor{darkblue}{Class7}, 
\textcolor{darkblue}{Class9}, 
\textcolor{darkblue}{Class8}, 
\textcolor{darkblue}{Class2}, 
\\
&&
\textcolor{darkred}{Class3}, 
\textcolor{darkred}{Class5}, 
\textcolor{darkred}{Class1}, 
\textcolor{darkred}{Class4}, 
\textcolor{darkred}{Class6}
\\
\specialrule{.15em}{.05em}{.05em}
\end{tabular}
}
\captionof{table}{
\textbf{\textcolor{darkblue}{$\gC^{\tt head}$} and \textcolor{darkred}{$\gC^{\tt tail}$} of MVTec, VisA, and DAGM.}
}
\vspace{-0.02cm}
\label{table:head-tail}

\end{table}

\begin{figure}[h]
    \centering
    \includegraphics[width=\linewidth]{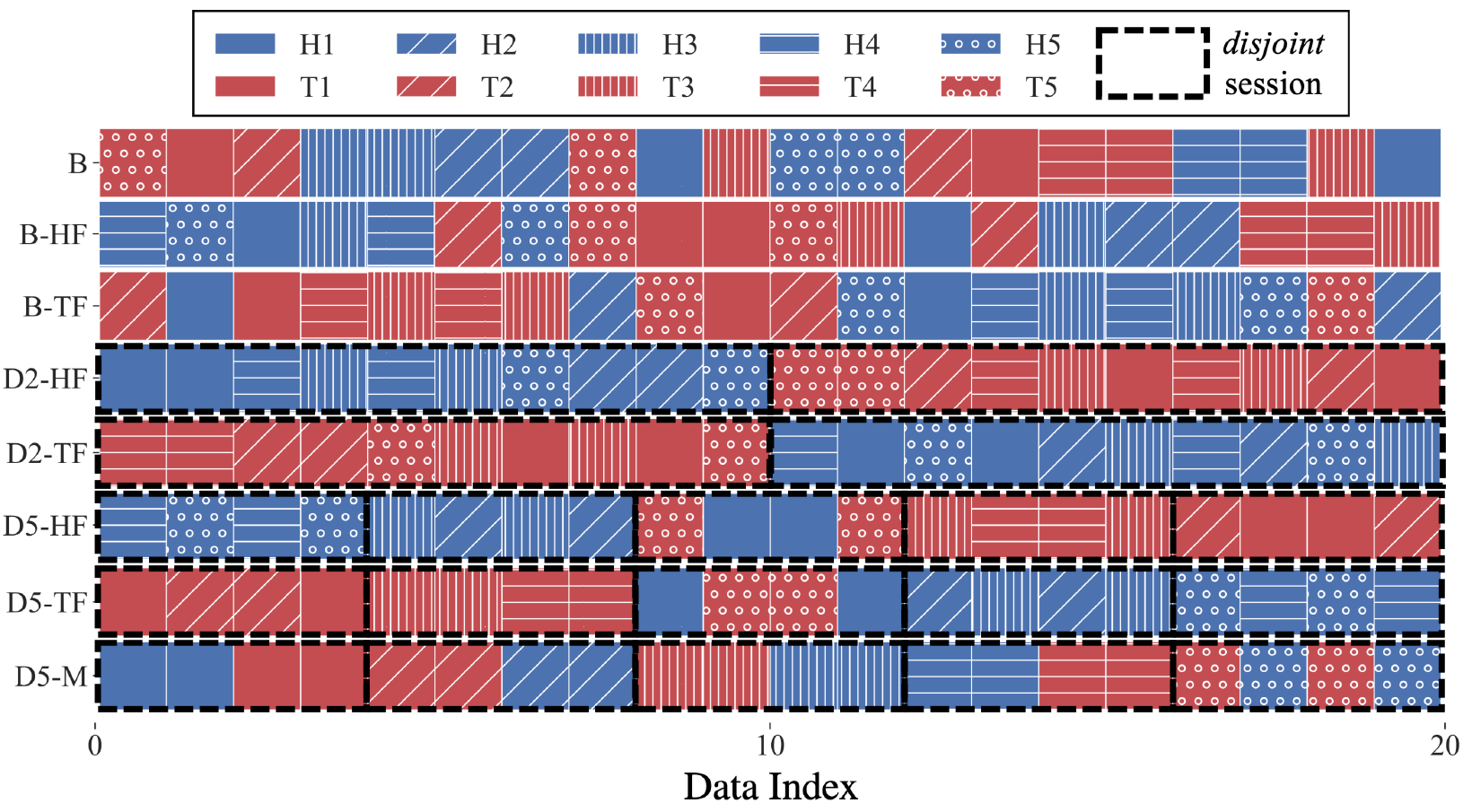}
    \caption{
        \textbf{The configurations for $\gD^{\tt O}$.} 
        We define the 8 configurations with combinations of different session type $\in \{ {\it blurry}, {\it disjoint} \}$ and ordering type $\in \{ \textit{{\color{darkblue}{head}}-first}, \: \textit{{\color{darkred}{head}}-first}, \: \textit{else} \}$. We visualize them (see Tab.~\ref{table:configuration}) with a toy example on $\gD^{\tt O}$. {\color{darkblue} $\gC^{\tt head}$} and {\color{darkred} $\gC^{\tt tail}$ } each has 5 classes. 
    }
    \label{fig:toy}
\end{figure}
In~\figref{fig:toy}, we visualize the proposed 8 configurations in Tab.~\ref{table:configuration} with a toy example for better understanding.

\subsection{MVTec}
MVTec is an industrial anomaly detection dataset, containing 15 classes. In the $1^{\tt st}$ row of ~\tabref{table:head-tail}, we list out all the classes and whether each of them belongs to $\gC^{\tt head}$ or $\gC^{\tt tail}$ in LTAD~\cite{ho2024ltad} benchmark.
We first detail the selected classes of each {\it D5} configuration in \ours. We then discuss the random function we used to sample {\it B-HF} and {\it B-TF} configurations.

In {\it D5-HF}, we divide the $\gD^{\tt O}$ into 5 disjoint sessions and we put the \textcolor{darkblue}{$\gC^{\tt head}$} in the first few sessions. We organize the classes in each session as the following list so that the number of images in each session of $\gD^{\tt O}$ is roughly the same.
\begin{enumerate}
    \item \textcolor{darkblue}{carpet}, \textcolor{darkblue}{hazelnut}, \textcolor{darkblue}{metal nut}
    
    \item \textcolor{darkblue}{tile}, \textcolor{darkblue}{leather}, \textcolor{darkblue}{wood}
    
    \item \textcolor{darkblue}{bottle}, \textcolor{darkred}{screw}, \textcolor{darkred}{capsule}
    
    \item \textcolor{darkred}{zipper},             \textcolor{darkred}{cable}, \textcolor{darkred}{toothbrush} 

    \item \textcolor{darkred}{pill}, \textcolor{darkred}{transistor}, \textcolor{darkred}{grid}
\end{enumerate}

On the other hand, in {\it D5-TF}, we divide the $\gD^{\tt O}$ into 5 disjoint sessions and we put the \textcolor{darkred}{$\gC^{\tt tail}$} in the first few sessions. We organize the classes in each session as the following list, \ie {\it D5-TF} in reverse order.
\begin{enumerate}
    \item \textcolor{darkred}{pill}, \textcolor{darkred}{transistor}, \textcolor{darkred}{grid}

    \item \textcolor{darkred}{zipper},             \textcolor{darkred}{cable}, \textcolor{darkred}{toothbrush}
        
    \item \textcolor{darkred}{screw}, \textcolor{darkred}{capsule}, \textcolor{darkblue}{bottle}

    \item \textcolor{darkblue}{tile}, \textcolor{darkblue}{leather}, \textcolor{darkblue}{wood}

    \item \textcolor{darkblue}{carpet}, \textcolor{darkblue}{hazelnut}, \textcolor{darkblue}{metal nut}
\end{enumerate}

In {\it D5-M}, we divide the $\gD^{\tt O}$ into 5 disjoint sessions and we put both \textcolor{darkblue}{$\gC^{\tt head}$} and \textcolor{darkred}{$\gC^{\tt tail}$} in each session. We organize the classes  in each session as the following list
\begin{enumerate}
    \item 
    \textcolor{darkblue}{carpet}, ~\textcolor{darkred}{cable}, ~\textcolor{darkred}{grid}

    \item 
    \textcolor{darkblue}{hazelnut},
    ~\textcolor{darkred}{transistor},  ~\textcolor{darkred}{capsule}, 

    \item 
    \textcolor{darkblue}{metal nut}, 
    ~\textcolor{darkblue}{wood}, 
    ~\textcolor{darkred}{screw} 

    \item 
    \textcolor{darkblue}{leather},  ~\textcolor{darkred}{pill}, ~\textcolor{darkred}{toothbrush}

    \item 
    \textcolor{darkblue}{bottle}, ~\textcolor{darkblue}{tile}, ~\textcolor{darkred}{zipper}   
\end{enumerate}

In {\it B-HF}, for each $i \in 1,\cdots,N^{\tt O}$ where $N^{\tt O} = |\gD^{\tt O}|$, we use the following operation to determine where $\tilde \mX_i$ should be sampled from $\gC^{\tt head}$ or $\gC^{\tt tail}$.
Let the number of remaining $\gC^{\tt head}$ images at step $t$, \ie $\gD^{\tt O}[t:]$ be $N^{\tt head}_t$ and the one for 
$\gC^{\tt tail}$ be $N^{\tt tail}_t$. We sample from $\gU(0, 1)$ repeatedly for $r$ times and keep the largest number as $u$.
If $u \geq \frac{N^{\tt head}_t}{N^{\tt head}_t + N^{\tt tail}_t}$, then we sample $\tilde \mX_i$ from $\gC^{\tt head}$ and otherwise. We set $r = 5$.

In {\it B-TF}, we use a similar operation. However, we sample from $\gU(0, 1)$ repeatedly for $r$ times and keep the smallest number as $u$ instead.
If $u \leq \frac{N^{\tt head}_t}{N^{\tt head}_t + N^{\tt tail}_t}$, then we sample $\tilde \mX_i$ from $\gC^{\tt tail}$ and otherwise.

\subsection{VisA}
VisA is an industrial anomaly detection dataset, containing 12 classes. We list out the $\gC$ in the $2^{\tt nd}$ row of ~\tabref{table:head-tail}.

Here we use the same design logic for the configurations in MVTec. In {\it D5-HF}, 
we organize the classes in each session as the following list.
\begin{enumerate}
    \item 
    \textcolor{darkblue}{pcb3}, 
    ~\textcolor{darkblue}{pcb4}
    
    \item 
    \textcolor{darkblue}{pcb1}, 
    ~\textcolor{darkblue}{pcb2}
    
    \item 
    \textcolor{darkblue}{macaroni1}, 
    ~\textcolor{darkblue}{macaroni2}
    
    \item 
    \textcolor{darkred}{chewinggum},             
    ~\textcolor{darkred}{fryum}, 
    ~\textcolor{darkred}{pipe fryum} 

    \item 
    \textcolor{darkred}{candle}, 
    ~\textcolor{darkred}{capsules}, 
    ~\textcolor{darkred}{cashew}
\end{enumerate}
In {\it D5-TF}, 
we organize the classes in each session as the following list.
\begin{enumerate}
    \item 
    \textcolor{darkred}{candle}, 
    ~\textcolor{darkred}{capsules}, 
    ~\textcolor{darkred}{cashew}

    \item 
    \textcolor{darkred}{chewinggum},             
    ~\textcolor{darkred}{fryum},  
    ~\textcolor{darkred}{pipe fryum} 

    \item 
    \textcolor{darkblue}{macaroni1}, ~\textcolor{darkblue}{macaroni2}

    \item 
    \textcolor{darkblue}{pcb1}, ~\textcolor{darkblue}{pcb2}
    
    \item 
    \textcolor{darkblue}{pcb3}, ~\textcolor{darkblue}{pcb4}
\end{enumerate}
In {\it D5-M}, we organize the classes in each session as the following list.
\begin{enumerate}
    \item 
    \textcolor{darkred}{candle}, 
    ~\textcolor{darkblue}{macaroni1}

    \item 
    \textcolor{darkred}{capsules}, 
    ~\textcolor{darkblue}{macaroni2},
    ~\textcolor{darkblue}{pcb1}

    \item 
    \textcolor{darkred}{cashew},
    ~\textcolor{darkblue}{pcb2}

    \item 
    \textcolor{darkred}{chewinggum},
    ~\textcolor{darkblue}{pcb3}

    \item 
    \textcolor{darkred}{fryum},
    ~\textcolor{darkred}{pipe fryum},
    ~\textcolor{darkblue}{pcb4}
\end{enumerate}

\subsection{DAGM}
DAGM is a synthetic anomaly detection dataset, containing 10 classes. We list out the $\gC$ in the $3^{\tt rd}$ row of ~\tabref{table:head-tail}. Each class has a different synthetic pattern. 

Here we use the same design logic for the configurations in MVTec. In {\it D5-HF}, we organize the classes in each session as the following list.
\begin{enumerate}
    \item 
    ~\textcolor{darkblue}{Class9}, ~\textcolor{darkblue}{Class10}

    \item 
    ~\textcolor{darkblue}{Class7},             
    ~\textcolor{darkblue}{Class8}

    \item 
    ~\textcolor{darkblue}{Class2} 
    ~\textcolor{darkred}{Class6},

    \item 
    ~\textcolor{darkred}{Class4}, 
    ~\textcolor{darkred}{Class5}
    
    \item 
    ~\textcolor{darkred}{Class1}, 
    ~\textcolor{darkred}{Class3}
\end{enumerate}

In {\it D5-TF}, we organize the classes in each session as the following list.
\begin{enumerate}
    \item 
    ~\textcolor{darkred}{Class1}, 
    ~\textcolor{darkred}{Class3}
    
    \item 
    ~\textcolor{darkred}{Class4}, 
    ~\textcolor{darkred}{Class5}
    
    \item 
    ~\textcolor{darkred}{Class6},
    ~\textcolor{darkblue}{Class2} 
    
    \item 
    ~\textcolor{darkblue}{Class7},             
    ~\textcolor{darkblue}{Class8}

    \item 
    ~\textcolor{darkblue}{Class9}, ~\textcolor{darkblue}{Class10}
\end{enumerate}

In {\it D5-M}, we organize the classes in each session as the following list.
\begin{enumerate}
    \item 
    ~\textcolor{darkblue}{Class2},
    ~\textcolor{darkred}{Class1}
    
    \item
    ~\textcolor{darkblue}{Class7},
    ~\textcolor{darkred}{Class3}
    
    \item 
    ~\textcolor{darkblue}{Class8},
    ~\textcolor{darkred}{Class4}
    
    \item 
    ~\textcolor{darkblue}{Class9},
    ~\textcolor{darkred}{Class5}

    \item 
    ~\textcolor{darkblue}{Class10},
    ~\textcolor{darkred}{Class6}
\end{enumerate}

\section{Offline experiments}
\label{supp:offline_results}
We provide additional analysis of our approach.
We then report the performances on $\gC^{\tt head}$, on $\gC^{\tt tail}$, and on each $c \in \gC$ in the following tables under various datasets and long-tailed imbalance~\cite{ho2024ltad} settings.

\subsection{Additional analysis}
\myparagraph{Alternative approach to our concept learning}
A naive baseline for our concept learning is to replace the VLM encoder, \ie ALIGN~\cite{jia2021scaling}, with simpler vision architectures like ResNet~\cite{he2016deep} using clustering methods. However, this alternative approach loses all the language information for each class and removes the capability of the semantics module ${\tt S}$. Additionally, long-tailed data distribution is challenging for simple clustering methods. We provide the empirical comparison on MVTec {\it exp100}. While LTOAD achieves image-level AUROC of 85.26 for Det. , the ResNet-based K-means alternative only achieves 57.16.

\myparagraph{Computational cost}
Our method does not introduce computational overhead. We pre-compute the $\hat C$ before training and it takes only 3 minutes to run on one single NVIDIA RTX 6000 GPU. 
Additionally, we use fewer parameters than LTAD~\cite{ho2024ltad} since our $|\hat C| < |\gC|$ thus requires fewer learned prompts.
Lastly, the throughput of LTOAD is 61.74 fps while LTAD’s is 32.78.

\subsection{MVTec, VisA, and DAGM}
\myparagraph{Experiment setup}
We follow the offline LTAD~\cite{ho2024ltad} benchmarks and report on MVTec~\cite{bergmann2019mvtec}, VisA~\cite{zou2022spot}, and DAGM~\cite{wieler2007weakly}. 

\myparagraph{Additional quantitative comparison}
The results are in the corresponding tables:
\begin{itemize}[topsep=0pt, leftmargin=16pt]      
    \item MVTec
    \begin{enumerate}
        \item {\it exp100}: Tab.~\ref{table:mvtec-exp100-det} and Tab.~\ref{table:mvtec-exp100-seg}.
        \item {\it exp200}: Tab.~\ref{table:mvtec-exp200-det} and Tab.~\ref{table:mvtec-exp200-seg}.
        \item {\it step100}: Tab.~\ref{table:mvtec-step100-det} and Tab.~\ref{table:mvtec-step100-seg}.
        \item {\it step200}: Tab.~\ref{table:mvtec-step200-det} and Tab.~\ref{table:mvtec-step200-seg}.
    \end{enumerate}
    \item VisA
    \begin{enumerate}
        \item {\it exp100}: Tab.~\ref{table:visa-exp100-det} and Tab.~\ref{table:visa-exp100-seg}.
        \item {\it exp200}: Tab.~\ref{table:visa-exp200-det} and Tab.~\ref{table:visa-exp200-seg}.
        \item {\it exp500}: Tab.~\ref{table:visa-exp500-det} and Tab.~\ref{table:visa-exp500-seg}.
        \item {\it step100}: Tab.~\ref{table:visa-step100-det} and Tab.~\ref{table:visa-step100-seg}.
        \item {\it step200}: Tab.~\ref{table:visa-step200-det} and Tab.~\ref{table:visa-step200-seg}.
        \item {\it step500}: Tab.~\ref{table:visa-step500-det} and Tab.~\ref{table:visa-step500-seg}.
    \end{enumerate}
    \item DAGM
    \begin{enumerate}
        \item {Overall}: Tab.~\ref{table:dagm-main-offline}.
        \item {\it exp50}: Tab.~\ref{table:dagm-exp50-det} and Tab.~\ref{table:dagm-exp50-seg}.
        \item {\it exp100}: Tab.~\ref{table:dagm-exp100-det} and Tab.~\ref{table:dagm-exp100-seg}.
        \item {\it exp200}: Tab.~\ref{table:dagm-exp200-det} and Tab.~\ref{table:dagm-exp200-seg}.
        \item {\it exp500}: Tab.~\ref{table:dagm-exp500-det} and Tab.~\ref{table:dagm-exp500-seg}.
        \item {\it reverse exp200}: Tab.~\ref{table:dagm-reverse-exp200-det} and Tab.~\ref{table:dagm-reverse-exp200-seg}.
        \item {\it step50}: Tab.~\ref{table:dagm-step50-det} and Tab.~\ref{table:dagm-step50-seg}.
        \item {\it step100}: Tab.~\ref{table:dagm-step100-det} and Tab.~\ref{table:dagm-step100-seg}.
        \item {\it step200}: Tab.~\ref{table:dagm-step200-det} and Tab.~\ref{table:dagm-step200-seg}..
        \item {\it step500}: Tab.~\ref{table:dagm-step500-det} and Tab.~\ref{table:dagm-step500-seg}..
        \item {\it reverse step200}: Tab.~\ref{table:dagm-reverse-step200-det} and Tab.~\ref{table:dagm-reverse-step200-seg}..
    \end{enumerate}
\end{itemize}
We note that in {\it reverse exp200} and {\it reverse step200}, LTAD switches the $\gC^{\tt head}$ and $\gC^{\tt tail}$. We observe a consistent improvement of \ours in comparison to all baselines on most configurations. Notably, \ours excels on the more challenging $\gC^{\tt tail}$ which is the focus of long-tailed settings.

\begin{figure}[t]
    \vspace{-0.2cm}
    \centering
    \includegraphics[width=0.99\linewidth]{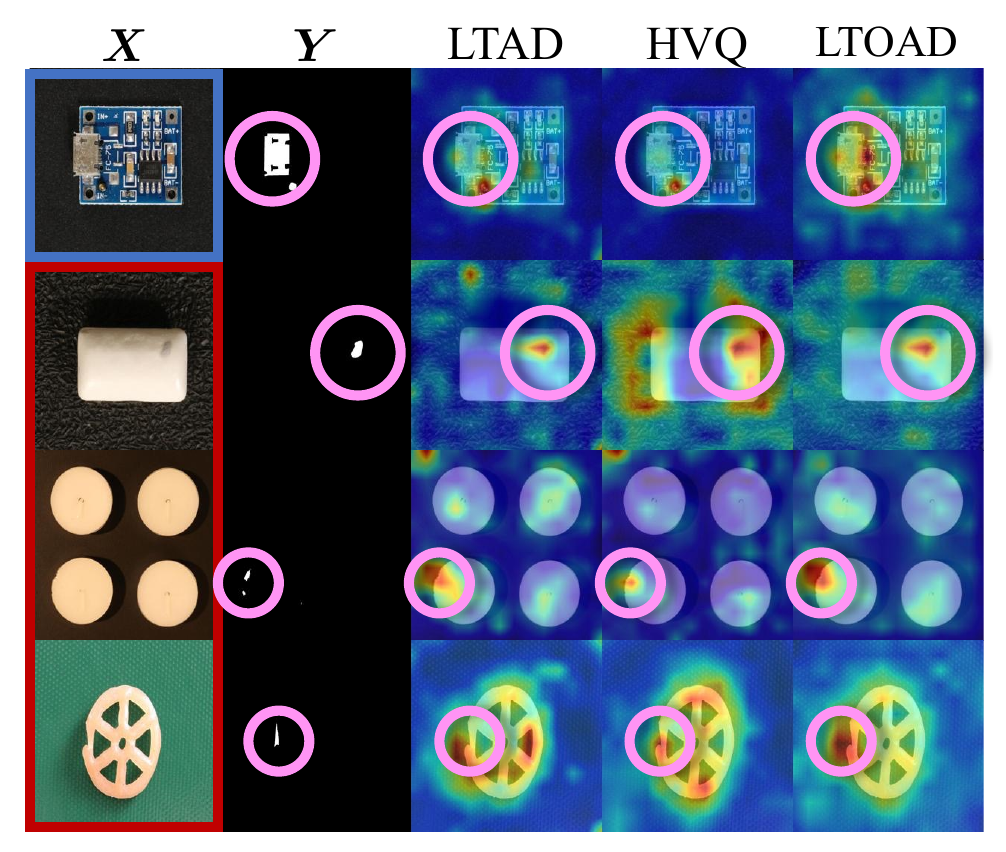}
    \vspace{-0.2cm}
    \captionof{figure}{
        \textbf{Qualitative comparison} among LTAD~\cite{ho2024ltad}, HVQ~\cite{lu2023hierarchical}, and \ours on VisA offline LTAD~\cite{ho2024ltad} {\it exp100} benchmark. Inputs from {\color{darkblue}{$\gC^{\tt head}$}} / {\color{darkred}{$\gC^{\tt tail}$}} are outlined in {\color{darkblue}{blue}} / {\color{darkred}{red}}. Smaller ground truth anomaly masks are circled in {\color{pink}{pink}}.
    }
    \label{fig:qual-visa}     
\end{figure}

\myparagraph{Additional Qualitative Comparison}
Finally, we show additional qualitative results in~\figref{fig:qual-visa}. 

\myparagraph{More evaluation metrics.}
We report more evaluation metrics on MVTec following~\cite{zhang2024ader}, \ie, pixel-level AUPRO (AUPRO$_{\text P}$), mean average precision (mAP), and mean F1-score (mF1-max), for comprehensive comparisons in Tab.~\ref{tab:mvtec-more_metrics}. We observe consistent improvement under all metrics.

\begin{table*}[h!]
    \centering
    \resizebox{0.95\linewidth}{!}{
    \begin{tabular}{c c ccc ccc ccc ccc ccc}
    \hline
        \multirow{2}{*}{Method} & \multicolumn{3}{c}{\it exp100} & \multicolumn{3}{c}{\it exp200} & \multicolumn{3}{c}{\it step100} & \multicolumn{3}{c}{\it step200}
        \\
        \cmidrule(lr){3-5}
        \cmidrule(lr){6-8}
        \cmidrule(lr){9-11}
        \cmidrule(lr){12-14}
        & 
        mAP & mF1-max & AUPRO$_{\text P}$ &
        mAP & mF1-max & AUPRO$_{\text P}$ &
        mAP & mF1-max & AUPRO$_{\text P}$ &
        mAP & mF1-max & AUPRO$_{\text P}$
        \\
        \cmidrule(lr){1-14}
        MoEAD~\cite{mengmoead} & 35.63 & 41.19 & 82.72 & 33.87 & 39.74 & 80.77 & 33.50 & 39.17 & 80.54 & 31.55 & 36.86 & 77.88
        \\
        LTOAD & \bf 38.21 & \bf 43.86 & \bf 85.98 & \bf 37.57 & \bf 42.65 & \bf 85.30 & \bf 38.22 & \bf 43.32 & \bf 85.71 & \bf 35.29 & \bf 40.45 & \bf 82.86
        \\
    \hline
    \end{tabular}
    }
    \caption{
        \textbf{Comparison ($\uparrow$) on MVTec} under more evaluation metrics, \ie mAP, mF1-max, and AUPRO$_{\text P}$.
        \label{tab:mvtec-more_metrics}
    }
\end{table*}

\begin{table}[h]

\centering
\setlength{\tabcolsep}{3pt}

\resizebox{\linewidth}{!}{
\begin{tabular}{ c@{}c c ccc c ccc c ccc c cccc} 
\specialrule{.15em}{.05em}{.05em}
\multirow{2}{*}{Method} & \multirow{2}{*}{CA} &
\multicolumn{2}{c}{\it exp100} && 
\multicolumn{2}{c}{\it exp200} &&
\multicolumn{2}{c}{\it step100} && 
\multicolumn{2}{c}{\it step200}
\\
& & 
Det. & Seg. &&
Det. & Seg. &&
Det. & Seg. &&
Det. & Seg. 
\\
\cmidrule(lr){1-17}
RegAD~\cite{huang2022registration} & \ccross &
84.86 & 90.29 && {84.86} & 90.29 &&
84.86 & 90.29 && 84.86 & 90.29
\\
{AnomalyGPT~\cite{gu2024agpt}} & \ccross &
85.31 & 77.20 && 83.29 & 77.16 &&
86.48 & 78.76 && 84.73 & 78.29 
\\
{LTAD~\cite{ho2024ltad}} & \ccross & 
{94.40} & \ul{97.30} && \ul{94.29} & \bf 97.19 &&
{93.97} & \ul{97.07} && {93.79} & \ul{96.84}
\\
HVQ~\cite{lu2023hierarchical} & \ccross &
75.42 & 85.54 && 76.19 & 85.32 &&
75.78 & 85.23 && 75.92 & 84.63
\\
\rowcolor{OursColor} {\ours*} & \ccross &
\ul{94.92} & {96.81} && 94.13 & {96.77} &&
\ul{94.18} & 96.90 && \ul{93.88} & {96.64}
\\
\cmidrule(lr){1-17}
UniAD~\cite{you2022unified} & \ccheck &
84.34 & 90.13 && 83.56 & 89.73 &&
81.11 & 89.11 && 80.33 & 89.07
\\
MoEAD~\cite{mengmoead} & \ccheck &
73.26 & 84.99 && 72.61 & 84.18 &&
70.83 & 83.34 && 70.01 & 82.27
\\
\rowcolor{OursColor} {\ours} & \ccheck &
\bf 95.58 & \bf{97.32} && \bf 94.71 & \ul{97.11} &&
\bf 94.83 & \bf 97.30 && \bf 94.12 & \bf 97.14
\\
\specialrule{.15em}{.05em}{.05em}
\end{tabular}
}

\caption{
    \textbf{Comparison ($\uparrow$) on offline DAGM} in the same format as Tab.~\ref{table:mvtec-main-offline}
}
\vspace{-0.25cm}
\label{table:dagm-main-offline}

\end{table}

\subsection{Uni-Medical}

\myparagraph{Experiment setup} We provided additional results on  Uni-Medical~\cite{zhang2023exploring, bao2024bmad} under long-tailed training settings. Uni-Medical is a medical anomaly detection dataset extracted from BMAD~\cite{bao2024bmad}. It consists of the following benchmarks of medical imaging: Brain MRI~\cite{baid2021rsna}, Liver CT BTCV~\cite{landman2015miccai, bilic2023liver}, and Retinal OCT~\cite{hu2019automated}.
We use the same design logic of LTAD benchmark~\cite{ho2024ltad} and create the long-tailed training configurations for Uni-Medical. 
Since there are only 3 classes in Uni-Medical, we experiment with cases of each being the head class.
We document the configurations of $\gD^{\tt O}$ in Tab.~\ref{table:uni_medical-configs}. 
\begin{table}[h]

\centering
\setlength{\tabcolsep}{3pt}

\resizebox{\linewidth}{!}{%
\begin{tabular}[t]{ c cc cc ccc } 
\specialrule{.15em}{.05em}{.05em}
\multirow{2}{*}{Config.} & \multirow{2}{*}{\textcolor{darkblue}{$\gC^{\tt head}$}} & \multirow{2}{*}{Imbalance type} & \multicolumn{3}{c}{$N_{c}$}
\\
\cmidrule(lr){4-6}
& & & brain & liver & retinal
\\
\cmidrule(lr){1-6}
{\it brain exp100} & brain & exponential decay & \textcolor{darkblue}{1542} & \textcolor{darkred}{154} & \textcolor{darkred}{15}
\\
{\it brain step100} & brain & step function & \textcolor{darkblue}{1542} & \textcolor{darkred}{15} & \textcolor{darkred}{15}
\\
{\it liver exp100} & liver & exponential decay  & \textcolor{darkred}{15} & \textcolor{darkblue}{1542} & \textcolor{darkred}{154}
\\
{\it liver step100} & liver & step function & \textcolor{darkred}{15} & \textcolor{darkblue}{1542} & \textcolor{darkred}{15}
\\
{\it retinal exp100} & retinal & exponential decay & \textcolor{darkred}{154} & \textcolor{darkred}{15} & \textcolor{darkblue}{1542}
\\
{\it retinal step100} & retinal & step function & \textcolor{darkred}{15} & \textcolor{darkred}{15} & \textcolor{darkblue}{1542}
\\
\specialrule{.15em}{.05em}{.05em}
\end{tabular}
}
\caption{
    \textbf{Configurtions of $\gD^{\tt T}$ for long-tailed Uni-Medical.} We denote the number of images per class $c$ as $N_c$ and highlight the $N_c$ of $c \in$ \textcolor{darkblue}{$\gC^{\tt head}$} or \textcolor{darkred}{$\gC^{tail}$}. We set all imbalance factors to 100.
}
\label{table:uni_medical-configs}

\end{table}

\myparagraph{Baselines \& training details}. We compare with the
SOTA MoEAD~\cite{mengmoead}~($\gG_1$) and HVQ~\cite{lu2023hierarchical}~($\gG_2$). We set $\hat K =10$.
We train \ours for 25 epochs on {\it liver exp100} and {\it liver step100}. On other configurations, we train for 5 epochs. For detection, we reduce $\hat \mY$ to $\hat y$ by calculating its standard deviation.

\myparagraph{Quantitative comparison}
The results are in Tab.~\ref{table:uni_medical-brain-exp100}-\ref{table:uni_medical-retinal-step100} in the corresponding tables:
\begin{itemize}[topsep=0pt, leftmargin=16pt]
    \item {\it brain exp100}: Tab.~\ref{table:uni_medical-brain-exp100}.
    \item {\it brain step100}: Tab.~\ref{table:uni_medical-brain-step100}.
    \item {\it liver exp100}: Tab.~\ref{table:uni_medical-liver-exp100}.
    \item {\it liver step100}: Tab.~\ref{table:uni_medical-liver-step100}.
    \item {\it retinal exp100}: Tab.~\ref{table:uni_medical-retinal-exp100}.
    \item {\it retinal step100}: Tab.~\ref{table:uni_medical-retinal-step100}.
\end{itemize}
On most configurations, \ours outperforms class-agnostic MoEAD while being competitive to class-aware HVQ which requires additional information.

\begin{table*}[ht]
    \centering
    \begin{minipage}[t]{0.48\linewidth}

\centering
\setlength{\tabcolsep}{3pt}

\resizebox{\linewidth}{!}{%
\begin{tabular}[h]{*{100}{c}}
\specialrule{.15em}{.05em}{.05em}
\multirow{2}{*}{Method} & \multirow{2}{*}{CA} &
\multicolumn{2}{c}{\it $\gC$} &
\multicolumn{2}{c}{\color{darkblue}$\gC^{\tt head}$} &
\multicolumn{2}{c}{\color{darkred}$\gC^{\tt tail}$} &
\multicolumn{2}{c}{\color{darkblue}\it brain} &
\multicolumn{2}{c}{\color{darkred}\it liver} &
\multicolumn{2}{c}{\color{darkred}\it retinal}
\\
& & 
Det. & Seg. &
Det. & Seg. &
Det. & Seg. &
Det. & Seg. &
Det. & Seg. &
Det. & Seg.
\\
\cmidrule(lr){1-38}
HVQ~\cite{lu2023hierarchical} & \ccross &
65.21 & 93.84 & 83.15 & 97.03 & 56.23 & 92.25 & 83.15 & 97.03 & 45.72 & \bf 95.42 & 66.74 & 89.07
\\
MoEAD~\cite{mengmoead} & \ccheck &
60.16 & 91.09 & \bf 86.74 & \bf 97.50 & 46.88 & 87.88 & \bf 86.74 & \bf 97.50 & 43.60 & 94.82 & 50.15 & 80.95
\\
\rowcolor{OursColor} {\ours} & \ccheck &
\bf 72.81 & \bf 94.93 & 80.77 & 96.32 & \bf 68.83 & \bf 94.24 & 80.77 & 96.32 & \bf 58.05 & 95.07 & \bf 79.60 & \bf 93.42
\\
\specialrule{.15em}{.05em}{.05em}
\end{tabular}
}

\captionof{table}{
    \textbf{Comparison ($\uparrow$) on Uni-Medical \it{brain exp100}} using the same evaluation metrics as Tab.~\ref{table:mvtec-exp100-det} and Tab.~\ref{table:mvtec-exp100-seg}. 
}
\label{table:uni_medical-brain-exp100}


    \end{minipage}
    \hfill
    \begin{minipage}[t]{0.48\linewidth}

\centering
\setlength{\tabcolsep}{3pt}

\resizebox{\linewidth}{!}{%
\begin{tabular}[h]{*{100}{c}}
\specialrule{.15em}{.05em}{.05em}
\multirow{2}{*}{Method} & \multirow{2}{*}{CA} &
\multicolumn{2}{c}{\it $\gC$} &
\multicolumn{2}{c}{\color{darkblue}$\gC^{\tt head}$} &
\multicolumn{2}{c}{\color{darkred}$\gC^{\tt tail}$} &
\multicolumn{2}{c}{\color{darkblue}\it brain} &
\multicolumn{2}{c}{\color{darkred}\it liver} &
\multicolumn{2}{c}{\color{darkred}\it retinal}
\\
& & 
Det. & Seg. &
Det. & Seg. &
Det. & Seg. &
Det. & Seg. &
Det. & Seg. &
Det. & Seg.
\\
\cmidrule(lr){1-38}
HVQ~\cite{lu2023hierarchical} & \ccross &
69.23 & 93.89 & \bf 79.60 & \bf 97.22 & 64.04 & 92.23 & \bf 79.60 & \bf 97.22 & \bf 57.74 & \bf 95.70 & 70.34 & 88.76 
\\
MoEAD~\cite{mengmoead} & \ccheck &
61.11 & 90.99 & 78.21 & 96.88 & 52.56 & 88.05 & 50.88 & 92.83 & 54.24 & 83.26 & 78.21 & \bf 96.88 
\\
\rowcolor{OursColor} {\ours} & \ccheck &
\bf 73.44 & \bf 94.61 & 78.13 & 96.08 & \bf 71.10 & \bf 93.88 & 78.13 & 96.08 & 56.31 & 93.94 & \bf 85.89 & 93.81 
\\
\specialrule{.15em}{.05em}{.05em}
\end{tabular}
}

\captionof{table}{
    \textbf{Comparison ($\uparrow$) on Uni-Medical \it{brain step100}} using the same evaluation metrics as Tab.~\ref{table:uni_medical-brain-exp100}. 
}
\label{table:uni_medical-brain-step100}


    \end{minipage}

    \begin{minipage}[t]{0.48\linewidth}

\centering
\setlength{\tabcolsep}{3pt}

\resizebox{\linewidth}{!}{%
\begin{tabular}[h]{*{100}{c}}
\specialrule{.15em}{.05em}{.05em}
\multirow{2}{*}{Method} & \multirow{2}{*}{CA} &
\multicolumn{2}{c}{\it $\gC$} &
\multicolumn{2}{c}{\color{darkblue}$\gC^{\tt head}$} &
\multicolumn{2}{c}{\color{darkred}$\gC^{\tt tail}$} &
\multicolumn{2}{c}{\color{darkblue}\it liver} &
\multicolumn{2}{c}{\color{darkred}\it retinal} &
\multicolumn{2}{c}{\color{darkred}\it brain}
\\
& & 
Det. & Seg. &
Det. & Seg. &
Det. & Seg. &
Det. & Seg. &
Det. & Seg. &
Det. & Seg.
\\
\cmidrule(lr){1-38}
HVQ~\cite{lu2023hierarchical} & \ccross &
\bf 72.50 & 94.49 & \bf 75.10 & 94.45 & 71.20 & 94.51 & \bf 59.10 & \bf 96.92 & 83.30 & 92.11 & \bf 75.10 & 94.45
\\
MoEAD~\cite{mengmoead} & \ccheck &
59.22 & 91.44 & 58.14 & 92.11 & 59.76 & 91.10 & 52.48 & 95.40 & 67.05 & 86.80 & 58.14 & 92.11
\\
\rowcolor{OursColor} {\ours} & \ccheck &
69.19 & \bf 94.53 & 63.91 & \bf 94.52 & \bf 71.82 & \bf 94.54 & 58.91 & 95.69 & \bf 84.74 & \bf 93.38 & 63.91 & \bf 94.52
\\
\specialrule{.15em}{.05em}{.05em}
\end{tabular}
}

\captionof{table}{
    \textbf{Comparison ($\uparrow$) on Uni-Medical \it{liver exp100}} using the same evaluation metrics as Tab.~\ref{table:uni_medical-brain-exp100}. 
}
\label{table:uni_medical-liver-exp100}


    \end{minipage}
    \hfill
    \begin{minipage}[t]{0.48\linewidth}

\centering
\setlength{\tabcolsep}{3pt}

\resizebox{\linewidth}{!}{%
\begin{tabular}[h]{*{100}{c}}
\specialrule{.15em}{.05em}{.05em}
\multirow{2}{*}{Method} & \multirow{2}{*}{CA} &
\multicolumn{2}{c}{\it $\gC$} &
\multicolumn{2}{c}{\color{darkblue}$\gC^{\tt head}$} &
\multicolumn{2}{c}{\color{darkred}$\gC^{\tt tail}$} &
\multicolumn{2}{c}{\color{darkblue}\it liver} &
\multicolumn{2}{c}{\color{darkred}\it retinal} &
\multicolumn{2}{c}{\color{darkred}\it brain}
\\
& & 
Det. & Seg. &
Det. & Seg. &
Det. & Seg. &
Det. & Seg. &
Det. & Seg. &
Det. & Seg.
\\
\cmidrule(lr){1-38}
HVQ~\cite{lu2023hierarchical} & \ccross &
66.88 & 93.34 & \bf 73.63 & 94.31 & 63.50 & 92.86 & 58.94 & \bf 96.80 & 68.06 & 88.92 & \bf 73.63 & 94.31
\\
MoEAD~\cite{mengmoead} & \ccheck &
57.01 & 90.39 & 61.12 & 92.19 & 54.96 & 89.49 & 54.60 & 95.86 & 55.32 & 83.13 & 61.12 & 92.19
\\
\rowcolor{OursColor} {\ours} & \ccheck &
\bf 70.27 & \bf 94.53 & 67.26 & \bf 94.42 & \bf 71.77 & \bf 94.58 & \bf 59.19 & 95.76 & \bf 84.35 & \bf 93.40 & 67.26 & \bf 94.42
\\
\specialrule{.15em}{.05em}{.05em}
\end{tabular}
}

\captionof{table}{
    \textbf{Comparison ($\uparrow$) on Uni-Medical \it{liver step100}} using the same evaluation metrics as Tab.~\ref{table:uni_medical-brain-exp100}. 
}
\label{table:uni_medical-liver-step100}


    \end{minipage}

    \begin{minipage}[t]{0.48\linewidth}

\centering
\setlength{\tabcolsep}{3pt}

\resizebox{\linewidth}{!}{%
\begin{tabular}[h]{*{100}{c}}
\specialrule{.15em}{.05em}{.05em}
\multirow{2}{*}{Method} & \multirow{2}{*}{CA} &
\multicolumn{2}{c}{\it $\gC$} &
\multicolumn{2}{c}{\color{darkblue}$\gC^{\tt head}$} &
\multicolumn{2}{c}{\color{darkred}$\gC^{\tt tail}$} &
\multicolumn{2}{c}{\color{darkblue}\it retinal} &
\multicolumn{2}{c}{\color{darkred}\it brain} &
\multicolumn{2}{c}{\color{darkred}\it liver}
\\
& & 
Det. & Seg. &
Det. & Seg. &
Det. & Seg. &
Det. & Seg. &
Det. & Seg. &
Det. & Seg.
\\
\cmidrule(lr){1-38}
HVQ~\cite{lu2023hierarchical} & \ccross &
74.80 & \bf 95.04 & 80.91 & \bf 96.14 & 71.75 & \bf 94.49 & \bf 86.15 & 93.19 & 80.91 & \bf 96.14 & 57.34 & \bf 95.78
\\
MoEAD~\cite{mengmoead} & \ccheck &
69.87 & 93.57 & 78.12 & 95.27 & 65.75 & 92.72 & 79.55 & 92.20 & 78.12 & 95.27 & 51.95 & 93.24
\\
\rowcolor{OursColor} {\ours} & \ccheck &
\bf 75.55 & 94.49 & \bf 83.03 & 95.53 & \bf 71.81 & 93.97 & 85.88 & \bf 94.19 & \bf 83.03 & 95.53 & \bf 57.75 & 93.75
\\
\specialrule{.15em}{.05em}{.05em}
\end{tabular}
}

\captionof{table}{
    \textbf{Comparison ($\uparrow$) on Uni-Medical \it{retinal exp100}} using the same evaluation metrics as Tab.~\ref{table:uni_medical-brain-exp100}. 
}
\label{table:uni_medical-retinal-exp100}


    \end{minipage}
    \hfill
    \begin{minipage}[t]{0.48\linewidth}

\centering
\setlength{\tabcolsep}{3pt}

\resizebox{\linewidth}{!}{%
\begin{tabular}[h]{*{100}{c}}
\specialrule{.15em}{.05em}{.05em}
\multirow{2}{*}{Method} & \multirow{2}{*}{CA} &
\multicolumn{2}{c}{\it $\gC$} &
\multicolumn{2}{c}{\color{darkblue}$\gC^{\tt head}$} &
\multicolumn{2}{c}{\color{darkred}$\gC^{\tt tail}$} &
\multicolumn{2}{c}{\color{darkblue}\it retinal} &
\multicolumn{2}{c}{\color{darkred}\it brain} &
\multicolumn{2}{c}{\color{darkred}\it liver}
\\
& & 
Det. & Seg. &
Det. & Seg. &
Det. & Seg. &
Det. & Seg. &
Det. & Seg. &
Det. & Seg.
\\
\cmidrule(lr){1-38}
HVQ~\cite{lu2023hierarchical} & \ccross &
\bf 71.44 & \bf 94.41 & \bf 72.07 & 93.96 & 71.13 & \bf 94.63 & \bf 87.65 & 93.68 & \bf 72.07 & 93.96 & 54.61 & \bf 95.58
\\
MoEAD~\cite{mengmoead} & \ccheck &
64.96 & 92.87 & 62.23 & 92.79 & 66.32 & 92.91 & 81.71 & 92.28 & 62.23 & 92.79 & 50.94 & 93.55
\\
\rowcolor{OursColor} {\ours} & \ccheck &
71.31 & 93.69 & 71.11 & \bf 94.39 & \bf 71.41 & 93.34 & 86.17 & \bf 94.70 & 71.11 & \bf 94.39 & \bf 56.66 & 91.98
\\
\specialrule{.15em}{.05em}{.05em}
\end{tabular}
}

\captionof{table}{
    \textbf{Comparison ($\uparrow$) on Uni-Medical \it{retinal step100}} using the same evaluation metrics as Tab~\ref{table:uni_medical-brain-exp100}. 
}
\label{table:uni_medical-retinal-step100}


    \end{minipage}
\end{table*}

\begin{table*}[ht]

\centering
\setlength{\tabcolsep}{3pt}

\resizebox{\linewidth}{!}{%
\begin{tabular}[h]{*{100}{c}}
\specialrule{.15em}{.05em}{.05em}
Method & CA &
\it $\gC$ &
\color{darkblue}$\gC^{\tt head}$ &
\color{darkred}$\gC^{\tt tail}$ &
\color{darkblue}\it hazelnut &
\color{darkblue}\it leather &
\color{darkblue}\it bottle &
{\color{darkblue}\it wood} &
{\color{darkblue}\it carpet} &
{\color{darkblue}\it tile} &
{\color{darkblue}\it metal nut} &
{\color{darkred}\it toothbrush} &
{\color{darkred}\it zipper} &
{\color{darkred}\it transistor} &
{\color{darkred}\it grid} &
{\color{darkred}\it pill} &
{\color{darkred}\it capsule} &
{\color{darkred}\it cable} &
{\color{darkred}{\it screw}}
\\
\cmidrule(lr){1-38}
MKD~\cite{salehi2021multiresolution} & \ccross & 78.92 & 83.11 & 75.26 & 97.00 & 78.94 & 98.97 & 91.05 & 74.48 & 78.94 & 62.37 & 72.22 & 90.49 & 79.33 & 64.24 & 78.40 & 69.01 & 78.54 & 69.81
\\
DRAEM~\cite{zavrtanik2021draem} & \ccross & 79.57 & 95.22 & 65.87 & 98.32 & 98.91 & 99.68 & \bf 99.91 & 92.17 & 97.94 & 79.66 & 83.61 & 93.01 & 62.54 & 43.02 & 68.14 & 41.76 & 53.31 & \bf 81.57
\\
RegAD~\cite{huang2022registration} & \ccross & 82.43 & 90.04 & 70.51 & 91.84 & 98.61 & 99.20 & 97.86 & 97.44 & 97.59 & 89.77 & 75.69 & 85.58 & 81.68 & 68.80 & 69.75 & 64.67 & 64.67 & 53.30
\\
AnomalyGPT~\cite{gu2024agpt} & \ccross & 87.44 & 96.40 & 79.60 & 96.21 & \bf 100.00 & 96.75 & 88.07 & 98.19 & 97.66 & 97.95 & 95.00 & 87.16 & 79.77 & 91.98 & 84.44 & 67.03 & 69.33 & 62.10
\\
LTAD~\cite{ho2024ltad} & \ccross & 88.86 & 99.09 & 79.90 & 99.82 & \bf 100.00 & 99.92 & 99.30 & 99.88 & 99.24 & 95.50 & 87.50 & 89.23 & 93.79 & 87.80 & 84.04 & 64.82 & 77.49 & 54.56
\\
HVQ~\cite{lu2023hierarchical} & \ccross & 87.43 & 99.34 & 77.00 & 99.93 & \bf 100.00 & \bf 100.00 & 97.54 & 99.88 & 98.99 & 99.02 & 85.00 & \bf 93.67 & 89.71 & 78.61 & 90.51 & 70.12 & 60.04 & 48.35
\\
\cmidrule(lr){1-38}
UniAD~\cite{you2022unified} & \ccheck & 87.70 & 99.27 & 77.58 & \bf 100.00 & \bf 100.00 & 99.76 & 99.56 & 99.79 & \bf 99.38 & 96.43 & 88.61 & 86.65 & 91.58 & 92.64 & 81.58 & 63.78 & 63.64 & 52.22
\\
MoEAD~\cite{mengmoead} & \ccheck & 84.73 & 98.94 & 72.29 & 99.21 & \bf 100.00 & 99.52 & 97.54 & \bf 100.00 & 97.84 & 98.48 & 90.28 & 88.79 & 89.75 & 65.16 & 66.91 & 66.69 & 61.15 & 49.62
\\
\rowcolor{OursColor} {\ours} & \ccheck & \bf 93.42 & \bf 99.45 & \bf 88.14 & \bf 100.00 & \bf 100.00 & 99.52 & 98.51 & \bf 100.00 & 98.67 & \bf 99.46 & \bf 96.39 & 91.81 & \bf 96.50 & \bf 97.33 & \bf 94.27 & \bf 83.05 & \bf 87.82 & 57.98
\\
\specialrule{.15em}{.05em}{.05em}
\end{tabular}
}

\caption{
    \textbf{Comparison ($\uparrow$) on MVTec \it{exp100}}~\cite{ho2024ltad} in image-level AUROC for anomaly detection (Det.). The column CA (class-agnostic) indicates whether a method requires class names or the number of classes during training or not (require: \ccross; not require: $\ccheck$). We report the class-wise performance on {\color{darkblue}{$\gC^{\tt head}$}} and {\color{darkred}{$\gC^{\tt tail}$}}. The classes are sorted from having the most images (left) to the least (right).
}
\label{table:mvtec-exp100-det}
\end{table*}

\begin{table*}[ht]

\centering
\setlength{\tabcolsep}{3pt}

\resizebox{\linewidth}{!}{%
\begin{tabular}[h]{*{100}{c}}
\specialrule{.15em}{.05em}{.05em}
{Method} & {CA} &
{\it $\gC$} &
{\color{darkblue}$\gC^{\tt head}$} &
{\color{darkred}$\gC^{\tt tail}$} &
{\color{darkblue}\it hazelnut} &
{\color{darkblue}\it leather} &
{\color{darkblue}\it bottle} &
{\color{darkblue}\it wood} &
{\color{darkblue}\it carpet} &
{\color{darkblue}\it tile} &
{\color{darkblue}\it metal nut} &
{\color{darkred}\it toothbrush} &
{\color{darkred}\it zipper} &
{\color{darkred}\it transistor} &
{\color{darkred}\it grid} &
{\color{darkred}\it pill} &
{\color{darkred}\it capsule} &
{\color{darkred}\it cable} &
{\color{darkred}{\it screw}}
\\
\cmidrule(lr){1-38}
MKD~\cite{salehi2021multiresolution} & \ccross &
85.95 & 87.39 & 84.69 & 95.00 & 96.87 & 93.10 & 77.74 & 93.05 & 76.84 & 79.13 & 93.86 & 90.14 & 73.69 & 72.36 & 88.75 & 93.16 & 71.67 & 93.95
\\
DRAEM~\cite{zavrtanik2021draem} & \ccross & 
85.17 & 93.68 & 77.73 & 97.80 & 97.56 & 95.18 & \bf 97.33 & 94.33 & \bf 97.36 & 76.21 & 97.34 & \bf 97.43 & 66.95 & 76.55 & 89.00 & 44.65 & 59.90 & 90.08
\\
RegAD~\cite{huang2022registration} & \ccross & 
95.20 & \bf 96.92 & 93.69 & 98.07 & 99.15 & 98.12 & 95.46 & 98.67 & 94.18 & 94.81 & 96.88 & 97.12 & 92.44 & 79.05 & \bf 97.29 & 96.42 & \bf 96.42 & \bf 93.96
\\
{AnomalyGPT~\cite{gu2024agpt}} & \ccross &
89.68 & 93.64 & 86.21 & 92.06 & \bf 99.58 & 93.42 & 89.47 & \bf 98.97 & 94.00 & 88.02 & 97.06 & 93.22 & 67.18 & 94.33 & 75.00 & 87.84 & 85.00 & 89.17
\\
{LTAD~\cite{ho2024ltad}} & \ccross &
94.46 & 95.96 & 93.15 & 98.12 & 99.29 & 96.84 & 90.45 & 98.52 & 92.52 & 95.96 & 97.91 & 93.93 & 96.89 & 93.76 & 90.09 & 95.02 & 91.67 & 85.92
\\
HVQ~\cite{lu2023hierarchical} & \ccross &
\bf 95.25 & 96.36 & 94.28 & \bf 98.54 & 98.80 & \bf 98.26 & 92.26 & 98.69 & 91.55 & \bf 96.39 & 98.24 & 96.59 & \bf 97.38 & 91.72 & 93.39 & \bf 98.17 & 90.74 & 87.98
\\
\cmidrule(lr){1-38}
UniAD~\cite{you2022unified} & \ccheck &
93.95 & 95.46 & 92.64 & 97.83 & 98.79 & 95.64 & 90.66 & 98.52 & 91.65 & 95.18 & 97.96 & 92.35 & 96.38 & 93.80 & 88.66 & 94.68 & 88.93 & 88.36
\\
MoEAD~\cite{mengmoead} & \ccheck &
94.34 & 96.02 & 92.87 & 98.00 & 98.86 & 98.14 & 92.21 & 98.80 & 90.34 & 95.78 & \bf 98.31 & 95.24 & 96.69 & 87.33 & 89.91 & 97.14 & 90.59 & 87.71
\\
\rowcolor{OursColor} {\ours} & \ccheck &
95.21 & 95.64 & \bf 94.83 & 98.16 & 98.57 & 96.69 & 91.01 & 98.72 & 93.57 & 92.79 & 98.14 & 93.24 & 96.82 & \bf 95.14 & 93.71 & 95.86 & 94.08 & 91.70
\\
\specialrule{.15em}{.05em}{.05em}
\end{tabular}
}

\caption{
    \textbf{Comparison ($\uparrow$) on MVTec \it{exp100}}~\cite{ho2024ltad} in pixel-level AUROC for anomaly segmentation (Seg.). The column CA (class-agnostic) indicates whether a method requires class names or the number of classes during training or not (require: \ccross; not require: $\ccheck$). We report the class-wise performance on {\color{darkblue}{$\gC^{\tt head}$}} and {\color{darkred}{$\gC^{\tt tail}$}}. The classes are sorted from having the most images (left) to the least (right).
}
\label{table:mvtec-exp100-seg}

\end{table*}


\begin{table*}[ht]

\centering
\setlength{\tabcolsep}{3pt}

\resizebox{\linewidth}{!}{%
\begin{tabular}[h]{*{100}{c}}
\specialrule{.15em}{.05em}{.05em}
{Method} & {CA} &
{\it $\gC$} &
{\color{darkblue}$\gC^{\tt head}$} &
{\color{darkred}$\gC^{\tt tail}$} &
{\color{darkblue}\it hazelnut} &
{\color{darkblue}\it leather} &
{\color{darkblue}\it bottle} &
{\color{darkblue}\it wood} &
{\color{darkblue}\it carpet} &
{\color{darkblue}\it tile} &
{\color{darkblue}\it metal nut} &
{\color{darkred}\it toothbrush} &
{\color{darkred}\it zipper} &
{\color{darkred}\it transistor} &
{\color{darkred}\it grid} &
{\color{darkred}\it pill} &
{\color{darkred}\it capsule} &
{\color{darkred}\it cable} &
{\color{darkred}{\it screw}}
\\
\cmidrule(lr){1-38}
MKD~\cite{salehi2021multiresolution} & \ccross & 79.93 & 84.47 & 75.96 & 96.93 & 78.19 & 98.97 & 91.23 & 74.36 & 86.98 & 64.66 & 72.78 & 89.97 & 79.08 & 67.84 & 76.70 & 70.56 & 78.84 & 71.90
\\
DRAEM~\cite{zavrtanik2021draem} & \ccross & 78.82 & 96.39 & 63.44 & 97.10 & 99.89 & 99.76 & \bf 99.91 & 95.78 & 98.05 & 84.26 & 80.00 & 88.07 & 60.00 & 46.70 & 50.19 & 33.58 & 54.96 & \bf 94.09
\\
AnomalyGPT~\cite{gu2024agpt} & \ccross & 85.80 & 97.01 & 76.00 & 98.57 & \bf 100.00 & 96.59 & 89.04 & 98.92 & 97.84 & 98.09 & 88.61 & 86.52 & 77.06 & 90.14 & 88.54 & 62.76 & 57.01 & 57.38
\\
LTAD~\cite{ho2024ltad} & \ccross & 86.05 & 98.99 & 74.73 & 99.29 & 99.93 & 99.76 & 99.56 & 99.96 & 99.39 & 95.06 & 86.94 & 85.56 & 79.67 & 88.47 & 81.91 & 64.42 & 58.62 & 52.22
\\
HVQ~\cite{lu2023hierarchical} & \ccross & 84.90 & 99.14 & 72.43 & \bf 99.89 & \bf 100.00 & \bf 100.00 & 98.16 & 99.76 & \bf 99.93 & 96.24 & 84.44 & 95.01 & 66.62 & 81.45 & 77.69 & 68.01 & 61.08 & 45.15
\\
\end{tabular}
}

\caption{
    \textbf{Comparison ($\uparrow$) on MVTec \it{exp200}}~\cite{ho2024ltad} in image-level AUROC using the same evaluation metrics as Tab.~\ref{table:mvtec-exp100-det}.
}
\label{table:mvtec-exp200-det}

\end{table*}

\begin{table*}[ht]

\centering
\setlength{\tabcolsep}{3pt}

\resizebox{\linewidth}{!}{%
\begin{tabular}[h]{*{100}{c}}
\specialrule{.15em}{.05em}{.05em}
{Method} & {CA} &
{\it $\gC$} &
{\color{darkblue}$\gC^{\tt head}$} &
{\color{darkred}$\gC^{\tt tail}$} &
{\color{darkblue}\it hazelnut} &
{\color{darkblue}\it leather} &
{\color{darkblue}\it bottle} &
{\color{darkblue}\it wood} &
{\color{darkblue}\it carpet} &
{\color{darkblue}\it tile} &
{\color{darkblue}\it metal nut} &
{\color{darkred}\it toothbrush} &
{\color{darkred}\it zipper} &
{\color{darkred}\it transistor} &
{\color{darkred}\it grid} &
{\color{darkred}\it pill} &
{\color{darkred}\it capsule} &
{\color{darkred}\it cable} &
{\color{darkred}{\it screw}}
\\
\cmidrule(lr){1-38}
MKD~\cite{salehi2021multiresolution} & \ccross & 86.01 & 87.15 & 85.02 & 94.01 & 96.81 & 93.02 & 77.79 & 93.25 & 77.04 & 78.10 & 93.53 & 90.26 & 75.28 & 72.88 & 88.81 & 93.16 & 72.45 & \bf 93.76
\\
DRAEM~\cite{zavrtanik2021draem} & \ccross & 82.95 & 93.98 & 73.30 & 96.96 & 99.15 & 94.34 & \bf 95.62 & 95.25 & \bf 95.34 & 81.24 & 96.43 & 96.15 & 59.05 & 77.78 & 83.77 & 38.02 & 44.28 & 90.94
\\
AnomalyGPT~\cite{gu2024agpt} & \ccross & 90.15 & 94.01 & 86.76 & 92.47 & \bf 99.54 & 94.34 & 89.70 & \bf 98.96 & 94.62 & 88.48 & 96.87 & 92.39 & 67.64 & 94.95 & 76.63 & 89.01 & 85.46 & 91.16
\\
LTAD~\cite{ho2024ltad} & \ccross & 94.18 & 96.25 & 92.37 & 97.87 & 99.04 & 96.79 & 92.16 & 98.65 & 94.23 & 95.00 & 97.83 & 92.86 & 93.45 & 93.85 & 93.11 & 95.37 & 85.92 & 86.55
\\
HVQ~\cite{lu2023hierarchical} & \ccross & 94.79 & \bf 96.32 & 93.44 & \bf 98.61 & 98.78 & \bf 98.23 & 92.14 & 98.66 & 91.53 & \bf 96.32 & 98.25 & \bf 96.16 & 93.08 & 91.34 & 92.10 & \bf 97.86 & 91.26 & 87.47
\\
\end{tabular}
}

\caption{
    \textbf{Comparison ($\uparrow$) on MVTec \it{exp200}}~\cite{ho2024ltad} in pixel-level AUROC  using the same evaluation metrics as Tab.~\ref{table:mvtec-exp100-seg}.
}
\label{table:mvtec-exp200-seg}

\end{table*}

\begin{table*}[ht]

\centering
\setlength{\tabcolsep}{3pt}

\resizebox{\linewidth}{!}{%
\begin{tabular}[h]{*{100}{c}}
\specialrule{.15em}{.05em}{.05em}
{Method} & {CA} &
{\it $\gC$} &
{\color{darkblue}$\gC^{\tt head}$} &
{\color{darkred}$\gC^{\tt tail}$} &
{\color{darkblue}\it hazelnut} &
{\color{darkblue}\it leather} &
{\color{darkblue}\it bottle} &
{\color{darkblue}\it wood} &
{\color{darkblue}\it carpet} &
{\color{darkblue}\it tile} &
{\color{darkblue}\it metal nut} &
{\color{darkred}\it toothbrush} &
{\color{darkred}\it zipper} &
{\color{darkred}\it transistor} &
{\color{darkred}\it grid} &
{\color{darkred}\it pill} &
{\color{darkred}\it capsule} &
{\color{darkred}\it cable} &
{\color{darkred}{\it screw}}
\\
\cmidrule(lr){1-38}
MKD~\cite{salehi2021multiresolution} & \ccross & 79.61 & 84.34 & 75.46 & 96.61 & 78.43 & 98.81 & 91.32 & 75.96 & 85.14 & 64.13 & 76.39 & 90.36 & 79.33 & 62.91 & 78.18 & 69.57 & 78.86 & 68.15
\\
DRAEM~\cite{zavrtanik2021draem} & \ccross & 69.82 & 84.00 & 57.41 & \bf 99.96 & 95.78 & 99.44 & \bf 100.00 & 97.47 & 99.67 & 94.72 & 72.77 & 70.79 & 51.12 & 28.15 & 57.50 & 37.89 & 50.50 & \bf 90.57
\\
RegAD~\cite{huang2022registration} & \ccross & 81.54 & 96.34 & 68.59 & 90.53 & \bf 100.00 & 99.76 & 98.50 & 95.94 & 95.90 & 93.76 & 70.27 & 74.69 & 83.70 & 66.41 & 68.97 & 64.33 & 63.05 & 57.31
\\
AnomalyGPT~\cite{gu2024agpt} & \ccross & 85.95 & 98.69 & 74.79 & 98.93 & \bf 100.00 & 97.94 & 95.96 & 99.60 & 99.39 & 99.07 & 88.06 & 86.19 & 70.69 & 92.65 & 86.82 & 61.47 & 51.20 & 61.28
\\
LTAD~\cite{ho2024ltad} & \ccross & 87.36 & 99.33 & 76.89 & 99.79 & 99.93 & 99.76 & 99.12 & 99.96 & 99.17 & 97.61 & 78.89 & 86.50 & 87.17 & 90.48 & 74.96 & 66.61 & 76.31 & 54.21
\\
HVQ~\cite{lu2023hierarchical} & \ccross & 82.14 & 99.30 & 67.12 & 99.79 & \bf 100.00 & 99.76 & 97.54 & 99.60 & \bf 99.86 & 98.53 & 83.33 & 80.86 & 69.00 & 81.04 & 68.60 & 44.52 & 66.12 & 43.49
\\
\end{tabular}
}

\caption{
    \textbf{Comparison ($\uparrow$) on MVTec \it{step100}}~\cite{ho2024ltad} in image-level AUROC using the same evaluation metrics as Tab.~\ref{table:mvtec-exp100-det}.
}
\label{table:mvtec-step100-det}

\end{table*}

\begin{table*}[ht]

\centering
\setlength{\tabcolsep}{3pt}

\resizebox{\linewidth}{!}{%
\begin{tabular}[h]{*{100}{c}}
\specialrule{.15em}{.05em}{.05em}
{Method} & {CA} &
{\it $\gC$} &
{\color{darkblue}$\gC^{\tt head}$} &
{\color{darkred}$\gC^{\tt tail}$} &
{\color{darkblue}\it hazelnut} &
{\color{darkblue}\it leather} &
{\color{darkblue}\it bottle} &
{\color{darkblue}\it wood} &
{\color{darkblue}\it carpet} &
{\color{darkblue}\it tile} &
{\color{darkblue}\it metal nut} &
{\color{darkred}\it toothbrush} &
{\color{darkred}\it zipper} &
{\color{darkred}\it transistor} &
{\color{darkred}\it grid} &
{\color{darkred}\it pill} &
{\color{darkred}\it capsule} &
{\color{darkred}\it cable} &
{\color{darkred}{\it screw}}
\\
\cmidrule(lr){1-38}
MKD~\cite{salehi2021multiresolution} & \ccross & 85.90 & 86.92 & 85.01 & 94.61 & 96.24 & 92.63 & 77.85 & 93.20 & 76.96 & 77.01 & 93.87 & 90.92 & 73.73 & 72.96 & 89.09 & 93.41 & 72.46 & 93.60
\\
DRAEM~\cite{zavrtanik2021draem} & \ccross & 79.65 & 92.81 & 69.79 & 94.54 & 97.87 & 89.75 & 95.11 & 98.08 & \bf 98.86 & 80.53 & 90.82 & 74.91 & 60.30 & 50.84 & 76.96 & 65.39 & 53.87 & 85.26
\\
RegAD~\cite{huang2022registration} & \ccross & 95.10 & \bf 97.25 & 93.21 & 98.24 & 99.22 & 97.92 & \bf 95.87 & 98.37 & 94.37 & \bf 96.78 & 95.69 & \bf 95.71 & 93.16 & 79.73 & \bf 96.53 & 95.91 & \bf 94.54 & \bf 94.48
\\
AnomalyGPT~\cite{gu2024agpt} & \ccross & 89.28 & 93.62 & 85.49 & 88.41 & \bf 99.56 & 94.05 & 89.61 & \bf 99.08 & 95.52 & 89.09 & 95.53 & 92.65 & 65.87 & 94.42 & 79.26 & 86.48 & 83.11 & 86.62
\\
LTAD~\cite{ho2024ltad} & \ccross & 93.83 & 95.90 & 92.01 & 98.24 & 99.01 & 95.69 & 91.46 & 98.68 & 91.82 & 96.43 & 96.71 & 90.17 & 93.68 & 91.90 & 91.10 & 95.73 & 91.80 & 85.02
\\
\end{tabular}
}

\caption{
    \textbf{Comparison ($\uparrow$) on MVTec \it{step100}}~\cite{ho2024ltad} in pixel-level AUROC using the same evaluation metrics as Tab.~\ref{table:mvtec-exp100-seg}.
}
\label{table:mvtec-step100-seg}

\end{table*}

\begin{table*}[ht]

\centering
\setlength{\tabcolsep}{3pt}

\resizebox{\linewidth}{!}{%
\begin{tabular}[h]{*{100}{c}}
\specialrule{.15em}{.05em}{.05em}
{Method} & {CA} &
{\it $\gC$} &
{\color{darkblue}$\gC^{\tt head}$} &
{\color{darkred}$\gC^{\tt tail}$} &
{\color{darkblue}\it hazelnut} &
{\color{darkblue}\it leather} &
{\color{darkblue}\it bottle} &
{\color{darkblue}\it wood} &
{\color{darkblue}\it carpet} &
{\color{darkblue}\it tile} &
{\color{darkblue}\it metal nut} &
{\color{darkred}\it toothbrush} &
{\color{darkred}\it zipper} &
{\color{darkred}\it transistor} &
{\color{darkred}\it grid} &
{\color{darkred}\it pill} &
{\color{darkred}\it capsule} &
{\color{darkred}\it cable} &
{\color{darkred}{\it screw}}
\\
\cmidrule(lr){1-38}
MKD~\cite{salehi2021multiresolution} & \ccross & 79.31 & 84.48 & 74.79 & 97.04 & 80.03 & 99.05 & 91.58 & 75.52 & 85.21 & 62.90 & 72.78 & \bf 90.70 & 77.67 & 64.75 & 76.38 & \bf 69.45 & 78.94 & 67.70
\\
DRAEM~\cite{zavrtanik2021draem} & \ccross & 71.64 & 90.90 & 54.79 & 96.96 & 96.39 & 86.50 & 98.24 & 94.98 & 97.58 & 65.68 & 64.44 & 46.90 & 39.29 & 57.81 & 63.39 & 40.76 & 51.55 & \bf 74.21
\\
AnomalyGPT~\cite{gu2024agpt} & \ccross & 82.47 & 98.07 & 68.82 & 98.82 & \bf 100.00 & 97.30 & 92.81 & 99.80 & 98.45 & 99.32 & 73.89 & 88.46 & 64.00 & 92.77 & 84.79 & 48.64 & 50.67 & 47.39
\\
LTAD~\cite{ho2024ltad} & \ccross & 85.60 & 99.15 & 73.74 & 99.43 & 99.93 & 99.68 & \bf 99.39 & 99.80 & 99.21 & 96.58 & 75.00 & 85.90 & 81.71 & 90.48 & 76.68 & 63.98 & 65.54 & 50.65
\\
HVQ~\cite{lu2023hierarchical} & \ccross & 83.00 & 99.34 & 68.70 & 99.96 & \bf 100.00 & \bf 100.00 & 97.02 & 99.40 & 99.24 & 99.76 & \bf 93.33 & 86.87 & 72.21 & 68.50 & 75.10 & 52.61 & 54.55 & 46.38
\\
\end{tabular}
}

\caption{
    \textbf{Comparison ($\uparrow$) on MVTec \it{step200}}~\cite{ho2024ltad} in image-level AUROC using the same evaluation metrics as Tab.~\ref{table:mvtec-exp100-det}.
}
\label{table:mvtec-step200-det}

\end{table*}

\begin{table*}[ht]

\centering
\setlength{\tabcolsep}{3pt}

\resizebox{\linewidth}{!}{%
\begin{tabular}[h]{*{100}{c}}
\specialrule{.15em}{.05em}{.05em}
{Method} & {CA} &
{\it $\gC$} &
{\color{darkblue}$\gC^{\tt head}$} &
{\color{darkred}$\gC^{\tt tail}$} &
{\color{darkblue}\it hazelnut} &
{\color{darkblue}\it leather} &
{\color{darkblue}\it bottle} &
{\color{darkblue}\it wood} &
{\color{darkblue}\it carpet} &
{\color{darkblue}\it tile} &
{\color{darkblue}\it metal nut} &
{\color{darkred}\it toothbrush} &
{\color{darkred}\it zipper} &
{\color{darkred}\it transistor} &
{\color{darkred}\it grid} &
{\color{darkred}\it pill} &
{\color{darkred}\it capsule} &
{\color{darkred}\it cable} &
{\color{darkred}{\it screw}}
\\
\cmidrule(lr){1-38}
MKD~\cite{salehi2021multiresolution} & \ccross & 86.03 & 87.22 & 84.99 & 94.80 & 96.73 & 93.34 & 78.61 & 93.33 & 76.63 & 77.16 & 93.93 & 90.73 & 73.95 & 74.24 & 88.43 & 93.35 & 71.44 & \bf 93.86
\\
DRAEM~\cite{zavrtanik2021draem} & \ccross & 76.79 & 84.23 & 70.29 & 95.21 & 88.39 & 76.81 & 82.64 & 94.62 & 89.27 & 62.70 & 89.65 & 70.96 & 56.29 & 57.39 & 85.65 & 54.66 & 56.30 & 91.44
\\
AnomalyGPT~\cite{gu2024agpt} & \ccross & 89.45 & 93.75 & 85.69 & 88.29 & \bf 99.56 & 95.12 & 89.94 & \bf 99.00 & \bf 95.49 & 88.82 & 94.98 & 91.61 & 66.40 & \bf 93.55 & 78.70 & 90.56 & 82.42 & 87.33
\\
LTAD~\cite{ho2024ltad} & \ccross & 92.12 & 95.45 & 89.20 & 97.83 & 99.06 & 95.76 & 91.00 & 98.72 & 90.59 & 95.22 & 96.45 & 89.00 & 84.38 & 89.24 & 89.94 & 92.37 & 84.92 & 87.32
\\
HVQ~\cite{lu2023hierarchical} & \ccross & 92.61 & \bf 96.42 & 89.28 & \bf 98.66 & 98.86 & \bf 98.24 & 92.21 & 98.58 & 91.75 & \bf 96.62 & 98.31 & 91.54 & 83.37 & 84.53 & 88.15 & 95.09 & 88.13 & 85.11
\\
\end{tabular}
}

\caption{
    \textbf{Comparison ($\uparrow$) on MVTec \it{step200}}~\cite{ho2024ltad} in pixel-level AUROC using the same evaluation metrics as Tab.~\ref{table:mvtec-exp100-seg}.
}
\label{table:mvtec-step200-seg}

\end{table*}

\begin{table*}[ht]

\centering
\setlength{\tabcolsep}{3pt}

\resizebox{\linewidth}{!}{%
\begin{tabular}[h]{*{100}{c}}
\specialrule{.15em}{.05em}{.05em}
{Method} & {CA} &
{\it $\gC$} &
{\color{darkblue}$\gC^{\tt head}$} &
{\color{darkred}$\gC^{\tt tail}$} &
{\color{darkblue}\it PCB3} &
{\color{darkblue}\it PCB2} &
{\color{darkblue}\it PCB1} &
{\color{darkblue}\it PCB4} &
{\color{darkblue}\it macaroni1} &
{\color{darkblue}\it macaroni2} &
{\color{darkred}\it capsule} &
{\color{darkred}\it candle} &
{\color{darkred}\it fryum} &
{\color{darkred}\it capsules} &
{\color{darkred}\it chewinggum} &
{\color{darkred}\it pipe fryum} 
\\
\cmidrule(lr){1-38}
RegAD~\cite{huang2022registration} & \ccross & 71.36 & 66.39 & 76.34 & 62.71 & 63.79 & 73.89 & 82.82 & 62.37 & 52.79 & 76.12 & 71.99 & 73.82 & 56.07 & 92.92 & 87.13
\\
AnomalyGPT~\cite{gu2024agpt} & \ccross & 70.34 & 70.13 & 70.56 & 62.69 & 78.55 & 79.57 & 69.66 & \bf 78.76 & 51.54 & 49.96 & 77.15 & 80.77 & 63.88 & 71.76 & 79.84
\\
LTAD~\cite{ho2024ltad} & \ccross & 80.00 & 76.29 & 83.71 & 82.18 & 81.42 & 85.02 & 93.63 & 56.46 & 59.02 & 84.46 & 82.46 & 85.70 & 69.37 & 96.16 & 84.10
\\
HVQ~\cite{lu2023hierarchical} & \ccross & 78.63 & \bf 81.75 & 75.51 & \bf 83.03 & \bf 92.40 & \bf 85.08 & \bf 98.36 & 75.28 & 56.36 & \bf 93.38 & 61.10 & 79.50 & 63.77 & 86.64 & 68.70
\\
\cmidrule(lr){1-38}
UniAD~\cite{you2022unified} & \ccheck & 77.31 & 77.42 & 77.19 & 78.34 & 84.69 & 82.02 & 95.96 & 70.68 & 52.88 & 87.10 & 81.44 & 68.60 & 52.75 & 95.54 & 77.74
\\
MoEAD~\cite{mengmoead} & \ccheck & 73.89 & 72.37 & 75.42 & 71.75 & 76.87 & 84.26 & 86.87 & 59.18 & 55.27 & 89.08 & 71.06 & 76.40 & 70.17 & 81.72 & 64.08
\\
\rowcolor{OursColor} {\ours} & \ccheck & \bf 85.26 & 79.64 & \bf 90.87 & 80.84 & 81.15 & 84.28 & 96.42 & 70.71 & \bf 64.43 & 91.17 & \bf 92.96 & \bf 86.88 & \bf 79.07 & \bf 97.98 & \bf 97.18
\\
\specialrule{.15em}{.05em}{.05em}
\end{tabular}
}

\caption{
    \textbf{Comparison ($\uparrow$) on VisA \it{exp100}}~\cite{ho2024ltad} in image-level AUROC using the same evaluation metrics as Tab~\ref{table:mvtec-exp100-det}. 
}
\label{table:visa-exp100-det}

\end{table*}

\begin{table*}[ht]

\centering
\setlength{\tabcolsep}{3pt}

\resizebox{\linewidth}{!}{%
\begin{tabular}[h]{*{100}{c}}
\specialrule{.15em}{.05em}{.05em}
{Method} & {CA} &
{\it $\gC$} &
{\color{darkblue}$\gC^{\tt head}$} &
{\color{darkred}$\gC^{\tt tail}$} &
{\color{darkblue}\it PCB3} &
{\color{darkblue}\it PCB2} &
{\color{darkblue}\it PCB1} &
{\color{darkblue}\it PCB4} &
{\color{darkblue}\it macaroni1} &
{\color{darkblue}\it macaroni2} &
{\color{darkred}\it capsule} &
{\color{darkred}\it candle} &
{\color{darkred}\it fryum} &
{\color{darkred}\it capsules} &
{\color{darkred}\it chewinggum} &
{\color{darkred}\it pipe fryum} 
\\
\cmidrule(lr){1-38}
RegAD~\cite{huang2022registration} & \ccross & 94.40 & 93.72 & 95.07 & 95.27 & 92.58 & \bf 97.26 & 91.82 & 94.88 & 90.53 & 95.03 & 96.82 & 94.97 & 87.20 & 97.37 & 99.07
\\
AnomalyGPT~\cite{gu2024agpt} & \ccross & 80.32 & 77.43 & 83.20 & 79.41 & 84.80 & 77.09 & 82.17 & 70.68 & 70.44 & 77.41 & 86.70 & 90.21 & 62.72 & 92.10 & 90.07
\\
LTAD~\cite{ho2024ltad} & \ccross & 95.56 & 95.16 & 95.97 & \bf 97.62 & 96.06 & 92.18 & 95.96 & 96.58 & 92.58 & 97.90 & 99.08 & 95.39 & 85.77 & 98.82 & 98.83
\\
HVQ~\cite{lu2023hierarchical} & \ccross & 96.18 & \bf 95.99 & 96.37 & 97.52 & \bf 97.21 & 91.59 & 96.65 & \bf 98.01 & 94.96 & \bf 98.54 & 98.16 & 96.14 & 90.66 & 95.68 & 99.05
\\
\cmidrule(lr){1-38}
UniAD~\cite{you2022unified} & \ccheck & 95.03 & 94.57 & 95.48 & 97.05 & 96.12 & 91.61 & \bf 96.80 & 97.04 & 88.84 & 98.19 & 98.82 & 95.25 & 83.21 & 98.68 & 98.78
\\
MoEAD~\cite{mengmoead} & \ccheck & 94.34 & 93.36 & 95.31 & 95.73 & 96.25 & 91.52 & 94.98 & 92.93 & 88.72 & 97.53 & 98.98 & 95.10 & 83.05 & 97.86 & \bf 99.37
\\
\rowcolor{OursColor} {\ours} & \ccheck & \bf 96.91 & 95.84 & \bf 97.97 & 97.43 & 96.45 & 93.37 & 96.50 & 96.16 & \bf 95.13 & 98.40 & \bf 99.16 & \bf 96.67 & \bf 95.40 & \bf 98.87 & 99.33
\\
\specialrule{.15em}{.05em}{.05em}
\end{tabular}
}

\caption{
    \textbf{Comparison ($\uparrow$) on VisA \it{exp100}}~\cite{ho2024ltad} in pixel-level AUROC using the same evaluation metrics as Tab~\ref{table:mvtec-exp100-seg}. 
}
\label{table:visa-exp100-seg}

\end{table*}

\begin{table*}[ht]

\centering
\setlength{\tabcolsep}{3pt}

\resizebox{\linewidth}{!}{%
\begin{tabular}[h]{*{100}{c}}
\specialrule{.15em}{.05em}{.05em}
{Method} & {CA} &
{\it $\gC$} &
{\color{darkblue}$\gC^{\tt head}$} &
{\color{darkred}$\gC^{\tt tail}$} &
{\color{darkblue}\it PCB3} &
{\color{darkblue}\it PCB2} &
{\color{darkblue}\it PCB1} &
{\color{darkblue}\it PCB4} &
{\color{darkblue}\it macaroni1} &
{\color{darkblue}\it macaroni2} &
{\color{darkred}\it capsule} &
{\color{darkred}\it candle} &
{\color{darkred}\it fryum} &
{\color{darkred}\it capsules} &
{\color{darkred}\it chewinggum} &
{\color{darkred}\it pipe fryum} 
\\
\cmidrule(lr){1-38}
RegAD~\cite{huang2022registration} & \ccross & 72.10 & 67.19 & 77.02 & 61.92 & 64.41 & 73.16 & 79.95 & 67.55 & 56.16 & 76.53 & 77.21 & 79.73 & 55.81 & 88.92 & 83.94
\\
AnomalyGPT~\cite{gu2024agpt} & \ccross & 69.78 & 69.22 & 70.35 & 61.64 & 75.82 & 79.42 & 68.41 & \bf 79.60 & 50.40 & 45.90 & 76.59 & 79.80 & 63.34 & 72.83 & 83.63
\\
LTAD~\cite{ho2024ltad} & \ccross & 80.21 & 76.81 & 83.60 & \bf 84.98 & 83.77 & \bf 86.86 & 95.37 & 57.66 & 52.20 & 85.73 & 81.82 & 83.30 & 63.23 & 96.82 & 90.72
\\
HVQ~\cite{lu2023hierarchical} & \ccross & 78.03 & 79.35 & 76.71 & 78.50 & \bf 88.97 & 82.56 & \bf 98.07 & 64.13 & \bf 63.84 & \bf 92.95 & 63.76 & 83.30 & 63.85 & 87.04 & 69.36
\\
\cmidrule(lr){1-38}
UniAD~\cite{you2022unified} & \ccheck & 76.87 & 76.25 & 77.49 & 78.91 & 84.96 & 81.85 & 96.34 & 62.01 & 53.44 & 86.91 & 78.60 & 74.84 & 54.08 & 94.28 & 76.28
\\
MoEAD~\cite{mengmoead} & \ccheck & 73.58 & 73.36 & 73.80 & 74.45 & 77.77 & 84.22 & 91.63 & 55.66 & 56.44 & 88.07 & 73.94 & 79.30 & 65.67 & 71.76 & 64.06
\\
\rowcolor{OursColor} {\ours} & \ccheck & \bf 85.24 & \bf 80.14 & \bf 90.34 & 80.46 & 82.09 & 84.80 & 97.03 & 74.46 & 61.99 & 91.85 & \bf 93.10 & \bf 85.04 & \bf 75.52 & \bf 98.74 & \bf 97.82
\\
\specialrule{.15em}{.05em}{.05em}
\end{tabular}
}

\caption{
    \textbf{Comparison ($\uparrow$) on VisA \it{exp200}}~\cite{ho2024ltad} in image-level AUROC using the same evaluation metrics as Tab~\ref{table:mvtec-exp100-det}. 
}
\label{table:visa-exp200-det}

\end{table*}

\begin{table*}[ht]

\centering
\setlength{\tabcolsep}{3pt}

\resizebox{\linewidth}{!}{%
\begin{tabular}[h]{*{100}{c}}
\specialrule{.15em}{.05em}{.05em}
{Method} & {CA} &
{\it $\gC$} &
{\color{darkblue}$\gC^{\tt head}$} &
{\color{darkred}$\gC^{\tt tail}$} &
{\color{darkblue}\it PCB3} &
{\color{darkblue}\it PCB2} &
{\color{darkblue}\it PCB1} &
{\color{darkblue}\it PCB4} &
{\color{darkblue}\it macaroni1} &
{\color{darkblue}\it macaroni2} &
{\color{darkred}\it capsule} &
{\color{darkred}\it candle} &
{\color{darkred}\it fryum} &
{\color{darkred}\it capsules} &
{\color{darkred}\it chewinggum} &
{\color{darkred}\it pipe fryum} 
\\
\cmidrule(lr){1-38}
RegAD~\cite{huang2022registration} & \ccross & 94.69 & 94.07 & 95.30 & 95.78 & 94.22 & \bf 97.87 & 90.41 & 96.18 & 90.01 & 95.64 & 96.80 & 95.89 & 87.54 & 96.73 & 99.23
\\
AnomalyGPT~\cite{gu2024agpt} & \ccross & 79.48 & 76.64 & 82.32 & 77.59 & 83.31 & 76.89 & 81.92 & 68.29 & 71.84 & 77.66 & 83.39 & 89.57 & 61.44 & 92.37 & 89.46
\\
LTAD~\cite{ho2024ltad} & \ccross & 95.36 & 94.98 & 95.69 & \bf 97.66 & 95.03 & 91.56 & 96.42 & 96.20 & 89.58 & 98.31 & 99.02 & 95.16 & 83.74 & 98.87 & 99.02
\\
HVQ~\cite{lu2023hierarchical} & \ccross & 95.32 & 95.30 & 95.35 & 96.94 & 95.71 & 92.34 & 96.20 & \bf 97.19 & 93.40 & 98.20 & 97.64 & 95.50 & 86.60 & 95.27 & 98.88
\\
\cmidrule(lr){1-38}
UniAD~\cite{you2022unified} & \ccheck & 94.80 & 94.34 & 95.26 & 97.07 & 96.06 & 91.64 & 96.53 & 96.18 & 88.56 & 98.06 & 98.61 & 95.24 & 82.72 & 98.32 & 98.66
\\
MoEAD~\cite{mengmoead} & \ccheck & 93.98 & 93.52 & 94.44 & 95.91 & 96.51 & 91.89 & 95.40 & 92.45 & 88.95 & 97.78 & 98.92 & 95.62 & 79.04 & 96.41 & 98.89
\\
\rowcolor{OursColor} {\ours} & \ccheck & \bf 96.97 & \bf 96.08 & \bf 97.86 & 97.57 & \bf 96.64 & 93.88 & \bf 96.93 & 96.22 & \bf 95.26 & \bf 98.45 & \bf 99.30 & \bf 96.52 & \bf 94.75 & \bf 98.89 & \bf 99.26
\\
\specialrule{.15em}{.05em}{.05em}
\end{tabular}
}

\caption{
    \textbf{Comparison ($\uparrow$) on VisA \it{exp200}}~\cite{ho2024ltad} in pixel-level AUROC using the same evaluation metrics as Tab~\ref{table:mvtec-exp100-seg}. 
}
\label{table:visa-exp200-seg}

\end{table*}

\begin{table*}[ht]

\centering
\setlength{\tabcolsep}{3pt}

\resizebox{\linewidth}{!}{%
\begin{tabular}[h]{*{100}{c}}
\specialrule{.15em}{.05em}{.05em}
{Method} & {CA} &
{\it $\gC$} &
{\color{darkblue}$\gC^{\tt head}$} &
{\color{darkred}$\gC^{\tt tail}$} &
{\color{darkblue}\it PCB3} &
{\color{darkblue}\it PCB2} &
{\color{darkblue}\it PCB1} &
{\color{darkblue}\it PCB4} &
{\color{darkblue}\it macaroni1} &
{\color{darkblue}\it macaroni2} &
{\color{darkred}\it capsule} &
{\color{darkred}\it candle} &
{\color{darkred}\it fryum} &
{\color{darkred}\it capsules} &
{\color{darkred}\it chewinggum} &
{\color{darkred}\it pipe fryum} 
\\
\cmidrule(lr){1-38}
AnomalyGPT~\cite{gu2024agpt} & \ccross & 68.18 & 68.35 & 68.00 & 60.83 & 76.03 & 78.94 & 66.37 & \bf 77.50 & 50.42 & 51.00 & 74.71 & 72.72 & 61.73 & 73.85 & 74.00
\\
LTAD~\cite{ho2024ltad} & \ccross & 78.53 & 76.56 & 80.50 & \bf 84.22 & 83.85 & \bf 87.05 & 94.55 & 50.11 & 59.56 & 81.07 & 77.84 & 77.36 & 66.48 & 94.24 & 85.98
\\
HVQ~\cite{lu2023hierarchical} & \ccross & 74.96 & 73.89 & 76.04 & 70.34 & 79.46 & 85.41 & 93.28 & 58.97 & 55.87 & \bf 89.86 & 73.38 & 77.96 & 59.58 & 90.20 & 65.24
\\
\cmidrule(lr){1-38}
UniAD~\cite{you2022unified} & \ccheck & 73.67 & 76.61 & 70.73 & 79.17 & \bf 85.11 & 82.97 & 96.38 & 62.94 & 53.13 & 85.70 & 65.38 & 76.92 & 53.11 & 89.22 & 54.06
\\
MoEAD~\cite{mengmoead} & \ccheck & 70.93 & 72.68 & 69.18 & 73.67 & 75.50 & 83.28 & 85.91 & 58.69 & 59.04 & 84.23 & 66.08 & 77.70 & 63.85 & 70.22 & 52.98
\\
\rowcolor{OursColor} {\ours} & \ccheck & \bf 83.89 & \bf 80.01 & \bf 87.77 & 81.59 & 82.56 & 85.75 & \bf 96.76 & 71.49 & \bf 61.93 & 89.14 & \bf 91.52 & \bf 87.04 & \bf 70.70 & \bf 96.90 & \bf 91.34
\\
\specialrule{.15em}{.05em}{.05em}
\end{tabular}
}

\caption{
    \textbf{Comparison ($\uparrow$) on VisA \it{exp500}}~\cite{ho2024ltad} in image-level AUROC using the same evaluation metrics as Tab~\ref{table:mvtec-exp100-det}. 
}
\label{table:visa-exp500-det}

\end{table*}

\begin{table*}[ht]

\centering
\setlength{\tabcolsep}{3pt}

\resizebox{\linewidth}{!}{%
\begin{tabular}[h]{*{100}{c}}
\specialrule{.15em}{.05em}{.05em}
{Method} & {CA} &
{\it $\gC$} &
{\color{darkblue}$\gC^{\tt head}$} &
{\color{darkred}$\gC^{\tt tail}$} &
{\color{darkblue}\it PCB3} &
{\color{darkblue}\it PCB2} &
{\color{darkblue}\it PCB1} &
{\color{darkblue}\it PCB4} &
{\color{darkblue}\it macaroni1} &
{\color{darkblue}\it macaroni2} &
{\color{darkred}\it capsule} &
{\color{darkred}\it candle} &
{\color{darkred}\it fryum} &
{\color{darkred}\it capsules} &
{\color{darkred}\it chewinggum} &
{\color{darkred}\it pipe fryum} 
\\
\cmidrule(lr){1-38}
AnomalyGPT~\cite{gu2024agpt} & \ccross & 78.83 & 76.04 & 81.62 & 78.63 & 84.05 & 75.66 & 83.38 & 65.46 & 69.04 & 77.46 & 81.01 & 89.62 & 58.23 & 93.03 & 90.36
\\
LTAD~\cite{ho2024ltad} & \ccross & 94.66 & 94.42 & 94.90 & \bf 97.75 & 95.60 & 92.76 & 96.23 & 94.80 & 89.39 & 97.54 & 98.86 & 95.16 & 80.98 & 98.69 & 98.18
\\
HVQ~\cite{lu2023hierarchical} & \ccross & 95.77 & 95.26 & 96.28 & 96.46 & 96.31 & \bf 94.94 & 95.69 & 94.87 & 93.30 & \bf 98.17 & 98.53 & \bf 96.10 & 89.62 & 97.44 & 97.81
\\
\cmidrule(lr){1-38}
UniAD~\cite{you2022unified} & \ccheck & 94.35 & 94.35 & 94.35 & 97.01 & 96.21 & 91.49 & 96.56 & \bf 95.92 & 88.95 & 97.96 & 98.35 & 94.94 & 81.48 & 97.76 & 95.62
\\
MoEAD~\cite{mengmoead} & \ccheck & 93.39 & 93.06 & 93.71 & 95.73 & 96.15 & 91.40 & 94.81 & 91.14 & 89.16 & 97.40 & 98.68 & 94.93 & 79.83 & 94.72 & 96.71
\\
\rowcolor{OursColor} {\ours} & \ccheck & \bf 96.60 & \bf 95.78 & \bf 97.41 & 97.42 & \bf 96.39 & 93.17 & \bf 96.81 & 95.86 & \bf 95.05 & 98.16 & \bf 99.09 & 96.06 & \bf 93.42 & \bf 98.82 & \bf 98.91
\\
\specialrule{.15em}{.05em}{.05em}
\end{tabular}
}

\caption{
    \textbf{Comparison ($\uparrow$) on VisA \it{exp500}}~\cite{ho2024ltad} in pixel-level AUROC using the same evaluation metrics as Tab~\ref{table:mvtec-exp100-seg}. 
}
\label{table:visa-exp500-seg}

\end{table*}

\begin{table*}[ht]

\centering
\setlength{\tabcolsep}{3pt}

\resizebox{\linewidth}{!}{%
\begin{tabular}[h]{*{100}{c}}
\specialrule{.15em}{.05em}{.05em}
{Method} & {CA} &
{\it $\gC$} &
{\color{darkblue}$\gC^{\tt head}$} &
{\color{darkred}$\gC^{\tt tail}$} &
{\color{darkblue}\it PCB3} &
{\color{darkblue}\it PCB2} &
{\color{darkblue}\it PCB1} &
{\color{darkblue}\it PCB4} &
{\color{darkblue}\it macaroni1} &
{\color{darkblue}\it macaroni2} &
{\color{darkred}\it capsule} &
{\color{darkred}\it candle} &
{\color{darkred}\it fryum} &
{\color{darkred}\it capsules} &
{\color{darkred}\it chewinggum} &
{\color{darkred}\it pipe fryum} 
\\
\cmidrule(lr){1-38}
RegAD~\cite{huang2022registration} & \ccross & 71.80 & 66.71 & 76.89 & 61.04 & 65.64 & 72.92 & 81.27 & 66.91 & 52.48 & 74.35 & 73.46 & 71.42 & 57.14 & 91.51 & 93.49
\\
AnomalyGPT~\cite{gu2024agpt} & \ccross & 71.89 & 71.41 & 72.37 & 65.90 & 75.90 & 74.12 & 75.06 & 82.88 & 54.61 & 50.73 & 80.22 & \bf 94.65 & 57.10 & 76.24 & 75.26
\\
LTAD~\cite{ho2024ltad} & \ccross & 84.80 & 85.98 & 83.63 & 85.21 & 82.08 & 87.91 & 96.97 & \bf 86.17 & 77.54 & 81.94 & 83.94 & 78.78 & 69.55 & 97.10 & 90.44
\\
HVQ~\cite{lu2023hierarchical} & \ccross & 76.68 & 87.24 & 66.12 & \bf 85.43 & \bf 91.02 & 92.17 & 99.02 & 83.15 & 72.63 & 64.42 & 53.44 & 76.22 & 58.80 & 73.76 & 70.06
\\
\cmidrule(lr){1-38}
UniAD~\cite{you2022unified} & \ccheck & 78.83 & 83.06 & 74.61 & 79.91 & 85.51 & 88.36 & 97.41 & 82.31 & 64.89 & 77.53 & 60.90 & 77.34 & 56.80 & 93.44 & 81.66
\\
MoEAD~\cite{mengmoead} & \ccheck & 74.12 & 86.58 & 61.67 & 81.54 & 88.64 & \bf 92.59 & 99.03 & 84.86 & 72.80 & 62.97 & 55.96 & 65.96 & 48.30 & 74.72 & 62.10
\\
\rowcolor{OursColor} {\ours} & \ccheck & \bf 88.32 & \bf 87.39 & \bf 89.25 & 85.21 & 87.04 & 90.80 & \bf 99.20 & 83.46 & \bf 78.63 & \bf 84.99 & \bf 91.56 & 86.88 & \bf 75.92 & \bf 98.72 & \bf 97.44
\\
\specialrule{.15em}{.05em}{.05em}
\end{tabular}
}

\caption{
    \textbf{Comparison ($\uparrow$) on VisA \it{step100}}~\cite{ho2024ltad} in image-level AUROC using the same evaluation metrics as Tab~\ref{table:mvtec-exp100-det}. 
}
\label{table:visa-step100-det}

\end{table*}

\begin{table*}[ht]

\centering
\setlength{\tabcolsep}{3pt}

\resizebox{\linewidth}{!}{%
\begin{tabular}[h]{*{100}{c}}
\specialrule{.15em}{.05em}{.05em}
{Method} & {CA} &
{\it $\gC$} &
{\color{darkblue}$\gC^{\tt head}$} &
{\color{darkred}$\gC^{\tt tail}$} &
{\color{darkblue}\it PCB3} &
{\color{darkblue}\it PCB2} &
{\color{darkblue}\it PCB1} &
{\color{darkblue}\it PCB4} &
{\color{darkblue}\it macaroni1} &
{\color{darkblue}\it macaroni2} &
{\color{darkred}\it capsule} &
{\color{darkred}\it candle} &
{\color{darkred}\it fryum} &
{\color{darkred}\it capsules} &
{\color{darkred}\it chewinggum} &
{\color{darkred}\it pipe fryum} 
\\
\cmidrule(lr){1-38}
RegAD~\cite{huang2022registration} & \ccross & 94.99 & 94.81 & 95.16 & 95.09 & 94.76 & 97.76 & 93.17 & 95.63 & 92.49 & 95.72 & 97.49 & 95.23 & 84.83 & 98.61 & 99.10
\\
AnomalyGPT~\cite{gu2024agpt} & \ccross & 82.30 & 79.41 & 85.19 & 80.35 & 86.10 & 78.21 & 80.42 & 76.54 & 74.83 & 81.56 & 91.79 & 91.02 & 60.06 & 93.07 & 93.63
\\
LTAD~\cite{ho2024ltad} & \ccross & 96.57 & 97.43 & 95.72 & 97.67 & 96.10 & 98.80 & 97.02 & 98.10 & 96.87 & \bf 96.57 & 98.97 & 95.78 & 84.87 & \bf 98.93 & 99.20
\\
HVQ~\cite{lu2023hierarchical} & \ccross & 95.63 & 97.69 & 93.57 & 97.56 & \bf 97.44 & 98.58 & 96.79 & \bf 98.40 & 97.40 & 89.61 & 97.35 & 94.26 & 87.80 & 93.57 & 98.84
\\
\cmidrule(lr){1-38}
UniAD~\cite{you2022unified} & \ccheck & 96.04 & 97.37 & 94.71 & 97.25 & 96.91 & \bf 99.18 & 97.42 & 98.21 & 95.30 & 95.89 & 96.89 & 94.24 & 83.91 & 98.51 & 98.84
\\
MoEAD~\cite{mengmoead} & \ccheck & 94.37 & 97.51 & 91.23 & 97.54 & 97.29 & 99.04 & 97.04 & \bf 98.40 & 95.75 & 89.52 & 96.43 & 93.91 & 75.33 & 93.66 & 98.55
\\
\rowcolor{OursColor} {\ours} & \ccheck & \bf 97.54 & \bf 97.78 & \bf 97.30 & \bf 97.74 & 96.89 & 99.11 & \bf 97.76 & 97.48 & \bf 97.67 & 96.23 & \bf 99.11 & \bf 96.10 & \bf 94.28 & 98.82 & \bf 99.25
\\
\specialrule{.15em}{.05em}{.05em}
\end{tabular}
}

\caption{
    \textbf{Comparison ($\uparrow$) on VisA \it{step100}}~\cite{ho2024ltad} in pixel-level AUROC using the same evaluation metrics as Tab~\ref{table:mvtec-exp100-seg}. 
}
\label{table:visa-step100-seg}

\end{table*}

\begin{table*}[ht]

\centering
\setlength{\tabcolsep}{3pt}

\resizebox{\linewidth}{!}{%
\begin{tabular}[h]{*{100}{c}}
\specialrule{.15em}{.05em}{.05em}
{Method} & {CA} &
{\it $\gC$} &
{\color{darkblue}$\gC^{\tt head}$} &
{\color{darkred}$\gC^{\tt tail}$} &
{\color{darkblue}\it PCB3} &
{\color{darkblue}\it PCB2} &
{\color{darkblue}\it PCB1} &
{\color{darkblue}\it PCB4} &
{\color{darkblue}\it macaroni1} &
{\color{darkblue}\it macaroni2} &
{\color{darkred}\it capsule} &
{\color{darkred}\it candle} &
{\color{darkred}\it fryum} &
{\color{darkred}\it capsules} &
{\color{darkred}\it chewinggum} &
{\color{darkred}\it pipe fryum} 
\\
\cmidrule(lr){1-38}
RegAD~\cite{huang2022registration} & \ccross & 71.65 & 67.02 & 76.28 & 57.60 & 65.20 & 72.95 & 81.89 & 67.74 & 56.76 & 72.69 & 72.51 & 73.50 & 56.74 & 89.39 & 92.86
\\
AnomalyGPT~\cite{gu2024agpt} & \ccross & 69.78 & 71.19 & 68.38 & 66.79 & 75.20 & 74.30 & 73.57 & 81.94 & 55.34 & 44.74 & 82.28 & 74.26 & 56.81 & 77.68 & 74.49
\\
LTAD~\cite{ho2024ltad} & \ccross & 84.03 & 87.34 & 80.72 & 85.00 & 82.47 & 90.49 & 96.12 & \bf 88.64 & \bf 81.33 & 81.65 & 84.50 & 71.84 & 61.10 & 95.40 & 89.82
\\
HVQ~\cite{lu2023hierarchical} & \ccross & 75.52 & \bf 88.66 & 62.38 & \bf 86.09 & \bf 92.04 & 92.99 & \bf 99.32 & 84.42 & 77.09 & 50.99 & 63.18 & 69.14 & 47.13 & 79.34 & 64.52
\\
\cmidrule(lr){1-38}
UniAD~\cite{you2022unified} & \ccheck & 77.64 & 84.51 & 70.78 & 81.51 & 86.73 & 89.78 & 97.83 & 83.51 & 67.72 & 64.57 & 55.58 & 73.70 & 56.20 & 91.30 & 83.36
\\
MoEAD~\cite{mengmoead} & \ccheck & 76.77 & 87.84 & 65.71 & 82.88 & 88.74 & \bf 93.24 & 99.25 & 84.96 & 77.96 & 69.15 & 55.90 & 65.86 & 58.28 & 77.46 & 67.58
\\
\rowcolor{OursColor} {\ours} & \ccheck & \bf 87.04 & 87.77 & \bf 86.31 & 85.80 & 86.74 & 91.24 & 99.03 & 84.68 & 79.11 & \bf 83.98 & \bf 88.78 & \bf 83.18 & \bf 67.02 & \bf 97.38 & \bf 97.50
\\
\specialrule{.15em}{.05em}{.05em}
\end{tabular}
}

\caption{
    \textbf{Comparison ($\uparrow$) on VisA \it{step200}}~\cite{ho2024ltad} in image-level AUROC using the same evaluation metrics as Tab~\ref{table:mvtec-exp100-det}. 
}
\label{table:visa-step200-det}

\end{table*}

\begin{table*}[ht]

\centering
\setlength{\tabcolsep}{3pt}

\resizebox{\linewidth}{!}{%
\begin{tabular}[h]{*{100}{c}}
\specialrule{.15em}{.05em}{.05em}
{Method} & {CA} &
{\it $\gC$} &
{\color{darkblue}$\gC^{\tt head}$} &
{\color{darkred}$\gC^{\tt tail}$} &
{\color{darkblue}\it PCB3} &
{\color{darkblue}\it PCB2} &
{\color{darkblue}\it PCB1} &
{\color{darkblue}\it PCB4} &
{\color{darkblue}\it macaroni1} &
{\color{darkblue}\it macaroni2} &
{\color{darkred}\it capsule} &
{\color{darkred}\it candle} &
{\color{darkred}\it fryum} &
{\color{darkred}\it capsules} &
{\color{darkred}\it chewinggum} &
{\color{darkred}\it pipe fryum} 
\\
\cmidrule(lr){1-38}
RegAD~\cite{huang2022registration} & \ccross & 94.52 & 93.73 & 95.32 & 94.80 & 93.28 & 97.06 & 90.95 & 94.87 & 91.44 & 95.24 & 98.00 & 95.30 & 87.24 & 97.00 & 99.14
\\
AnomalyGPT~\cite{gu2024agpt} & \ccross & 81.97 & 78.69 & 85.25 & 80.22 & 85.88 & 77.08 & 81.05 & 74.01 & 73.90 & 80.96 & 91.60 & 90.77 & 61.16 & 93.04 & 93.99
\\
LTAD~\cite{ho2024ltad} & \ccross & 96.27 & 97.63 & 94.92 & 97.78 & 96.36 & 98.89 & 96.90 & 98.45 & 97.39 & 95.87 & 98.65 & 95.23 & 81.87 & \bf 98.82 & 99.06
\\
HVQ~\cite{lu2023hierarchical} & \ccross & 95.50 & \bf 97.89 & 93.10 & 97.73 & 97.50 & \bf 99.34 & 97.00 & 98.19 & \bf 97.57 & 90.31 & 96.45 & 95.05 & 82.89 & 95.17 & 98.76
\\
\cmidrule(lr){1-38}
UniAD~\cite{you2022unified} & \ccheck & 95.66 & 97.45 & 93.87 & 97.41 & 96.87 & 99.15 & 97.53 & 98.38 & 95.40 & 95.37 & 95.29 & 93.63 & 82.25 & 98.10 & 98.61
\\
MoEAD~\cite{mengmoead} & \ccheck & 94.82 & 97.80 & 91.84 & 97.77 & \bf 97.60 & 99.22 & 96.94 & \bf 98.64 & 96.65 & 90.07 & 96.37 & 95.28 & 77.05 & 93.76 & 98.49
\\
\rowcolor{OursColor} {\ours} & \ccheck & \bf 97.23 & 97.74 & \bf 96.71 & \bf 97.79 & 96.89 & 99.03 & \bf 97.64 & 97.73 & 97.37 & \bf 96.25 & \bf 98.88 & \bf 96.03 & \bf 91.12 & 98.74 & \bf 99.20
\\
\specialrule{.15em}{.05em}{.05em}
\end{tabular}
}

\caption{
    \textbf{Comparison ($\uparrow$) on VisA \it{step200}}~\cite{ho2024ltad} in pixel-level AUROC using the same evaluation metrics as Tab~\ref{table:mvtec-exp100-seg}. 
}
\label{table:visa-step200-seg}

\end{table*}

\begin{table*}[ht]

\centering
\setlength{\tabcolsep}{3pt}

\resizebox{\linewidth}{!}{%
\begin{tabular}[h]{*{100}{c}}
\specialrule{.15em}{.05em}{.05em}
{Method} & {CA} &
{\it $\gC$} &
{\color{darkblue}$\gC^{\tt head}$} &
{\color{darkred}$\gC^{\tt tail}$} &
{\color{darkblue}\it PCB3} &
{\color{darkblue}\it PCB2} &
{\color{darkblue}\it PCB1} &
{\color{darkblue}\it PCB4} &
{\color{darkblue}\it macaroni1} &
{\color{darkblue}\it macaroni2} &
{\color{darkred}\it capsule} &
{\color{darkred}\it candle} &
{\color{darkred}\it fryum} &
{\color{darkred}\it capsules} &
{\color{darkred}\it chewinggum} &
{\color{darkred}\it pipe fryum} 
\\
\cmidrule(lr){1-38}
AnomalyGPT~\cite{gu2024agpt} & \ccross & 62.88 & 70.15 & 55.61 & 63.36 & 74.22 & 74.40 & 75.84 & 79.34 & 53.74 & 41.40 & 71.25 & 66.18 & 49.11 & 41.74 & 63.95
\\
LTAD~\cite{ho2024ltad} & \ccross & 83.33 & 88.40 & 78.26 & 85.68 & 85.19 & 89.95 & 96.68 & 87.91 & \bf 85.00 & 78.59 & 76.56 & 69.86 & \bf 68.30 & \bf 94.34 & 81.88
\\
HVQ~\cite{lu2023hierarchical} & \ccross & 72.09 & 83.60 & 60.57 & 76.29 & 87.24 & 86.35 & 98.62 & 82.87 & 70.26 & 68.64 & 48.32 & 66.22 & 53.28 & 70.96 & 55.98
\\
\cmidrule(lr){1-38}
UniAD~\cite{you2022unified} & \ccheck & 71.84 & 81.84 & 61.85 & 78.79 & 84.72 & 88.80 & 97.44 & 79.46 & 61.83 & 61.82 & 68.54 & 55.96 & 51.30 & 68.28 & 65.22
\\
MoEAD~\cite{mengmoead} & \ccheck & 72.75 & \bf 89.41 & 56.09 & 83.37 & \bf 90.37 & \bf 93.80 & 99.03 & \bf 88.14 & 81.77 & 67.31 & 38.26 & 57.20 & 54.97 & 64.60 & 54.18
\\
\rowcolor{OursColor} {\ours} & \ccheck & \bf 84.84 & 88.13 & \bf 81.56 & \bf 85.99 & 87.51 & 91.67 & \bf 99.29 & 85.84 & 78.47 & \bf 87.26 & \bf 78.94 & \bf 76.72 & 61.03 & 93.16 & \bf 92.26
\\
\specialrule{.15em}{.05em}{.05em}
\end{tabular}
}

\caption{
    \textbf{Comparison ($\uparrow$) on VisA \it{step500}}~\cite{ho2024ltad} in image-level AUROC using the same evaluation metrics as Tab~\ref{table:mvtec-exp100-det}. 
}
\label{table:visa-step500-det}

\end{table*}

\begin{table*}[ht]

\centering
\setlength{\tabcolsep}{3pt}

\resizebox{\linewidth}{!}{%
\begin{tabular}[h]{*{100}{c}}
\specialrule{.15em}{.05em}{.05em}
{Method} & {CA} &
{\it $\gC$} &
{\color{darkblue}$\gC^{\tt head}$} &
{\color{darkred}$\gC^{\tt tail}$} &
{\color{darkblue}\it PCB3} &
{\color{darkblue}\it PCB2} &
{\color{darkblue}\it PCB1} &
{\color{darkblue}\it PCB4} &
{\color{darkblue}\it macaroni1} &
{\color{darkblue}\it macaroni2} &
{\color{darkred}\it capsule} &
{\color{darkred}\it candle} &
{\color{darkred}\it fryum} &
{\color{darkred}\it capsules} &
{\color{darkred}\it chewinggum} &
{\color{darkred}\it pipe fryum} 
\\
\cmidrule(lr){1-38}
AnomalyGPT~\cite{gu2024agpt} & \ccross & 81.48 & 78.88 & 84.09 & 79.88 & 84.91 & 78.51 & 79.59 & 75.69 & 74.69 & 78.52 & 92.32 & 90.94 & 59.75 & 89.52 & 93.49
\\
LTAD~\cite{ho2024ltad} & \ccross & \bf 96.41 & 97.69 & \bf 95.12 & 97.91 & 96.48 & 98.95 & 96.63 & 98.69 & \bf 97.50 & 95.81 & 97.90 & 95.10 & 84.46 & \bf 98.52 & \bf 98.93
\\
HVQ~\cite{lu2023hierarchical} & \ccross & 95.06 & 97.34 & 92.77 & 97.06 & 97.15 & 99.03 & 96.87 & 97.90 & 96.05 & 91.24 & 96.33 & \bf 95.54 & 82.58 & 93.91 & 97.03
\\
\cmidrule(lr){1-38}
UniAD~\cite{you2022unified} & \ccheck & 95.06 & 97.21 & 92.91 & 97.15 & 96.79 & \bf 99.16 & 97.35 & 97.98 & 94.87 & 93.25 & 91.47 & 90.63 & \bf 90.63 & 96.35 & 95.16
\\
MoEAD~\cite{mengmoead} & \ccheck & 94.30 & \bf 98.10 & 90.51 & \bf 97.95 & \bf 97.78 & \bf 99.16 & 97.57 & \bf 98.96 & 97.16 & 90.07 & 94.81 & 94.26 & 74.40 & 93.55 & 95.99
\\
\rowcolor{OursColor} {\ours} & \ccheck & \bf 96.41 & 97.74 & 95.08 & 97.63 & 96.90 & 99.07 & \bf 97.68 & 97.71 & 97.44 & \bf 96.94 & \bf 98.01 & 94.39 & 84.83 & 98.34 & 97.96
\\
\specialrule{.15em}{.05em}{.05em}
\end{tabular}
}

\caption{
    \textbf{Comparison ($\uparrow$) on VisA \it{step500}}~\cite{ho2024ltad} in pixel-level AUROC using the same evaluation metrics as Tab~\ref{table:mvtec-exp100-seg}. 
}
\label{table:visa-step500-seg}

\end{table*}

\begin{table*}[ht]

\centering
\setlength{\tabcolsep}{3pt}

\resizebox{0.8\linewidth}{!}{%
\begin{tabular}[h]{*{100}{c}}
\specialrule{.15em}{.05em}{.05em}
{Method} & {CA} &
{\it $\gC$} &
{\color{darkblue}$\gC^{\tt head}$} &
{\color{darkred}$\gC^{\tt tail}$} &
{\color{darkblue}\it Class10} &
{\color{darkblue}\it Class7} &
{\color{darkblue}\it Class9} &
{\color{darkblue}\it Class8} &
{\color{darkblue}\it Class2} &
{\color{darkred}\it Class3} &
{\color{darkred}\it Class5} &
{\color{darkred}\it Class1} &
{\color{darkred}\it Class4} &
{\color{darkred}\it Class6} &
\\
\cmidrule(lr){1-38}
RegAD~\cite{huang2022registration} & \ccross & 84.86 & 86.88 & 82.84 & 94.45 & 91.34 & 99.23 & 50.94 & 98.44 & 97.95 & 72.07 & 48.76 & 99.75 & 95.66
\\
AnomalyGPT~\cite{gu2024agpt} & \ccross & 84.86 & 84.68 & 85.04 & 82.55 & \bf 100.00 & 87.28 & 70.07 & 83.52 & 93.61 & 96.88 & 83.91 & 53.00 & 97.79
\\
LTAD~\cite{ho2024ltad} & \ccross & 94.82 & 93.39 & 96.25 & 97.71 & 99.53 & \bf 99.91 & 69.81 & \bf 100.00 & \bf 99.69 & 97.68 & 84.08 & \bf 100.00 & 99.82
\\
HVQ~\cite{lu2023hierarchical} & \ccross & 76.65 & 76.78 & 76.52 & 95.41 & 74.27 & 66.92 & 50.13 & 97.17 & 61.06 & 73.82 & \bf 99.42 & 94.66 & 53.64
\\
\cmidrule(lr){1-38}
UniAD~\cite{you2022unified} & \ccheck & 84.51 & 87.34 & 81.69 & 99.25 & 95.29 & 91.61 & 52.97 & 97.61 & 79.73 & 77.55 & 51.19 & \bf 100.00 & \bf 99.99
\\
MoEAD~\cite{mengmoead} & \ccheck & 77.03 & 75.23 & 78.84 & 91.06 & 59.31 & 78.14 & 52.51 & 95.12 & 76.91 & 81.30 & 96.95 & 72.25 & 66.77
\\
\rowcolor{OursColor} {\ours} & \ccheck & \bf 96.00 & \bf 94.16 & \bf 97.83 & \bf 99.85 & 99.90 & 99.86 & \bf 71.21 & \bf 100.00 & 99.32 & \bf 98.17 & 92.47 & \bf 100.00 & 99.21
\\
\specialrule{.15em}{.05em}{.05em}
\end{tabular}
}

\caption{
    \textbf{Comparison ($\uparrow$) on DAGM \it{exp50}}~\cite{ho2024ltad} in image-level AUROC using the same evaluation metrics as Tab.~\ref{table:mvtec-exp100-det}. 
}
\label{table:dagm-exp50-det}

\end{table*}

\begin{table*}[ht]

\centering
\setlength{\tabcolsep}{3pt}

\resizebox{0.8\linewidth}{!}{%
\begin{tabular}[h]{*{100}{c}}
\specialrule{.15em}{.05em}{.05em}
{Method} & {CA} &
{\it $\gC$} &
{\color{darkblue}$\gC^{\tt head}$} &
{\color{darkred}$\gC^{\tt tail}$} &
{\color{darkblue}\it Class10} &
{\color{darkblue}\it Class7} &
{\color{darkblue}\it Class9} &
{\color{darkblue}\it Class8} &
{\color{darkblue}\it Class2} &
{\color{darkred}\it Class3} &
{\color{darkred}\it Class5} &
{\color{darkred}\it Class1} &
{\color{darkred}\it Class4} &
{\color{darkred}\it Class6} &
\\
\cmidrule(lr){1-38}
RegAD~\cite{huang2022registration} & \ccross & 90.29 & 91.91 & 88.66 & 98.74 & 88.41 & \bf 99.95 & 73.28 & 99.17 & 93.69 & 85.23 & 72.23 & \bf 99.73 & 92.44
\\
AnomalyGPT~\cite{gu2024agpt} & \ccross & 77.37 & 78.89 & 75.84 & 73.04 & 86.30 & 88.76 & 66.11 & 80.26 & 78.17 & 78.85 & 79.04 & 54.37 & 88.79
\\
LTAD~\cite{ho2024ltad} & \ccross & 97.40 & 97.11 & \bf 97.68 & 99.09 & 97.49 & 99.93 & \bf 89.18 & \bf 99.87 & \bf 98.94 & \bf 98.79 & 93.90 & 99.49 & 97.27
\\
HVQ~\cite{lu2023hierarchical} & \ccross & 86.03 & 84.32 & 87.75 & 98.81 & 73.45 & 95.16 & 55.29 & 98.85 & 83.15 & 85.11 & 92.47 & 98.86 & 79.17
\\
\cmidrule(lr){1-38}
UniAD~\cite{you2022unified} & \ccheck & 90.70 & 89.75 & 91.64 & \bf 99.76 & 93.54 & 99.29 & 56.45 & 99.74 & 90.87 & 89.53 & 79.49 & 98.91 & \bf 99.44
\\
MoEAD~\cite{mengmoead} & \ccheck & 86.08 & 84.02 & 88.15 & 98.46 & 68.93 & 97.60 & 56.45 & 98.66 & 87.27 & 87.73 & 89.69 & 97.00 & 79.03
\\
\rowcolor{OursColor} {\ours} & \ccheck & \bf 97.41 & \bf 97.34 & 97.48 & 99.48 & \bf 99.04 & 99.80 & 88.51 & 99.86 & 98.04 & 98.76 & \bf 95.76 & 99.45 & 95.38
\\
\specialrule{.15em}{.05em}{.05em}
\end{tabular}
}

\caption{
    \textbf{Comparison ($\uparrow$) on DAGM \it{exp50}}~\cite{ho2024ltad} in pixel-level AUROC using the same evaluation metrics as Tab.~\ref{table:mvtec-exp100-seg}. 
}
\label{table:dagm-exp50-seg}

\end{table*}

\begin{table*}[ht]

\centering
\setlength{\tabcolsep}{3pt}

\resizebox{0.8\linewidth}{!}{%
\begin{tabular}[h]{*{100}{c}}
\specialrule{.15em}{.05em}{.05em}
{Method} & {CA} &
{\it $\gC$} &
{\color{darkblue}$\gC^{\tt head}$} &
{\color{darkred}$\gC^{\tt tail}$} &
{\color{darkblue}\it Class10} &
{\color{darkblue}\it Class7} &
{\color{darkblue}\it Class9} &
{\color{darkblue}\it Class8} &
{\color{darkblue}\it Class2} &
{\color{darkred}\it Class3} &
{\color{darkred}\it Class5} &
{\color{darkred}\it Class1} &
{\color{darkred}\it Class4} &
{\color{darkred}\it Class6} &
\\
\cmidrule(lr){1-38}
RegAD~\cite{huang2022registration} & \ccross & 84.86 & 86.88 & 82.84 & 94.45 & 91.34 & 99.23 & 50.94 & 98.44 & 97.95 & 72.07 & 48.76 & 99.75 & 95.66
\\
AnomalyGPT~\cite{gu2024agpt} & \ccross & 85.31 & 86.05 & 84.58 & 86.56 & \bf 100.00 & 89.76 & \bf 71.90 & 82.03 & 94.29 & 95.42 & 84.22 & 52.27 & 96.69
\\
LTAD~\cite{ho2024ltad} & \ccross & 94.40 & 93.24 & 95.56 & 97.78 & 99.60 & \bf 99.89 & 68.92 & \bf 100.00 & \bf 99.66 & 97.35 & 81.01 & \bf 100.00 & 99.78
\\
HVQ~\cite{lu2023hierarchical} & \ccross & 75.42 & 75.88 & 74.96 & 94.17 & 70.69 & 67.04 & 51.00 & 96.49 & 62.20 & 74.47 & \bf 99.59 & 86.08 & 52.44
\\
\cmidrule(lr){1-38}
UniAD~\cite{you2022unified} & \ccheck & 84.34 & 87.66 & 81.02 & 99.15 & 96.01 & 92.45 & 53.31 & 97.40 & 79.79 & 74.26 & 51.15 & 99.91 & \bf 99.99
\\
MoEAD~\cite{mengmoead} & \ccheck & 73.26 & 72.53 & 73.99 & 88.93 & 56.55 & 73.94 & 50.73 & 92.50 & 68.31 & 77.81 & 97.08 & 61.74 & 65.02
\\
\rowcolor{OursColor} {\ours} & \ccheck & \bf 95.58 & \bf 93.75 & \bf 97.41 & \bf 99.60 & 99.92 & 99.79 & 69.45 & \bf 100.00 & 99.60 & \bf 97.66 & 90.94 & \bf 100.00 & 98.84
\\
\specialrule{.15em}{.05em}{.05em}
\end{tabular}
}

\caption{
    \textbf{Comparison ($\uparrow$) on DAGM \it{exp100}}~\cite{ho2024ltad} in image-level AUROC using the same evaluation metrics as Tab.~\ref{table:mvtec-exp100-det}. 
}
\label{table:dagm-exp100-det}

\end{table*}

\begin{table*}[ht]

\centering
\setlength{\tabcolsep}{3pt}

\resizebox{0.8\linewidth}{!}{%
\begin{tabular}[h]{*{100}{c}}
\specialrule{.15em}{.05em}{.05em}
{Method} & {CA} &
{\it $\gC$} &
{\color{darkblue}$\gC^{\tt head}$} &
{\color{darkred}$\gC^{\tt tail}$} &
{\color{darkblue}\it Class10} &
{\color{darkblue}\it Class7} &
{\color{darkblue}\it Class9} &
{\color{darkblue}\it Class8} &
{\color{darkblue}\it Class2} &
{\color{darkred}\it Class3} &
{\color{darkred}\it Class5} &
{\color{darkred}\it Class1} &
{\color{darkred}\it Class4} &
{\color{darkred}\it Class6} &
\\
\cmidrule(lr){1-38}
RegAD~\cite{huang2022registration} & \ccross & 90.28 & 91.91 & 88.66 & 98.74 & 88.40 & \bf 99.95 & 73.28 & 99.17 & 93.69 & 85.23 & 72.23 & \bf 99.73 & 92.44
\\
AnomalyGPT~\cite{gu2024agpt} & \ccross & 77.20 & 79.19 & 75.21 & 74.34 & 86.14 & 89.99 & 66.26 & 79.22 & 77.37 & 77.79 & 80.49 & 54.93 & 85.45
\\
LTAD~\cite{ho2024ltad} & \ccross & 97.30 & 97.01 & \bf 97.58 & 99.09 & 97.42 & 99.93 & \bf 88.74 & \bf 99.87 & \bf 98.95 & 98.62 & 93.76 & 99.59 & 96.98
\\
HVQ~\cite{lu2023hierarchical} & \ccross & 85.54 & 84.07 & 87.01 & 98.82 & 71.83 & 95.48 & 55.51 & 98.70 & 83.71 & 84.81 & 92.75 & 98.34 & 75.44
\\
\cmidrule(lr){1-38}
UniAD~\cite{you2022unified} & \ccheck & 90.13 & 89.66 & 90.60 & \bf 99.75 & 93.54 & 99.33 & 55.96 & 99.76 & 90.13 & 87.54 & 78.19 & 97.74 & \bf 99.42
\\
MoEAD~\cite{mengmoead} & \ccheck & 84.99 & 83.07 & 86.92 & 97.76 & 67.49 & 96.90 & 55.18 & 98.03 & 85.14 & 85.19 & 89.18 & 95.54 & 79.53
\\
\rowcolor{OursColor} {\ours} & \ccheck & \bf 97.32 & \bf 97.26 & 97.39 & 99.41 & \bf 98.86 & 99.79 & 88.37 & 99.86 & 98.05 & \bf 98.67 & \bf 95.61 & 99.41 & 95.20
\\
\specialrule{.15em}{.05em}{.05em}
\end{tabular}
}

\caption{
    \textbf{Comparison ($\uparrow$) on DAGM \it{exp100}}~\cite{ho2024ltad} in pixel-level AUROC using the same evaluation metrics as Tab.~\ref{table:mvtec-exp100-seg}. 
}
\label{table:dagm-exp100-seg}

\end{table*}

\begin{table*}[ht]

\centering
\setlength{\tabcolsep}{3pt}

\resizebox{0.8\linewidth}{!}{%
\begin{tabular}[h]{*{100}{c}}
\specialrule{.15em}{.05em}{.05em}
{Method} & {CA} &
{\it $\gC$} &
{\color{darkblue}$\gC^{\tt head}$} &
{\color{darkred}$\gC^{\tt tail}$} &
{\color{darkblue}\it Class10} &
{\color{darkblue}\it Class7} &
{\color{darkblue}\it Class9} &
{\color{darkblue}\it Class8} &
{\color{darkblue}\it Class2} &
{\color{darkred}\it Class3} &
{\color{darkred}\it Class5} &
{\color{darkred}\it Class1} &
{\color{darkred}\it Class4} &
{\color{darkred}\it Class6} &
\\
\cmidrule(lr){1-38}
RegAD~\cite{huang2022registration} & \ccross & 84.86 & 86.88 & 82.84 & 94.45 & 91.34 & 99.23 & 50.94 & 98.44 & 97.95 & 72.07 & 48.76 & 99.75 & 95.66
\\
AnomalyGPT~\cite{gu2024agpt} & \ccross & 83.29 & 83.45 & 83.13 & 84.31 & \bf 100.00 & 87.16 & 66.80 & 79.00 & 89.90 & 95.65 & 80.03 & 52.49 & 97.56
\\
LTAD~\cite{ho2024ltad} & \ccross & 94.29 & 93.08 & 95.51 & 97.78 & 99.62 & 99.80 & 68.21 & \bf 100.00 & 99.51 & 97.34 & 80.90 & \bf 100.00 & 99.82
\\
HVQ~\cite{lu2023hierarchical} & \ccross & 76.19 & 76.36 & 76.03 & 95.78 & 73.96 & 64.36 & 50.82 & 96.90 & 62.25 & 75.00 & \bf 99.57 & 88.06 & 55.25
\\
\cmidrule(lr){1-38}
UniAD~\cite{you2022unified} & \ccheck & 83.56 & 87.09 & 80.02 & 99.40 & 95.42 & 89.82 & 53.32 & 97.53 & 79.13 & 72.48 & 50.52 & 97.98 & \bf 99.99
\\
MoEAD~\cite{mengmoead} & \ccheck & 72.61 & 72.11 & 73.10 & 89.16 & 54.77 & 73.66 & 51.42 & 91.55 & 67.92 & 77.01 & 96.98 & 61.91 & 61.69
\\
\rowcolor{OursColor} {\ours} & \ccheck & \bf 94.71 & \bf 93.79 & \bf 95.63 & \bf 99.51 & 99.88 & \bf 99.89 & \bf 69.66 & \bf 100.00 & \bf 99.57 & \bf 98.27 & 81.53 & \bf 100.00 & 98.80
\\
\specialrule{.15em}{.05em}{.05em}
\end{tabular}
}

\caption{
    \textbf{Comparison ($\uparrow$) on DAGM \it{exp200}}~\cite{ho2024ltad} in image-level AUROC using the same evaluation metrics as Tab.~\ref{table:mvtec-exp100-det}. 
}
\label{table:dagm-exp200-det}

\end{table*}

\begin{table*}[ht]

\centering
\setlength{\tabcolsep}{3pt}

\resizebox{0.8\linewidth}{!}{%
\begin{tabular}[h]{*{100}{c}}
\specialrule{.15em}{.05em}{.05em}
{Method} & {CA} &
{\it $\gC$} &
{\color{darkblue}$\gC^{\tt head}$} &
{\color{darkred}$\gC^{\tt tail}$} &
{\color{darkblue}\it Class10} &
{\color{darkblue}\it Class7} &
{\color{darkblue}\it Class9} &
{\color{darkblue}\it Class8} &
{\color{darkblue}\it Class2} &
{\color{darkred}\it Class3} &
{\color{darkred}\it Class5} &
{\color{darkred}\it Class1} &
{\color{darkred}\it Class4} &
{\color{darkred}\it Class6} &
\\
\cmidrule(lr){1-38}
RegAD~\cite{huang2022registration} & \ccross & 90.29 & 91.91 & 88.66 & 98.74 & 88.40 & \bf 99.95 & 73.28 & 99.17 & 93.69 & 85.23 & 72.23 & \bf 99.73 & 92.44
\\
AnomalyGPT~\cite{gu2024agpt} & \ccross & 77.16 & 79.58 & 74.73 & 74.85 & 85.83 & 89.38 & 65.89 & 81.93 & 73.30 & 77.54 & 81.12 & 56.02 & 85.69
\\
LTAD~\cite{ho2024ltad} & \ccross & \bf 97.19 & 96.98 & \bf 97.39 & 99.02 & 97.38 & 99.92 & \bf 88.72 & 99.86 & \bf 98.88 & 98.50 & 93.26 & 99.52 & 96.80
\\
HVQ~\cite{lu2023hierarchical} & \ccross & 85.32 & 84.22 & 86.42 & 98.89 & 72.99 & 94.89 & 55.59 & 98.75 & 84.66 & 85.63 & 91.97 & 98.11 & 71.70
\\
\cmidrule(lr){1-38}
UniAD~\cite{you2022unified} & \ccheck & 89.73 & 89.57 & 89.88 & \bf 99.77 & 93.28 & 99.22 & 55.84 & 99.76 & 90.30 & 86.78 & 76.53 & 96.44 & \bf 99.39
\\
MoEAD~\cite{mengmoead} & \ccheck & 84.18 & 82.54 & 85.81 & 97.70 & 64.61 & 96.94 & 55.46 & 97.99 & 84.78 & 85.19 & 88.57 & 95.36 & 75.15
\\
\rowcolor{OursColor} {\ours} & \ccheck & 97.11 & \bf 97.29 & 96.94 & 99.41 & \bf 98.85 & 99.81 & 88.48 & \bf 99.88 & 98.01 & \bf 98.64 & \bf 93.88 & 99.41 & 94.76
\\
\specialrule{.15em}{.05em}{.05em}
\end{tabular}
}

\caption{
    \textbf{Comparison ($\uparrow$) on DAGM \it{exp200}}~\cite{ho2024ltad} in pixel-level AUROC using the same evaluation metrics as Tab.~\ref{table:mvtec-exp100-seg}. 
}
\label{table:dagm-exp200-seg}

\end{table*}

\begin{table*}[ht]

\centering
\setlength{\tabcolsep}{3pt}

\resizebox{0.8\linewidth}{!}{%
\begin{tabular}[h]{*{100}{c}}
\specialrule{.15em}{.05em}{.05em}
{Method} & {CA} &
{\it $\gC$} &
{\color{darkblue}$\gC^{\tt head}$} &
{\color{darkred}$\gC^{\tt tail}$} &
{\color{darkblue}\it Class10} &
{\color{darkblue}\it Class7} &
{\color{darkblue}\it Class9} &
{\color{darkblue}\it Class8} &
{\color{darkblue}\it Class2} &
{\color{darkred}\it Class3} &
{\color{darkred}\it Class5} &
{\color{darkred}\it Class1} &
{\color{darkred}\it Class4} &
{\color{darkred}\it Class6} &
\\
\cmidrule(lr){1-38}
RegAD~\cite{huang2022registration} & \ccross & 84.86 & 86.88 & 82.84 & 94.45 & 91.34 & 99.23 & 50.94 & 98.44 & 97.95 & 72.07 & 48.76 & 99.75 & 95.66
\\
AnomalyGPT~\cite{gu2024agpt} & \ccross & 83.47 & 84.38 & 82.55 & 86.38 & \bf 99.99 & 91.88 & 64.78 & 78.88 & 90.31 & 95.06 & 79.88 & 52.72 & 94.78
\\
LTAD~\cite{ho2024ltad} & \ccross & 93.54 & 93.23 & \bf 93.85 & 97.76 & 99.77 & 99.79 & 68.81 & \bf 100.00 & 99.36 & 97.04 & 73.00 & \bf 100.00 & 99.83
\\
HVQ~\cite{lu2023hierarchical} & \ccross & 76.27 & 77.51 & 75.04 & 96.24 & 75.46 & 67.12 & 51.39 & 97.32 & 61.73 & 76.55 & \bf 99.06 & 76.79 & 61.07
\\
\cmidrule(lr){1-38}
UniAD~\cite{you2022unified} & \ccheck & 81.35 & 88.75 & 73.96 & \bf 99.89 & 98.39 & 96.60 & 51.31 & 97.56 & 85.24 & 75.57 & 53.49 & 55.64 & \bf 99.84
\\
MoEAD~\cite{mengmoead} & \ccheck & 72.28 & 72.81 & 71.74 & 89.39 & 57.43 & 72.99 & 52.05 & 92.20 & 68.71 & 75.06 & 94.66 & 59.14 & 61.14
\\
\rowcolor{OursColor} {\ours} & \ccheck & \bf 93.82 & \bf 93.98 & 93.67 & 99.60 & 99.85 & \bf 99.82 & \bf 70.62 & \bf 100.00 & \bf 99.51 & \bf 97.67 & 72.66 & \bf 100.00 & 98.50
\\
\specialrule{.15em}{.05em}{.05em}
\end{tabular}
}

\caption{
    \textbf{Comparison ($\uparrow$) on DAGM \it{exp500}}~\cite{ho2024ltad} in image-level AUROC using the same evaluation metrics as Tab.~\ref{table:mvtec-exp100-det}. 
}
\label{table:dagm-exp500-det}

\end{table*}

\begin{table*}[ht]

\centering
\setlength{\tabcolsep}{3pt}

\resizebox{0.8\linewidth}{!}{%
\begin{tabular}[h]{*{100}{c}}
\specialrule{.15em}{.05em}{.05em}
{Method} & {CA} &
{\it $\gC$} &
{\color{darkblue}$\gC^{\tt head}$} &
{\color{darkred}$\gC^{\tt tail}$} &
{\color{darkblue}\it Class10} &
{\color{darkblue}\it Class7} &
{\color{darkblue}\it Class9} &
{\color{darkblue}\it Class8} &
{\color{darkblue}\it Class2} &
{\color{darkred}\it Class3} &
{\color{darkred}\it Class5} &
{\color{darkred}\it Class1} &
{\color{darkred}\it Class4} &
{\color{darkred}\it Class6} &
\\
\cmidrule(lr){1-38}
RegAD~\cite{huang2022registration} & \ccross & 90.29 & 91.91 & 88.66 & 98.74 & 88.40 & \bf 99.95 & 73.28 & 99.17 & 93.69 & 85.23 & 72.23 & \bf 99.73 & 92.44
\\
AnomalyGPT~\cite{gu2024agpt} & \ccross & 76.87 & 78.93 & 74.81 & 76.11 & 85.42 & 90.89 & 62.51 & 79.73 & 72.28 & 77.29 & 82.21 & 57.35 & 84.93
\\
LTAD~\cite{ho2024ltad} & \ccross & \bf 97.01 & 96.94 & \bf 97.07 & 99.02 & 97.33 & 99.92 & \bf 88.57 & \bf 99.87 & \bf 98.74 & 98.43 & 92.23 & 99.31 & 96.63
\\
HVQ~\cite{lu2023hierarchical} & \ccross & 85.41 & 84.74 & 86.08 & 98.94 & 74.08 & 95.27 & 56.52 & 98.89 & 83.16 & 85.12 & 89.74 & 96.55 & 75.87
\\
\cmidrule(lr){1-38}
UniAD~\cite{you2022unified} & \ccheck & 88.63 & 89.44 & 87.82 & \bf 99.77 & 93.10 & 99.22 & 55.37 & 99.73 & 89.05 & 84.62 & 75.15 & 90.93 & \bf 99.33
\\
MoEAD~\cite{mengmoead} & \ccheck & 83.34 & 83.11 & 83.57 & 97.82 & 68.03 & 96.68 & 55.14 & 97.89 & 84.76 & 84.24 & 85.67 & 93.03 & 70.12
\\
\rowcolor{OursColor} {\ours} & \ccheck & 96.98 & \bf 97.23 & 96.72 & 99.39 & \bf 98.84 & 99.83 & 88.25 & 99.86 & 98.10 & \bf 98.55 & \bf 93.45 & 99.13 & 94.30
\\
\specialrule{.15em}{.05em}{.05em}
\end{tabular}
}

\caption{
    \textbf{Comparison ($\uparrow$) on DAGM \it{exp500}}~\cite{ho2024ltad} in pixel-level AUROC using the same evaluation metrics as Tab.~\ref{table:mvtec-exp100-seg}. 
}
\label{table:dagm-exp500-seg}

\end{table*}

\begin{table*}[ht]

\centering
\setlength{\tabcolsep}{3pt}

\resizebox{0.8\linewidth}{!}{%
\begin{tabular}[h]{*{100}{c}}
\specialrule{.15em}{.05em}{.05em}
{Method} & {CA} &
{\it $\gC$} &
{\color{darkblue}$\gC^{\tt head}$} &
{\color{darkred}$\gC^{\tt tail}$} &
{\color{darkblue}\it Class6} &
{\color{darkblue}\it Class4} &
{\color{darkblue}\it Class1} &
{\color{darkblue}\it Class5} &
{\color{darkblue}\it Class3} &
{\color{darkred}\it Class2} &
{\color{darkred}\it Class8} &
{\color{darkred}\it Class9} &
{\color{darkred}\it Class7} &
{\color{darkred}\it Class10} &
\\
\cmidrule(lr){1-38}
RegAD~\cite{huang2022registration} & \ccross & 84.85 & 82.83 & 86.88 & 95.66 & 99.75 & 48.76 & 72.07 & 97.95 & 98.44 & 50.94 & 99.23 & 91.34 & 94.45
\\
AnomalyGPT~\cite{gu2024agpt} & \ccross & 84.65 & 86.58 & 82.72 & 99.11 & 67.43 & 74.65 & \bf 97.48 & 94.21 & 88.79 & 63.56 & 92.07 & \bf 99.96 & 69.20
\\
LTAD~\cite{ho2024ltad} & \ccross & 94.09 & \bf 96.85 & 91.34 & 99.76 & \bf 100.00 & 88.00 & 97.19 & 99.29 & \bf 100.00 & 67.80 & 99.75 & 95.03 & 94.12
\\
HVQ~\cite{lu2023hierarchical} & \ccross & 75.49 & 72.14 & 78.84 & 61.39 & 96.04 & \bf 99.65 & 74.83 & 62.26 & 96.37 & 49.37 & 66.89 & 62.88 & 85.20
\\
\cmidrule(lr){1-38}
UniAD~\cite{you2022unified} & \ccheck & 82.48 & 85.78 & 79.17 & \bf 99.93 & \bf 100.00 & 63.41 & 79.04 & 86.52 & 97.00 & 51.42 & 98.31 & 71.48 & 77.67
\\
MoEAD~\cite{mengmoead} & \ccheck & 71.11 & 66.07 & 76.15 & 66.77 & 69.09 & 96.90 & 78.52 & 69.49 & 91.20 & 47.45 & 62.87 & 49.29 & 79.56
\\
\rowcolor{OursColor} {\ours} & \ccheck & \bf 95.24 & 93.21 & \bf 97.27 & 99.19 & \bf 100.00 & 90.26 & 97.36 & \bf 99.54 & \bf 100.00 & \bf 68.78 & \bf 99.89 & 99.61 & \bf 97.78
\\
\specialrule{.15em}{.05em}{.05em}
\end{tabular}
}

\caption{
    \textbf{Comparison ($\uparrow$) on DAGM \it{reverse exp200}}~\cite{ho2024ltad} in image-level AUROC using the same evaluation metrics as Tab.~\ref{table:mvtec-exp100-det}. 
}
\label{table:dagm-reverse-exp200-det}

\end{table*}

\begin{table*}[ht]

\centering
\setlength{\tabcolsep}{3pt}

\resizebox{0.8\linewidth}{!}{%
\begin{tabular}[h]{*{100}{c}}
\specialrule{.15em}{.05em}{.05em}
{Method} & {CA} &
{\it $\gC$} &
{\color{darkblue}$\gC^{\tt head}$} &
{\color{darkred}$\gC^{\tt tail}$} &
{\color{darkblue}\it Class6} &
{\color{darkblue}\it Class4} &
{\color{darkblue}\it Class1} &
{\color{darkblue}\it Class5} &
{\color{darkblue}\it Class3} &
{\color{darkred}\it Class2} &
{\color{darkred}\it Class8} &
{\color{darkred}\it Class9} &
{\color{darkred}\it Class7} &
{\color{darkred}\it Class10} &
\\
\cmidrule(lr){1-38}
RegAD~\cite{huang2022registration} & \ccross & 90.28 & 88.66 & 91.91 & 92.44 & 99.73 & 72.23 & 85.23 & 93.69 & 99.17 & 73.28 & \bf 99.95 & 88.40 & 98.74
\\
AnomalyGPT~\cite{gu2024agpt} & \ccross & 78.41 & 76.49 & \bf 8043.00 & 94.92 & 57.21 & 69.48 & 81.27 & 79.10 & 89.15 & 65.79 & 95.83 & 83.85 & 67.53
\\
LTAD~\cite{ho2024ltad} & \ccross & 96.68 & \bf 97.63 & 95.72 & 97.16 & 99.43 & 94.76 & 98.30 & \bf 98.52 & \bf 99.87 & 85.23 & 99.92 & 95.11 & 98.49
\\
HVQ~\cite{lu2023hierarchical} & \ccross & 85.58 & 82.49 & 88.66 & 83.07 & 98.97 & 92.35 & 85.48 & 83.45 & 98.71 & 56.00 & 94.12 & 66.84 & 96.78
\\
\cmidrule(lr){1-38}
UniAD~\cite{you2022unified} & \ccheck & 89.98 & 92.69 & 87.27 & \bf 99.49 & \bf 99.86 & 84.59 & 90.23 & 89.30 & 99.65 & 53.89 & 99.55 & 85.42 & 97.87
\\
MoEAD~\cite{mengmoead} & \ccheck & 83.55 & 79.28 & 87.83 & 82.51 & 96.61 & 89.01 & 86.02 & 84.99 & 97.32 & 54.46 & 92.42 & 58.52 & 93.66
\\
\rowcolor{OursColor} {\ours} & \ccheck & \bf 97.20 & 96.97 & 97.43 & 95.92 & 99.33 & \bf 95.50 & \bf 98.43 & 97.96 & \bf 99.87 & \bf 87.62 & 99.83 & \bf 98.47 & \bf 99.08
\\
\specialrule{.15em}{.05em}{.05em}
\end{tabular}
}

\caption{
    \textbf{Comparison ($\uparrow$) on DAGM \it{reverse exp200}}~\cite{ho2024ltad} in pixel-level AUROC using the same evaluation metrics as Tab.~\ref{table:mvtec-exp100-seg}. 
}
\label{table:dagm-reverse-exp200-seg}

\end{table*}

\begin{table*}[ht]

\centering
\setlength{\tabcolsep}{3pt}

\resizebox{0.8\linewidth}{!}{%
%
}

\caption{
    \textbf{Comparison ($\uparrow$) on DAGM \it{step50}}~\cite{ho2024ltad} in image-level AUROC using the same evaluation metrics as Tab.~\ref{table:mvtec-exp100-det}. 
}
\label{table:dagm-step50-det}

\end{table*}

\begin{table*}[ht]

\centering
\setlength{\tabcolsep}{3pt}

\resizebox{0.8\linewidth}{!}{%
%
}

\caption{
    \textbf{Comparison ($\uparrow$) on DAGM \it{step50}}~\cite{ho2024ltad} in pixel-level AUROC using the same evaluation metrics as Tab.~\ref{table:mvtec-exp100-seg}. 
}
\label{table:dagm-step50-seg}

\end{table*}

\begin{table*}[ht]

\centering
\setlength{\tabcolsep}{3pt}

\resizebox{0.8\linewidth}{!}{%
%
}

\caption{
    \textbf{Comparison ($\uparrow$) on DAGM \it{step100}}~\cite{ho2024ltad} in image-level AUROC using the same evaluation metrics as Tab.~\ref{table:mvtec-exp100-det}. 
}
\label{table:dagm-step100-det}

\end{table*}

\begin{table*}[ht]

\centering
\setlength{\tabcolsep}{3pt}

\resizebox{0.8\linewidth}{!}{%
%
}

\caption{
    \textbf{Comparison ($\uparrow$) on DAGM \it{step100}}~\cite{ho2024ltad} in pixel-level AUROC using the same evaluation metrics as Tab.~\ref{table:mvtec-exp100-seg}. 
}
\label{table:dagm-step100-seg}

\end{table*}

\begin{table*}[ht]

\centering
\setlength{\tabcolsep}{3pt}

\resizebox{0.8\linewidth}{!}{%
%
}

\caption{
    \textbf{Comparison ($\uparrow$) on DAGM \it{step200}}~\cite{ho2024ltad} in image-level AUROC using the same evaluation metrics as Tab.~\ref{table:mvtec-exp100-det}. 
}
\label{table:dagm-step200-det}

\end{table*}

\begin{table*}[ht]

\centering
\setlength{\tabcolsep}{3pt}

\resizebox{0.8\linewidth}{!}{%
%
}

\caption{
    \textbf{Comparison ($\uparrow$) on DAGM \it{step200}}~\cite{ho2024ltad} in pixel-level AUROC using the same evaluation metrics as Tab.~\ref{table:mvtec-exp100-seg}. 
}
\label{table:dagm-step200-seg}

\end{table*}

\begin{table*}[ht]

\centering
\setlength{\tabcolsep}{3pt}

\resizebox{0.8\linewidth}{!}{%
%
}

\caption{
    \textbf{Comparison ($\uparrow$) on DAGM \it{reverse step200}}~\cite{ho2024ltad} in image-level AUROC using the same evaluation metrics as Tab.~\ref{table:mvtec-exp100-det}. 
}
\label{table:dagm-reverse-step200-det}

\end{table*}

\begin{table*}[ht]

\centering
\setlength{\tabcolsep}{3pt}

\resizebox{0.8\linewidth}{!}{%
%
}

\caption{
    \textbf{Comparison ($\uparrow$) on DAGM \it{reverse step200}}~\cite{ho2024ltad} in pixel-level AUROC using the same evaluation metrics as Tab.~\ref{table:mvtec-exp100-seg}. 
}
\label{table:dagm-reverse-step200-seg}

\end{table*}

\section{Online experiments}
\label{supp:online_results}

\begin{figure*}[h]
    \centering
    \includegraphics[width=\linewidth]{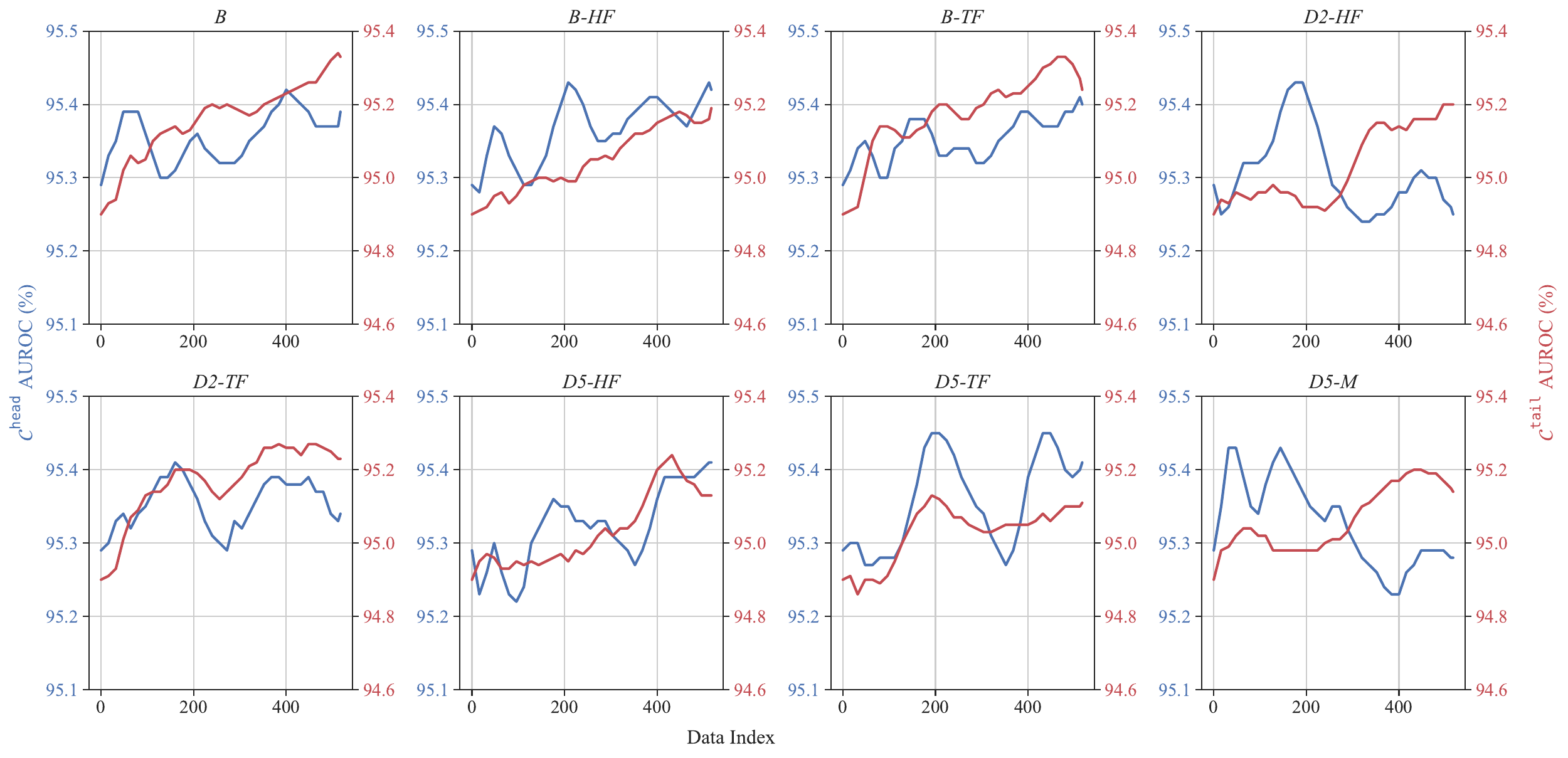}
    \caption{
        \textbf{Performance curves} of \ours in pixel-level AUROC
        on \textcolor{darkblue}{$\gC^{\tt head}$} and \textcolor{darkred}{$\gC^{\tt tail}$}. They are offline-trained on MVTec {\it exp100}~\cite{ho2024ltad} and online-trained on our online benchmarks.
    }
    \label{fig:online-mvtec-exp-100}
\end{figure*}

\begin{figure*}[h]
    \centering
    \includegraphics[width=\linewidth]{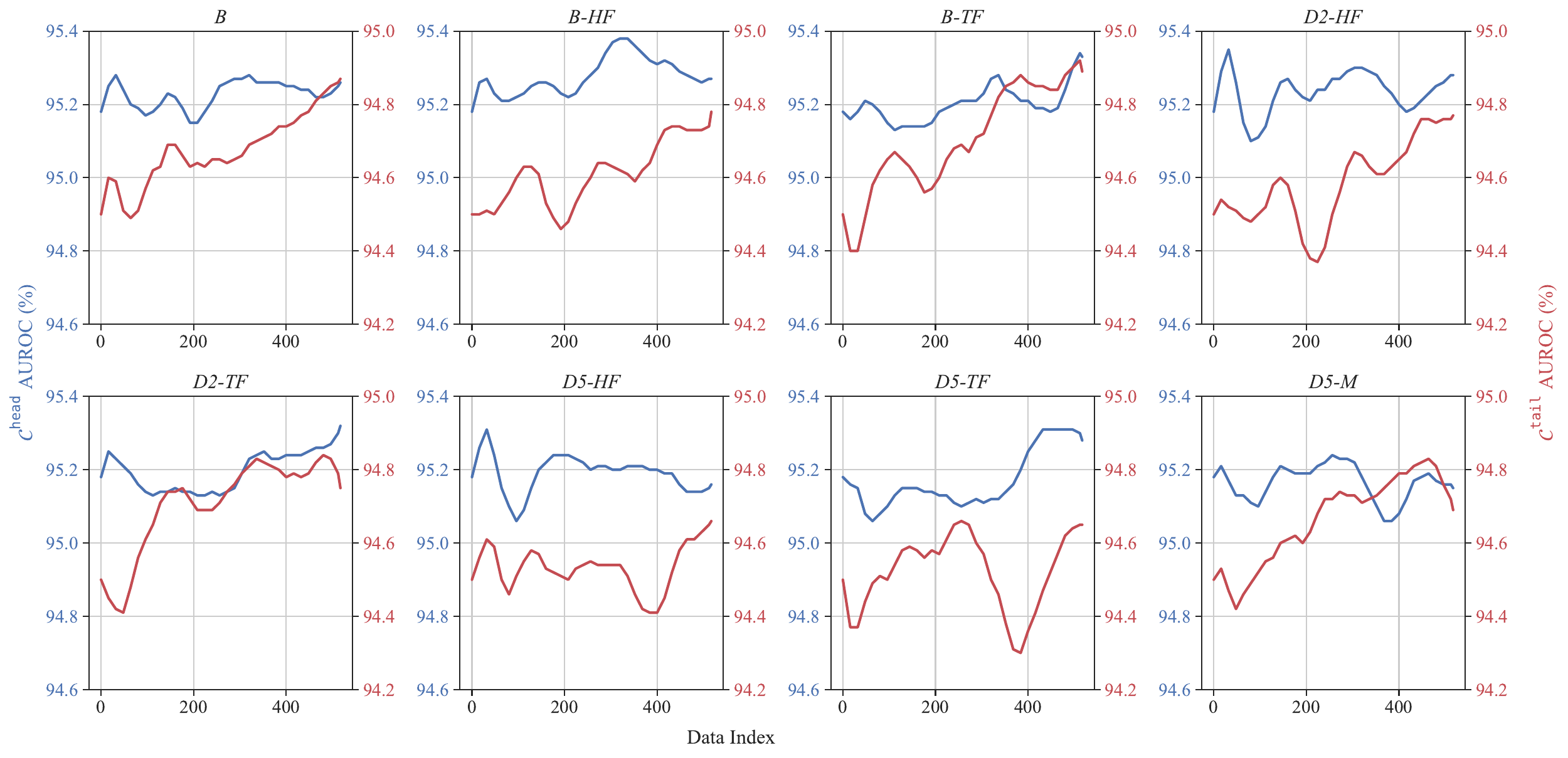}
    \caption{
        \textbf{Performance curves} 
        in the same format as Tab.~\ref{table:supp-mvtec-online}.
        They are offline-trained on MVTec {\it exp200}~\cite{ho2024ltad}.
    }
    \label{fig:online-mvtec-exp-200}
\end{figure*}

\begin{figure*}[h]
    \centering
    \includegraphics[width=\linewidth]{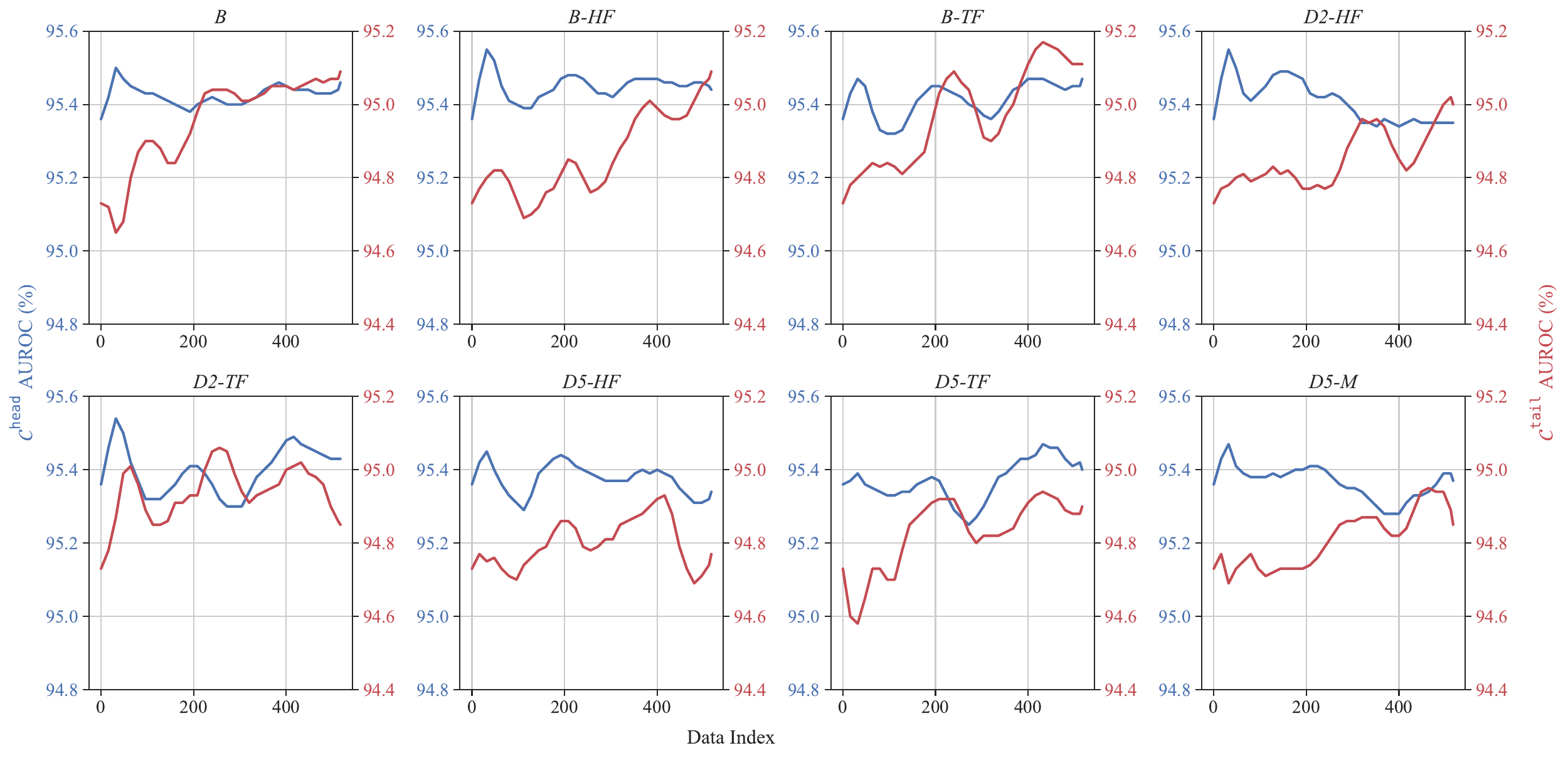}
    \caption{
        \textbf{Performance curves} in the same format as Tab.~\ref{table:supp-mvtec-online}.
        They are offline-trained on MVTec {\it step100}~\cite{ho2024ltad}.
    }
    \label{fig:online-mvtec-step-100}
\end{figure*}

\begin{figure*}[h]
    \centering
    \includegraphics[width=\linewidth]{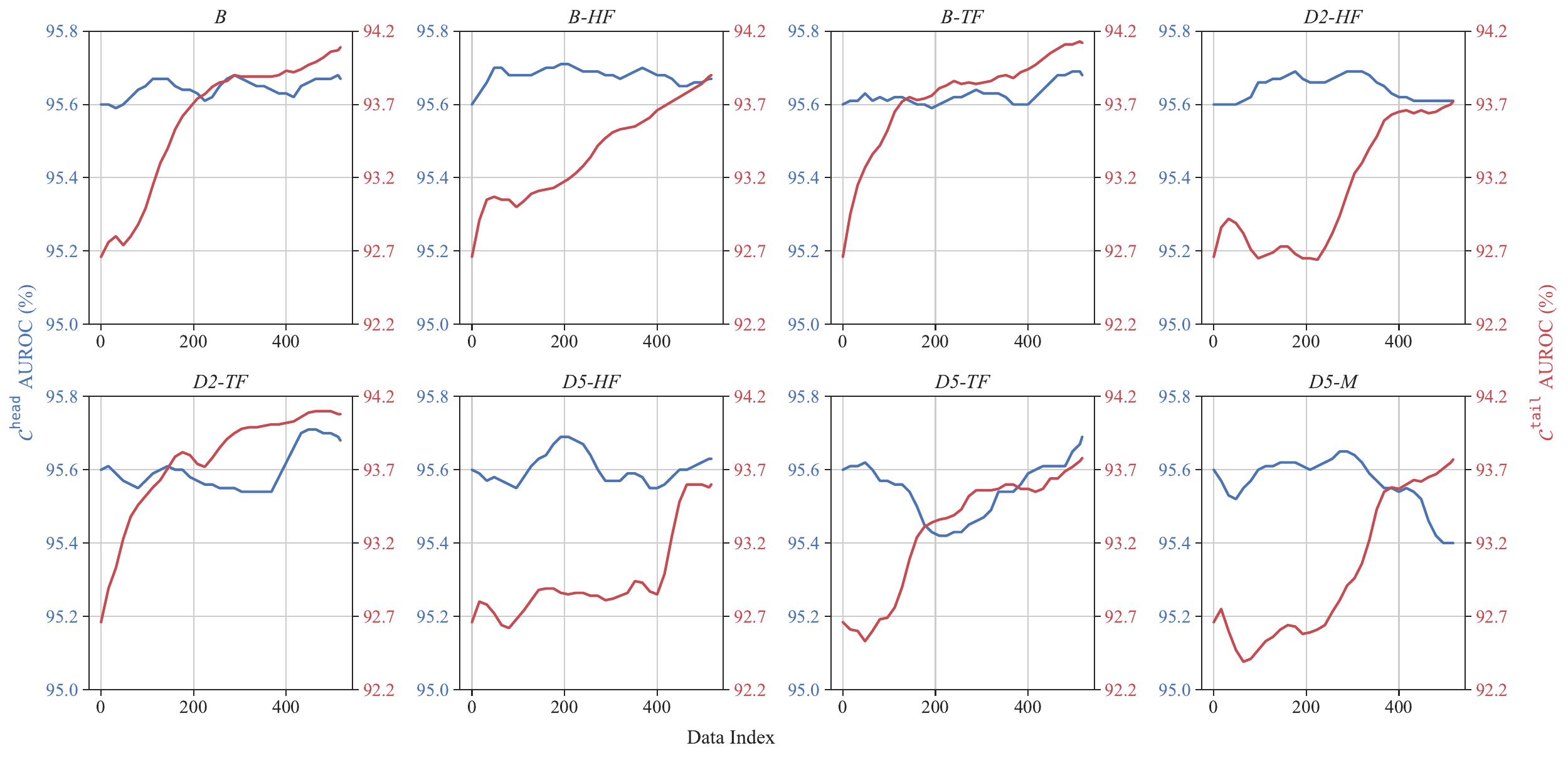}
    \caption{
        \textbf{Performance curves} in the same format as Tab.~\ref{table:supp-mvtec-online}.
        They are offline-trained on MVTec {\it step200}~\cite{ho2024ltad}.
    }
    \label{fig:online-mvtec-step-200}
\end{figure*}

\begin{figure*}[h]
    \centering
    \includegraphics[width=\linewidth]{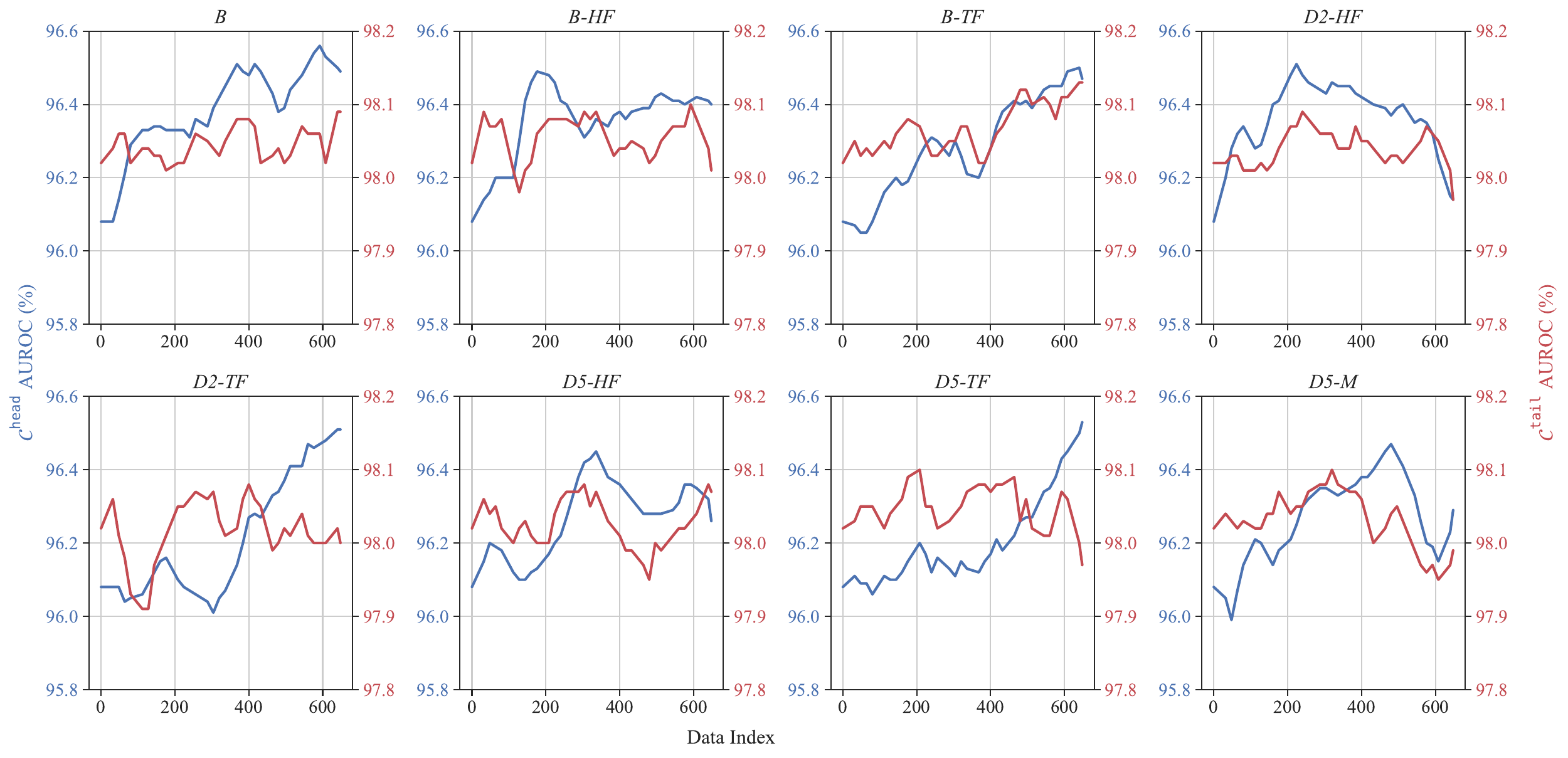}
    \caption{
        \textbf{Performance curves} in the same format as Tab.~\ref{table:supp-mvtec-online}.
        They are offline-trained on VisA {\it exp100}~\cite{ho2024ltad}.
    }
    \label{fig:online-visa-exp-100}
\end{figure*}

\begin{figure*}[h]
    \centering
    \includegraphics[width=\linewidth]{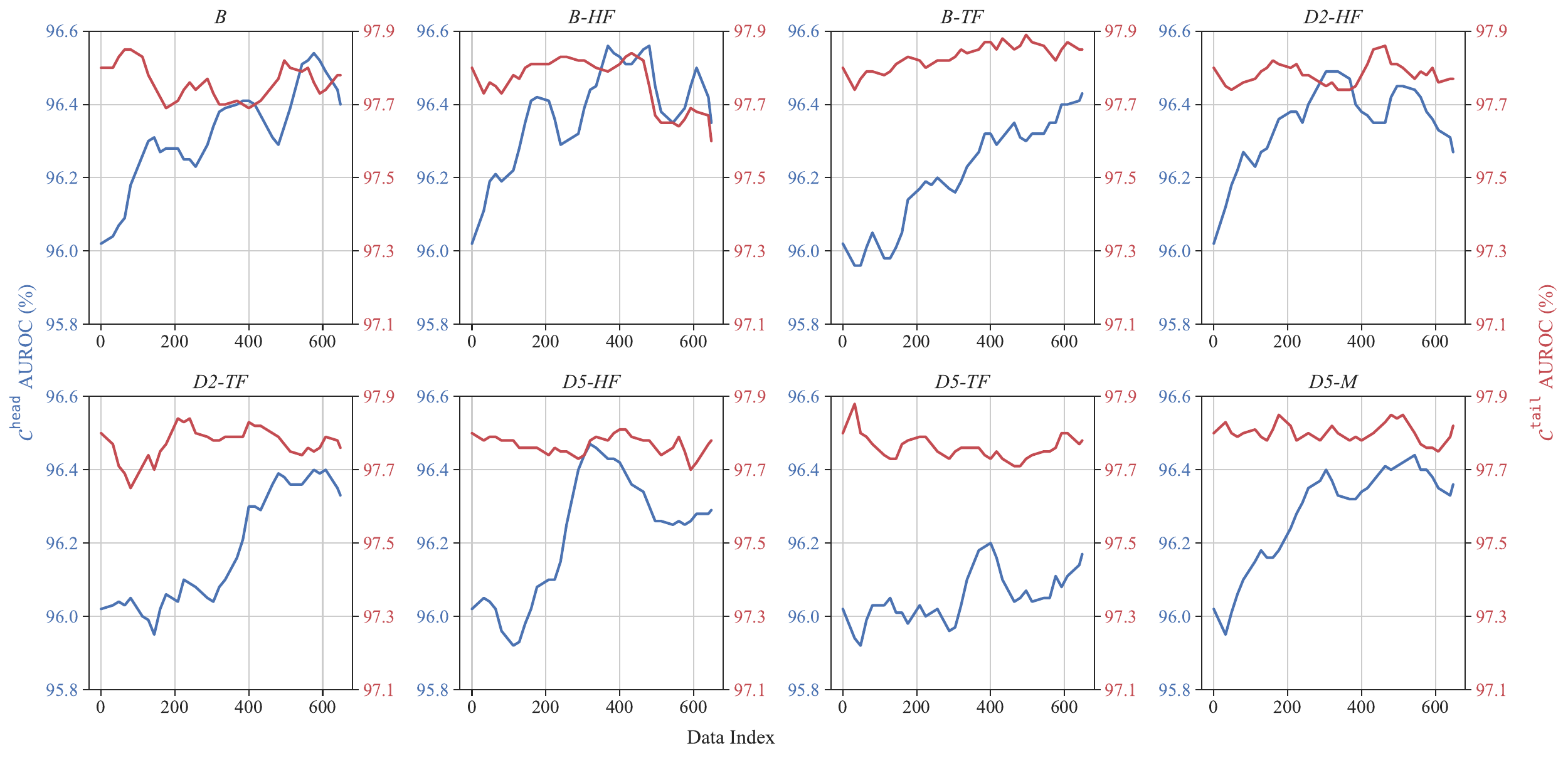}
    \caption{
        \textbf{Performance curves} in the same format as Tab.~\ref{table:supp-mvtec-online}.
        They are offline-trained on VisA {\it exp200}~\cite{ho2024ltad}.
    }
    \label{fig:online-visa-exp-200}
\end{figure*}

\begin{figure*}[h]
    \centering
    \includegraphics[width=\linewidth]{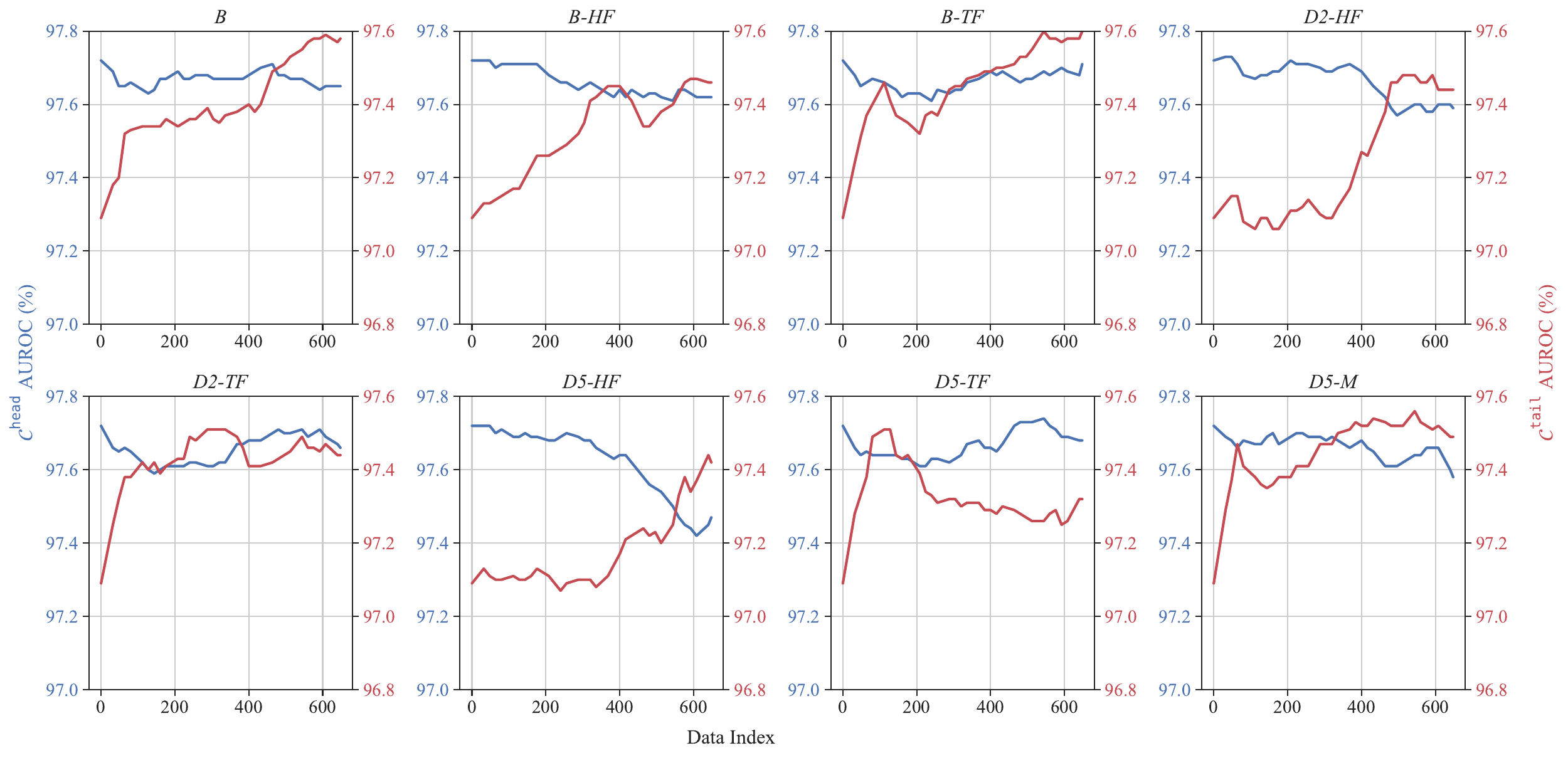}
    \caption{
        \textbf{Performance curves} in the same format as Tab.~\ref{table:supp-mvtec-online}.
        They are offline-trained on VisA {\it step100}~\cite{ho2024ltad}.
    }
    \label{fig:online-visa-step-100}
\end{figure*}

\begin{figure*}[h]
    \centering
    \includegraphics[width=\linewidth]{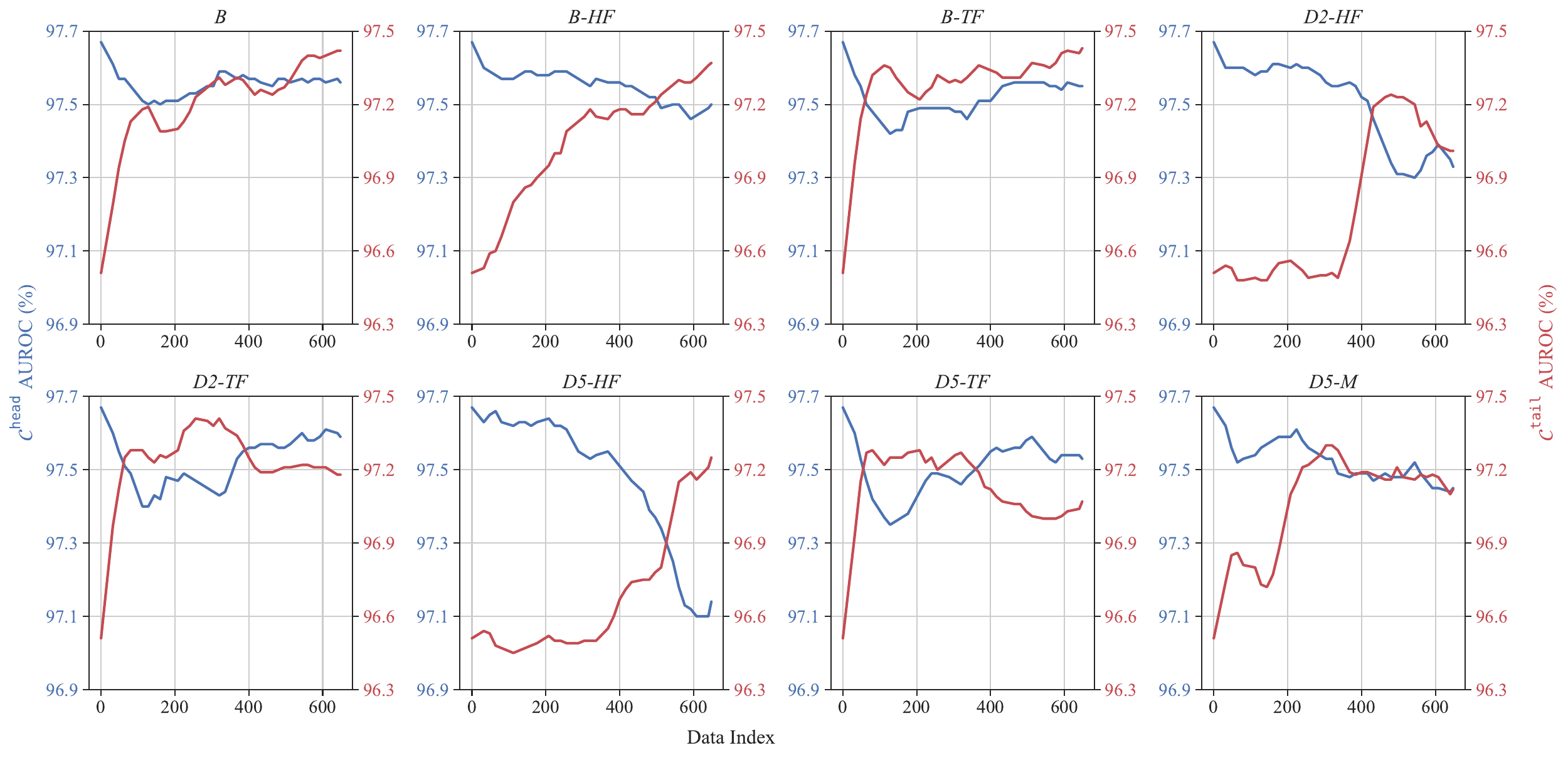}
    \caption{
        \textbf{Performance curves} in the same format as Tab.~\ref{table:supp-mvtec-online}.
        They are offline-trained on VisA {\it step200}~\cite{ho2024ltad}.
    }
    \label{fig:online-visa-step-200}
\end{figure*}

\begin{figure*}[h]
    \centering
    \includegraphics[width=\linewidth]{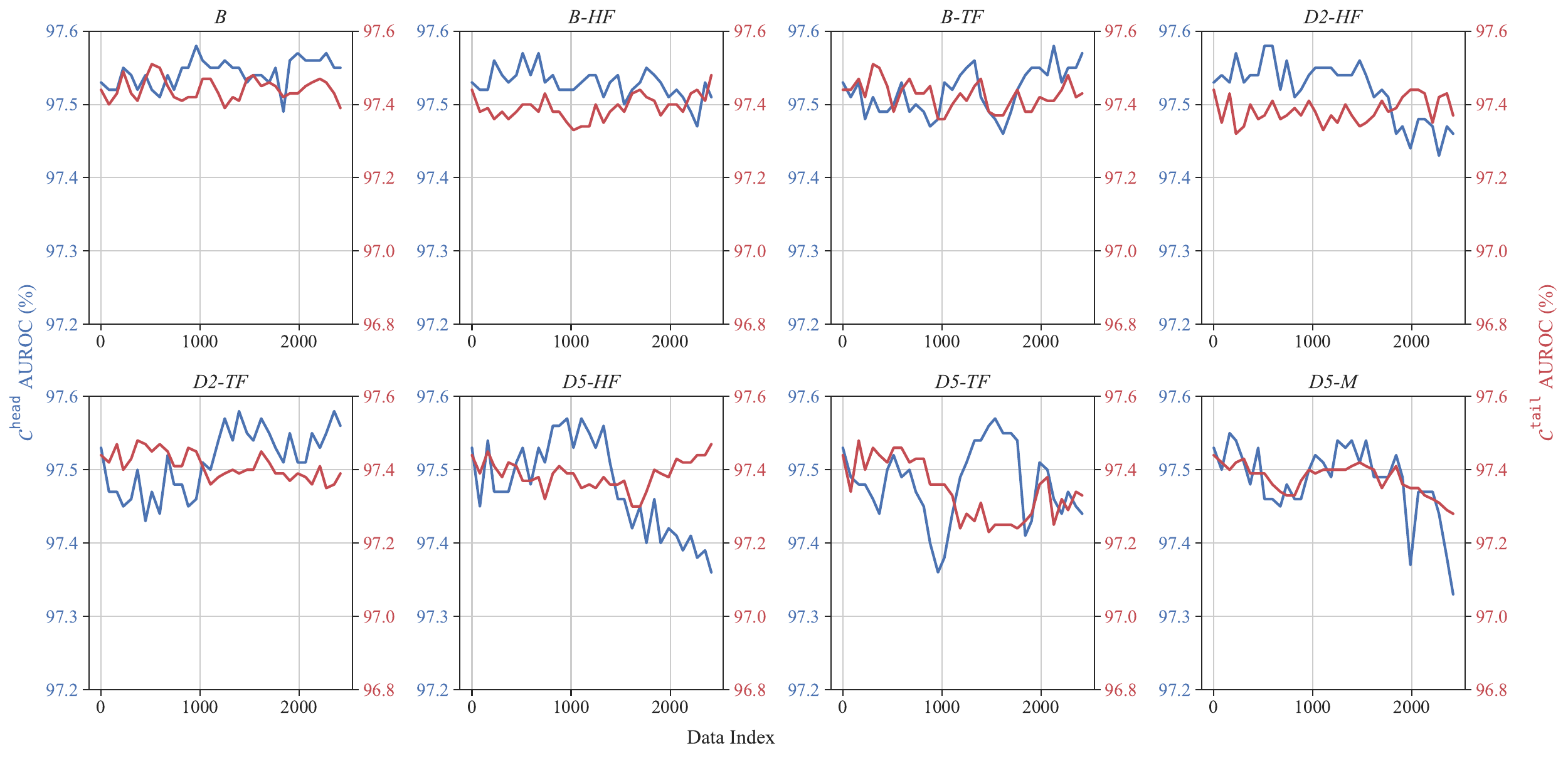}
    \caption{
        \textbf{Performance curves} in the same format as Tab.~\ref{table:supp-mvtec-online}.
        They are offline-trained on DAGM {\it exp100}~\cite{ho2024ltad}.
    }
    \label{fig:online-dagm-exp-100}
\end{figure*}

\begin{figure*}[t]
    \centering
    \includegraphics[width=\linewidth]{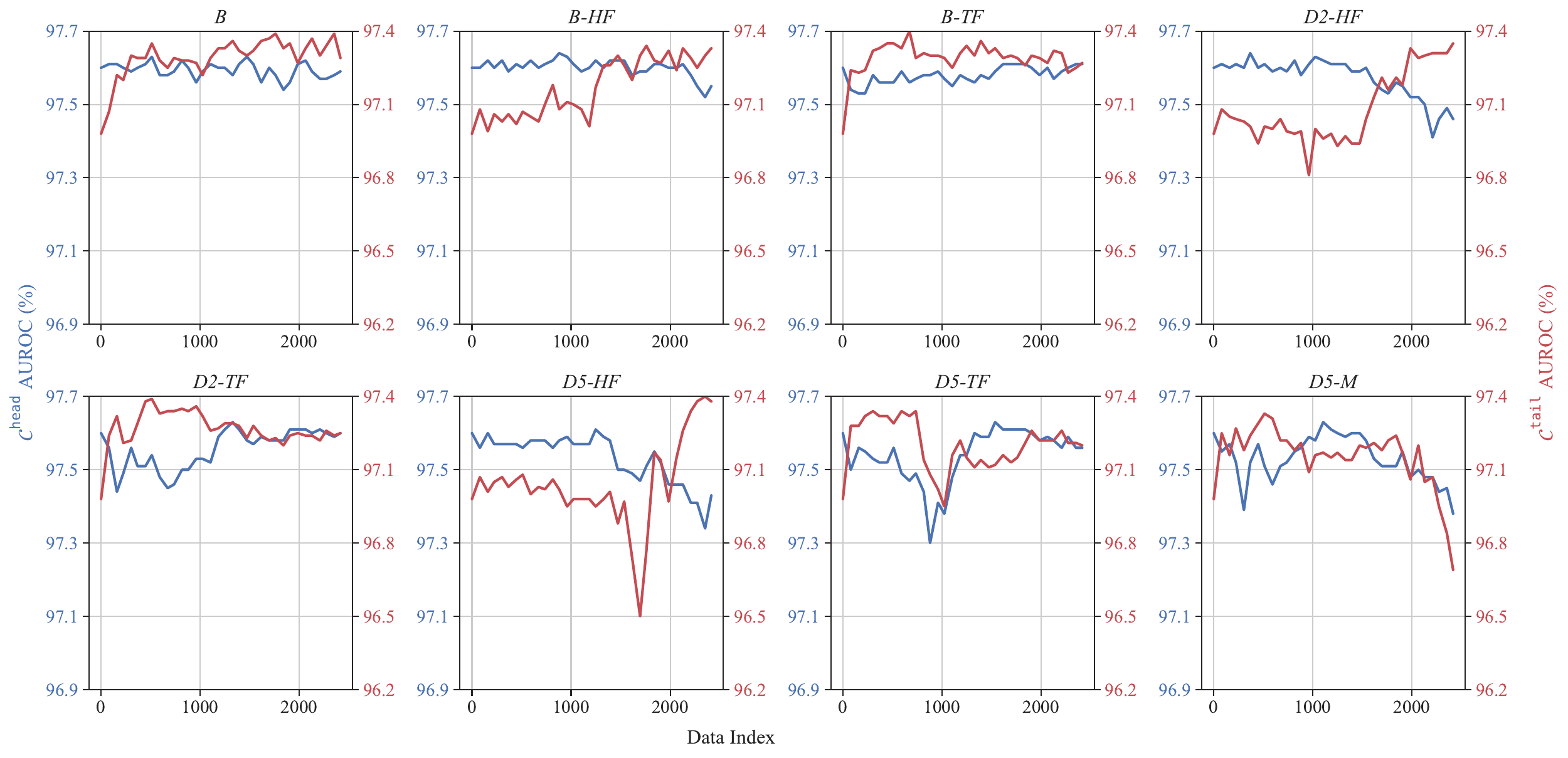}
    \caption{
        \textbf{Performance curves} in the same format as Tab.~\ref{table:supp-mvtec-online}.
        They are offline-trained on DAGM {\it exp200}~\cite{ho2024ltad}.
    }
    \label{fig:online-dagm-exp-200}
\end{figure*}

\begin{figure*}[h]
    \centering
    \includegraphics[width=\linewidth]{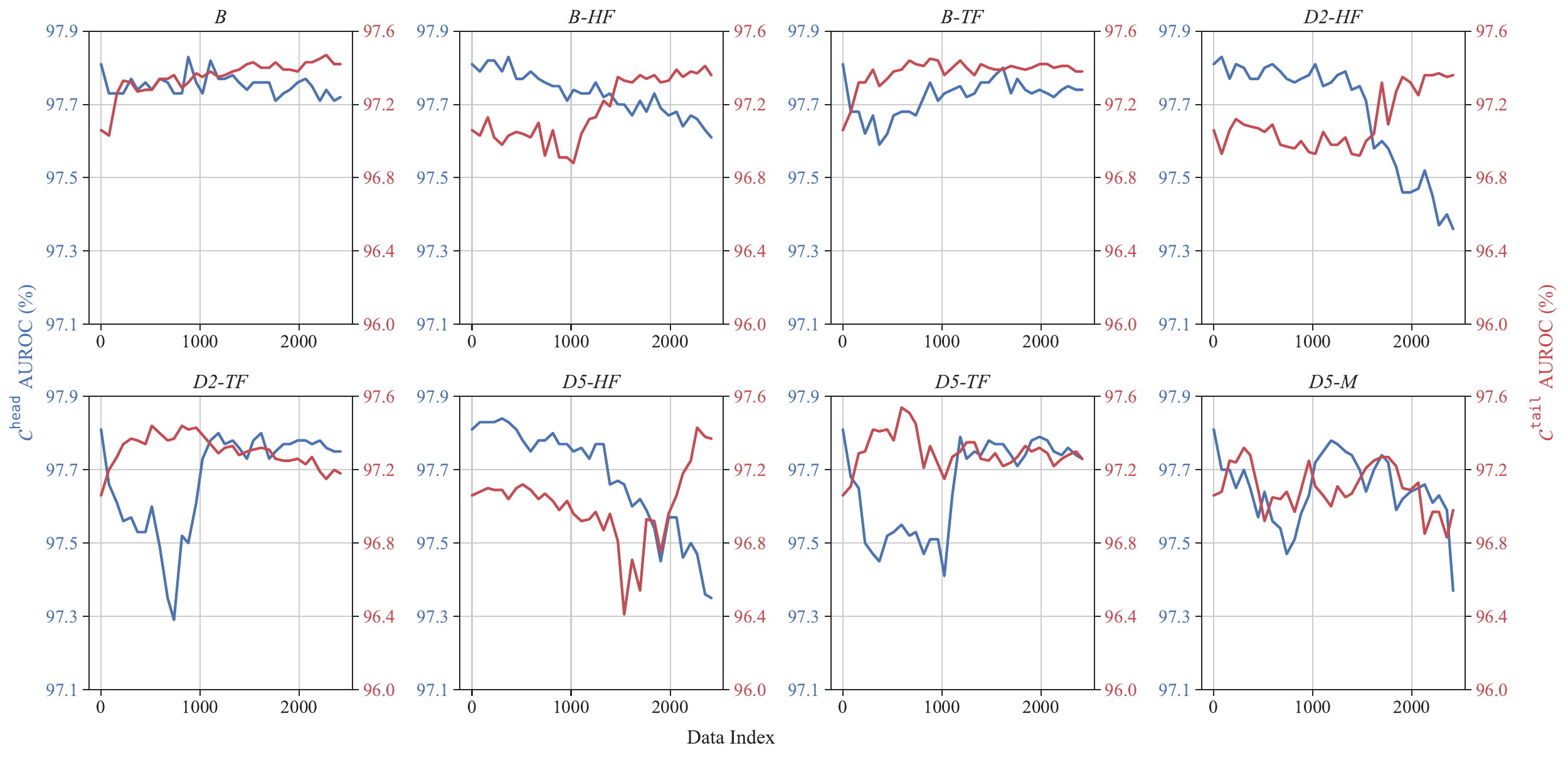}
    \caption{
        \textbf{Performance curves} in the same format as Tab.~\ref{table:supp-mvtec-online}.
        They are offline-trained on DAGM {\it step100}~\cite{ho2024ltad}.
    }
    \label{fig:online-dagm-step-100}
\end{figure*}

\begin{figure*}[h]
    \centering
    \includegraphics[width=\linewidth]{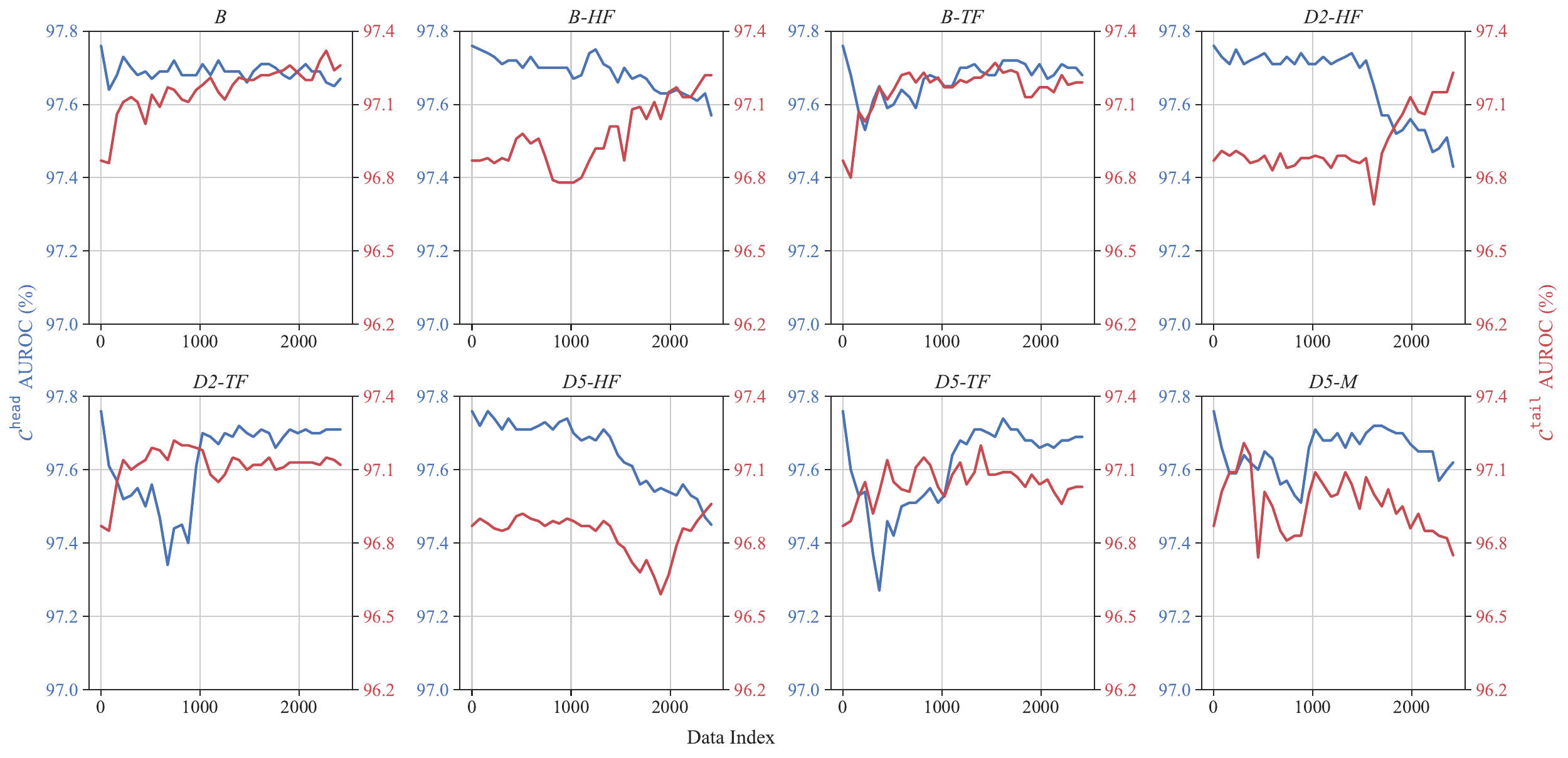}
    \caption{
        \textbf{Performance curves} in the same format as Tab.~\ref{table:supp-mvtec-online}.
        They are offline-trained on DAGM {\it step200}~\cite{ho2024ltad}.
    }
    \label{fig:online-dagm-step-200}
\end{figure*}

We report additional results on our proposed benchmark \ours under more configurations. For each dataset, we report the quantitative comparison at $t=T$ on all $\gD^{\tt O}$ configuration with or without our $\gA^{\tt AA}$ in Tab.~\ref{table:supp-mvtec-online}, Tab.~\ref{table:supp-visa-online}, and Tab.~\ref{table:supp-dagm-online}, respectively. We also plot the performance curves of $\tF_{\theta_t}$ on $\gC^{\tt head}$ and $\gC^{\tt tail}$ for each step $t \in [0, \cdots, T]$. We set $T = 33$ for all experiments.
Specifically, we list them as the following.
\begin{itemize}[topsep=0pt, leftmargin=16pt]
    \item MVTec
    \begin{enumerate}
        \item {\it exp100}: Fig.~\ref{fig:online-mvtec-exp-100}.
        \item {\it exp200}: Fig.~\ref{fig:online-mvtec-exp-200}.
        \item {\it step100}: Fig.~\ref{fig:online-mvtec-step-100}.
        \item {\it step200}: Fig.~\ref{fig:online-mvtec-step-200}.
    \end{enumerate}
    \item VisA
    \begin{enumerate}
        \item {\it exp100}: Fig.~\ref{fig:online-visa-exp-100}.
        \item {\it exp200}: Fig.~\ref{fig:online-visa-exp-200}.
        \item {\it step100}: Fig.~\ref{fig:online-visa-step-100}.
        \item {\it step200}: Fig.~\ref{fig:online-visa-step-200}.
    \end{enumerate}
    \item DAGM
    \begin{enumerate}
        \item {\it exp100}: 
        Fig.~\ref{fig:online-dagm-exp-100}.
        \item {\it exp200}: 
        Fig.~\ref{fig:online-dagm-exp-200}.
        \item {\it step100}: Fig.~\ref{fig:online-dagm-step-100}.
        \item {\it step200}: 
        Fig.~\ref{fig:online-dagm-step-200}.
    \end{enumerate}
\end{itemize}
Apart from the synthetic DAGM dataset where the offline performance is saturated, we observe that \ours improves the performances in most cases, especially under the more challenging offline long-tailed settings, \ie~{\it step200}.

\begin{table}[h]

\centering
\setlength{\tabcolsep}{3pt}

\resizebox{\linewidth}{!}{%
\begin{tabular}{ccccccccccc}
\specialrule{.15em}{.05em}{.05em}
{Config.} & {Online} &
{\it B} &
{\it B-HF} &
{\it B-TF} &
{\it D2-HF} &
{\it D2-TF} &
{\it D5-HF} &
{\it D5-TF} &
{\it D5-M} &
{Avg.}
\\
\cmidrule(lr){1-11}
& \ccross & 95.08 & 95.08 & 95.08 & 95.08 & 95.08 & 95.08 & 95.08 & 95.08 & 95.08
\\
\rowcolor{OursColor} \multirow{-2}{*}{\it exp100} \cellcolor{white} & \ccheck & 
\bf 95.36 & \bf 95.30 & \bf 95.32 & \bf 95.23 & \bf 95.28 & \bf 95.26 & \bf 95.25 & \bf 95.21 & \bf 95.28
\\
& \ccross & 94.82 & 94.82 & 94.82 & 94.82 & 94.82 & 94.82 & 94.82 & 94.82 & 94.82
\\
\rowcolor{OursColor} \multirow{-2}{*}{\it exp200} \cellcolor{white} & \ccheck &
\bf 95.05 & \bf 95.01 & \bf 95.10 & \bf 95.01 & \bf 95.01 & \bf 94.89 & \bf 94.94 & \bf 94.91 & \bf 94.99
\\
 & \ccross &
95.03 & 95.03 & 95.03 & 95.03 & 95.03 & 95.03 & 95.03 & 95.03 & 95.03
\\
\rowcolor{OursColor} \multirow{-2}{*}{\it step100} \cellcolor{white} & \ccheck &
\bf 95.26 & \bf 95.26 & \bf 95.28 & \bf 95.17 & \bf 95.12 & \bf 95.04 & \bf 95.14 & \bf \bf 95.09 & \bf 95.17
\\
& \ccross & 94.03 & 94.03 & 94.03 & 94.03 & 94.03 & 94.03 & 94.03 & 94.03 & 94.03
\\
\rowcolor{OursColor} \multirow{-2}{*}{\it step200} \cellcolor{white} & \ccheck & 
\bf 94.83 & \bf 94.72 & \bf 94.85 & \bf 94.60 & \bf 94.82 & \bf 94.55 & \bf \bf 94.67 & \bf 94.53 & \bf 94.70
\\
\specialrule{.15em}{.05em}{.05em}
\end{tabular}
}
\caption{
    \textbf{Comparison ($\uparrow$) on online MVTec} in pixel-level AUROC for anomaly segmentation across 4 configurations~\cite{ho2024ltad} of $\gD^{\tt T}$ and 8 configurations of $\gD^{\tt O}$, \ie~{\it B}, {\it B-HF}, {\it B-TF}, {\it D2-HF}, {\it D2-TF}, {\it D5-HF}, {\it D5-TF}, and {\it D5-M}. 
}
\label{table:supp-mvtec-online}

\end{table}

\begin{table}[h]

\centering
\setlength{\tabcolsep}{3pt}

\resizebox{\linewidth}{!}{%
\begin{tabular}{ccccccccccc}
\specialrule{.15em}{.05em}{.05em}
{Config.} & {Online} &
{\it B} &
{\it B-HF} &
{\it B-TF} &
{\it D2-HF} &
{\it D2-TF} &
{\it D5-HF} &
{\it D5-TF} &
{\it D5-M} &
{Avg.}
\\
\cmidrule(lr){1-11}
& \ccross &
97.05 & 97.05 & 97.05 & 97.05 & 97.05 & 97.05 & 97.05 & 97.05 & 97.05
\\
\rowcolor{OursColor} \multirow{-2}{*}{\it exp100} \cellcolor{white} & \ccheck & 
\bf 97.29 & \bf 97.20 & \bf 97.30 & \bf 97.06 & \bf 97.26 & \bf 97.16 & \bf 97.25 & \bf 97.14 & \bf 97.21
\\
& \ccross & 
96.91 & 96.91 & 96.91 & 96.91 & 96.91 & 96.91 & 96.91 & 96.91 & 96.91
\\
\rowcolor{OursColor} \multirow{-2}{*}{\it exp200} \cellcolor{white} & \ccheck &
\bf 97.09 & \bf 96.98 & \bf 97.14 & \bf 97.02 & \bf 97.05 & \bf 97.03 & \bf 96.97 & \bf 97.09 & \bf 97.05
\\
 & \ccross &
97.40 & 97.40 & 97.40 & 97.40 & 97.40 & 97.40 & 97.40 & 97.40 & 97.40
\\
\rowcolor{OursColor} \multirow{-2}{*}{\it step100} \cellcolor{white} & \ccheck &
\bf 97.61 & \bf 97.54 & \bf 97.65 & \bf 97.52 & \bf 97.55 & \bf 97.45 & \bf 97.50 & \bf 97.53 & \bf 97.54
\\
& \ccross & 
97.09 & 97.09 & 97.09 & 97.09 & 97.09 & 97.09 & 97.09 & 97.09 & 97.09
\\
\rowcolor{OursColor} \multirow{-2}{*}{\it step200} \cellcolor{white} & \ccheck & 
\bf 97.49 & \bf 97.43 & \bf 97.49 & \bf 97.17 & \bf 97.39 & \bf 97.19 & \bf 97.30 & \bf 97.29 & \bf 97.34
\\
\specialrule{.15em}{.05em}{.05em}
\end{tabular}
}
\caption{
    \textbf{Comparison ($\uparrow$) on online VisA} in the same format as Tab.~\ref{table:supp-mvtec-online}.
}
\label{table:supp-visa-online}

\end{table}

\begin{table}[h]

\centering
\setlength{\tabcolsep}{3pt}

\resizebox{\linewidth}{!}{%
\begin{tabular}{ccccccccccc}
\specialrule{.15em}{.05em}{.05em}
{Config.} & {Online} &
{\it B} &
{\it B-HF} &
{\it B-TF} &
{\it D2-HF} &
{\it D2-TF} &
{\it D5-HF} &
{\it D5-TF} &
{\it D5-M} &
{Avg.}
\\
\cmidrule(lr){1-11}
& \ccross & 
\bf 97.48 & 97.48 & 97.48 & \bf 97.48 & \bf 97.48 & \bf 97.48 & \bf 97.48 & \bf 97.48 & \bf 97.48
\\
\rowcolor{OursColor} \multirow{-2}{*}{\it exp100} \cellcolor{white} & \ccheck & 
97.47 & \bf 97.49 & \bf 97.50 & 97.42 & 97.47 & 97.42 & 97.38 & 97.30 & 97.43
\\
& \ccross & 
97.29 & 97.29 & 97.29 & 97.29 & 97.29 & 97.29 & 97.29 & 97.29 & 97.29
\\
\rowcolor{OursColor} \multirow{-2}{*}{\it exp200} \cellcolor{white} & \ccheck &
\bf 97.44 & \bf 97.44 & \bf 97.44 & \bf 97.40 & \bf 97.43 & \bf 97.40 & \bf 97.38 & \bf 97.03 & \bf 97.37
\\
 & \ccross &
 97.43 & 97.43 & 97.43 & 97.43 & 97.43 & 97.43 & 97.43 & \bf 97.43 & \bf 97.43
\\
\rowcolor{OursColor} \multirow{-2}{*}{\it step100} \cellcolor{white} & \ccheck &
\bf 97.57 & \bf 97.48 & \bf 97.56 & \bf 97.36 & \bf 97.47 & \bf 97.36 & \bf 97.50 & 97.17 & \bf 97.43
\\
& \ccross & 
97.31 & 97.31 & 97.31 & 97.31 & 97.31 & \bf 97.31 & 97.31 & \bf 97.31 & 97.31
\\
\rowcolor{OursColor} \multirow{-2}{*}{\it step200} \cellcolor{white} & \ccheck & 
\bf 97.47 & \bf 97.39 & \bf 97.43 & \bf 97.33 & \bf 97.42 & 97.21 & \bf 97.36 & 97.19 & \bf 97.35
\\
\specialrule{.15em}{.05em}{.05em}
\end{tabular}
}
\caption{
    \textbf{Comparison ($\uparrow$) on online DAGM} in the same format as Tab.~\ref{table:supp-mvtec-online}.
}
\label{table:supp-dagm-online}

\end{table}

\clearpage 

\end{document}